\documentclass[3p,times]{elsarticle}

\usepackage{graphicx,subfigure,epsfig,amsmath,amssymb,multirow,times}
\usepackage{epsfig,amsmath,amssymb,multirow,times}
\usepackage{algorithmic,balance}
\usepackage{algorithm}
\usepackage[update,prepend]{epstopdf}
\usepackage{color}
\usepackage[T1]{fontenc}
\usepackage{graphicx}
\usepackage{caption}
\usepackage{amssymb}

\usepackage[figuresright]{rotating}

\begin{document}

\begin{frontmatter}




\title{Fuzzy human motion analysis: A review}


\author{Chern Hong Lim*, Ekta Vats* and Chee Seng Chan**}

\address{Centre of Image and Signal Processing,\\
Faculty of Computer Science \& Information Technology, \\
University of Malaya, 50603 Kuala Lumpur, Malaysia \\
\textit{\{ch\_lim; ektavats\_2608\}@siswa.um.edu.my; cs.chan.um.edu.my}}
\let\thefootnote\relax\footnotetext{*Chern Hong Lim and Ekta Vats contributed equally to this paper.}
\let\thefootnote\relax\footnotetext{**Corresponding author.}

\begin{abstract}

Human Motion Analysis (HMA) is currently one of the most popularly active research domains as such significant research interests are motivated by a number of real world applications such as video surveillance, sports analysis, healthcare monitoring and so on. However, most of these real world applications face high levels of uncertainties that can affect the operations of such applications. Hence, the fuzzy set theory has been applied and showed great success in the recent past. In this paper, we aim at reviewing the fuzzy set oriented approaches for HMA, individuating how the fuzzy set may improve the HMA, envisaging and delineating the future perspectives. To the best of our knowledge, there is not found a single survey in the current literature that has discussed and reviewed fuzzy approaches towards the HMA. For ease of understanding, we conceptually classify the human motion into three broad levels: Low-Level (LoL), Mid-Level (MiL), and High-Level (HiL) HMA.

\end{abstract}

\end{frontmatter}

\section{Introduction}

\begin{figure}[htbp]
\centering
\includegraphics[scale=0.48]{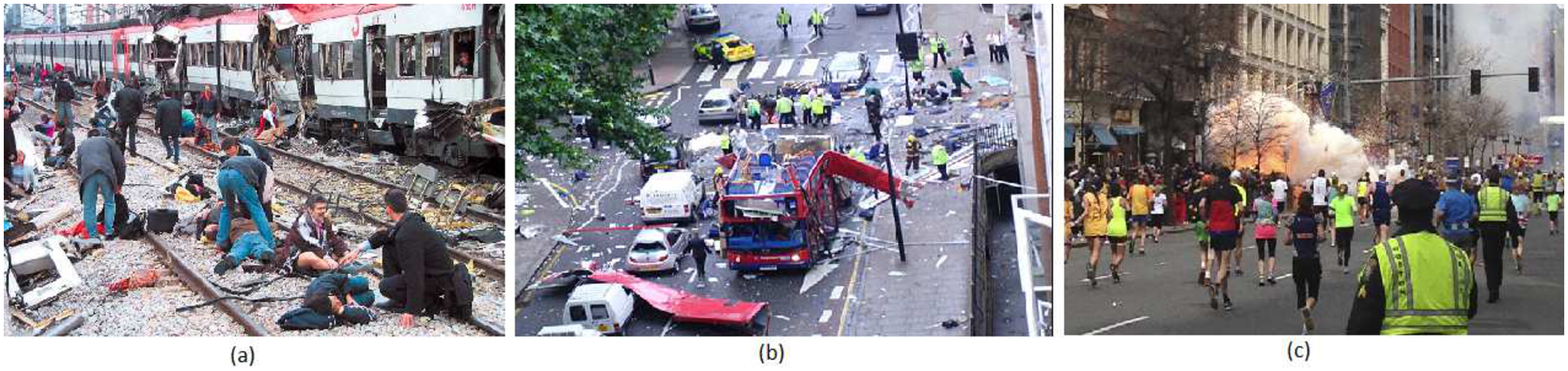}

\caption{(a) \textbf{Madrid train bombings:} On March 11 2004, Madrid commuter rail network was attacked and the explosions killed 191 people, injuring 1,800 others, (b) \textbf{London bombings:} July 7 2005 London bombings were a series of coordinated suicide attacks in the central London, which targeted civilians using the public transport system during the morning rush hour, (c) \textbf{Boston Marathon bombings:} On April 15 2013, two pressure cooker bombs exploded during the Boston Marathon, killing 3 people and injuring 264. Information source: \textit{http://en.wikipedia.org/}, Image source: \textit{http://images.google.com}.}

\label{fig:Fig_Bombings}
\end{figure}

Human motion analysis (HMA) refers to the analysis and interpretation of human movements over time. For decades, it has been a popular research topic that crossovers many domains such as biology \cite{bobick1997movement,troje2002decomposing}, psychology \cite{barclay1978temporal,blake2007perception}, multimedia \cite{kirtley2001application} and so on. In the computer vision domain, HMA has been emerging actively over the years due to the advancement of video camera technologies and the availability of more sophisticated computer vision algorithms in the public domain. Here, the HMA concerns the detection, tracking and recognition of human, and more generally the understanding of human behaviors, from image sequences involving humans. Amongst all, video surveillance is one of the most important real-time applications \cite{haering2008evolution,hu2004survey,kim2010intelligent,ko2008survey,popoola2012video}. For instance, as illustrated in Figure \ref{fig:Fig_Bombings}, the Madrid, London and Boston marathon bombing tragedies, happened in 2004, 2005 and 2013 respectively, would not have been worse if an intelligent video surveillance system that capable of automatically detecting abnormal human behavior was installed in the public areas. Apart from that, HMA also contributed in video retrieval \cite{geetha2008survey}, sports analysis \cite{efros2003recognizing,loy2004monocular,sullivan2008action}, healthcare monitoring \cite{anderson2006recognizing,anderson2009modeling}, human-computer interaction \cite{jaimes2007multimodal} and so on. 

\begin{table}[htbp]
	\centering
		\caption{Summarization of the survey papers on HMA.}
		\label{Table:Survey}
		\resizebox{15cm}{!} {
		\begin{tabular}{c p{4cm} p{7cm} p{11cm} c }\hline		
		
  		Paper & Author & Title & Description & Year \\ \hline \hline \\
  		 
		\cite{aggarwal1994articulated} & J.K. Aggarwal, Q. Cai, W. Liao \& B. Sabata & Articulated and elastic non-rigid motion: a review & The earliest survey on HMA, focusing on various methods used in the articulated and non-rigid motion. & 1994\\ \\
		
		\cite{cedras1995motion} & C. Cedras \& M. Shah & Motion-based recognition: a survey & An overview on various methods for motion extraction: action recognition, body parts recognition and body configuration estimation. & 1995 \\ \\
		
		\cite{aggarwal1997human} & J.K. Aggarwal \& Q. Cai & Human motion analysis: a review & Focus on motion analysis of human body parts, tracking moving human from a single view or multiple camera perspectives, and recognizing human activities from video. & 1997 \\ \\
		
		\cite{gavrila1999visual} & D.M. Gavrila & The visual analysis of human movement: a survey & Discussed various methodologies grouped into 2D approaches with or without explicit shape models as well as 3D approaches. & 1999 \\ \\
		
		\cite{pentland2000looking} & Alex Pentland & Looking at people: sensing for ubiquitous and wearable computing & Reviewed the state-of-the-art of "looking at people" focusing on person identification and surveillance monitoring. & 2000 \\ \\
		
		\cite{moeslund2001survey} & T.B. Moeslund \& E. Granum & A survey of computer vision-based human motion capture  & Overview on the taxonomy of system functionalities: initialization, tracking, pose estimation and recognition. & 2001 \\ \\
		
		\cite{wang2003recent} & H. Wang, W. Hu \& T. Tan & Recent Developments in Human Motion Analysis  & Focus on three major issues: human detection, tracking and activity understanding. & 2003 \\ \\
			
		\cite{hu2004survey} & W. Hu,  T. Tan, W. Liang \& S. Maybank & A survey on visual surveillance of object motion and behaviors & Reviewed recent developments in visual surveillance of object motion and behaviors in dynamic scenes and analyzed possible research directions. & 2004 \\ \\
		
		\cite{moeslund2006survey} & T. B. Moeslund, A. Hilton, \& V. Kruger & A survey of advances in vision-based human motion capture and analysis  & Discuss recent trends in video-based human motion capture and analysis. & 2006 \\ \\
		
		\cite{poppe2007vision} & R. Poppe & Vision-based human motion analysis: An overview  & HMA with two phases: modeling (concerned with construction of the likelihood function) and estimation (finding the most likely pose given the likelihood surface). & 2007 \\ \\
								
		\cite{turaga2008machine} & P. Turaga, R. Chellappa, V. Subrahmanian \& O. Udrea & Machine recognition of human activities: A survey & Addressed the problem of representation, recognition and learning of human activities from video and related applications. & 2008 \\ \\
								
		\cite{ji2010advances} & X. Ji \& H. Liu & Advances in view-invariant human motion analysis: A review & Emphasized on the recognition of poses and actions. Three major issues were addressed: human detection, view-invariant pose representation and estimation, and behavior understanding. & 2010 \\ \\
		
		\cite{poppe2010survey} & R. Poppe & A survey on vision-based human action recognition & Overview on current advances in vision-based human action recognition, addressing challenges faced due to variations in motion performance, recording settings and inter-personal differences. Also, discussed shortcomings of the state-of-the-art and outline promising directions of research. & 2010 \\ \\
		
		\cite{candamo2010understanding} & J. Candamo, M. Shreve, D. Goldgof, D. Sapper, \& R. Kasturi & Understanding transit scenes: A survey on human behavior-recognition algorithms & Reviewed automatic behavior recognition techniques, focusing on human activity surveillance in transit applications context. & 2010 \\ \\
		
		\cite{aggarwal2011human} & J. K. Aggarwal \& M. S. Ryoo & Human activity analysis: A review & Discussed methodologies developed for simple human actions as well as high-level activities. & 2011 \\ \\
				
		\cite{weinland2011survey} & D. Weinland, R. Ronfard \& E. Boyer & A survey of vision-based methods for action representation, segmentation and recognition & Concentrated on the approaches that aim at classification of full-body motions: kicking, punching and waving, and further categorized them according to spatial and temporal structure of actions, action segmentation from an input stream of visual data and view-invariant representation of actions. & 2011 \\ \\
		
		\cite{holte2011human} & M.B. Holte, T.B. Moeslund, C. Tran \& M.M. Trivedi & Human action recognition using multiple views: A comparative perspective on recent developments & Presented a review and comparative study of recent multi-view 2D and 3D approaches for HMA. & 2011 \\ \\
		
		\cite{lara2013survey} & O. Lara \& M. Labrador & A survey on human activity recognition using wearable sensors & Surveys human activity recognition based on wearable sensors. 28 systems were qualitatively evaluated in terms of recognition performance, energy consumption, and flexibility etc. & 2013 \\ \\
		
		\cite{chen2013survey} & L. Chen, H. Wei \& J. Ferryman & A survey of human motion analysis using depth imagery & Reviewed the research on the use of depth imagery for analyzing human activity (e.g. the Microsoft Kinect). Also listed publicly available datasets that include depth imagery. & 2013 \\ \\ 
		
		\cite{cristani2013human} & M. Cristani, R. Raghavendra, A. Del Bue \& V. Murino & Human behavior analysis in video surveillance: A social signal processing perspective & Analyzed the social signal processing perspective of the automated surveillance of human activities such as face expressions and gazing, body posture and gestures, vocal characteristics etc. & 2013 \\ \\
		
		\cite{chaquet2013survey} & J. M. Chaquet, E. J. Carmona \& A. F.-Caballero & A survey of video datasets for human action and activity recognition & Provide a complete description of the most important public datasets for video-based human activity and action recognition. & 2013 \\ \\
								
		\hline
								
		\end{tabular}
		}
\end{table}

\begin{table}[htbp]
	\centering
		\caption{Criterion on which the previous survey papers on HMA emphasized on (1994-2013). Note that those criterion without a `tick' means the topic is not discussed comprehensively in the corresponding survey paper, but might be touched indirectly in the contents.}
		\label{Table:Survey2}
		\resizebox{15cm}{!} {
		\begin{tabular}{c c c c c c c c c}\hline		
		
  		Year & Paper & Human Detection & Tracking & Behavior Understanding & Multi-view & Feature extraction & Datasets & Application\\ \hline \hline \\
  		 
		1994 & \cite{aggarwal1994articulated} & - & \checkmark & \checkmark & - & \checkmark & - & - \\ \\
		
		1995 & \cite{cedras1995motion} & - & \checkmark & \checkmark & - & \checkmark & - & - \\ \\
		
		1997 & \cite{aggarwal1997human} & \checkmark & \checkmark & \checkmark & \checkmark & \checkmark & - & - \\ \\
		
		1999 & \cite{gavrila1999visual} & \checkmark & \checkmark & \checkmark & \checkmark & \checkmark & - & \checkmark \\ \\
		
		2000 & \cite{pentland2000looking} & \checkmark & \checkmark & \checkmark & - & \checkmark & - & - \\ \\
		
		2001 & \cite{moeslund2001survey} & \checkmark & \checkmark & \checkmark & - & \checkmark & - & \checkmark \\ \\
		
		2003 & \cite{wang2003recent} & \checkmark & \checkmark & \checkmark & \checkmark & - & - & \checkmark \\ \\
		
		2004 & \cite{hu2004survey} & \checkmark & \checkmark & \checkmark & \checkmark & - & - & \checkmark \\ \\
		
		2006 & \cite{moeslund2006survey} & \checkmark & \checkmark & \checkmark & \checkmark & - & - & - \\ \\ 
		
		2007 & \cite{poppe2007vision} & \checkmark & \checkmark & - & - & \checkmark & - & - \\ \\
		
		2008 & \cite{turaga2008machine} & \checkmark & - & \checkmark & - & \checkmark & - & \checkmark \\ \\
		
		2010 & \cite{ji2010advances} & \checkmark & - & \checkmark & \checkmark & - & \checkmark & - \\ \\
		
		2010 & \cite{poppe2010survey} & - & - & \checkmark & \checkmark & \checkmark & \checkmark & - \\ \\ 
		
		2010 & \cite{candamo2010understanding} & \checkmark & \checkmark & \checkmark & - & - & - & - \\ \\ 
		
		2011 & \cite{aggarwal2011human} & - & - & \checkmark & - & \checkmark & \checkmark & \checkmark \\ \\
		
		2011 & \cite{weinland2011survey} & \checkmark & - & \checkmark & \checkmark & \checkmark & \checkmark & - \\ \\
		
		2011 & \cite{holte2011human} & - & - & \checkmark & \checkmark & \checkmark & \checkmark & - \\ \\
		
		2013 & \cite{lara2013survey} & - & - & \checkmark & - & \checkmark & \checkmark & - \\ \\ 
		
		2013 & \cite{chen2013survey} & \checkmark & \checkmark & \checkmark & - & - & \checkmark & - \\ \\ 
		
		2013 & \cite{cristani2013human} & \checkmark & \checkmark & \checkmark & - & - & - & \checkmark \\ \\ 
		
		2013 & \cite{chaquet2013survey} & - & - & - & - & - & \checkmark & - \\ \\
								
		\hline
								
		\end{tabular}
		}
\end{table}

The importance and popularity of the HMA system has led to several surveys in the literature, as indicated in Table \ref{Table:Survey}. One of the earliest surveys was \cite{aggarwal1994articulated}, focused on various methods employed in the analysis of the human body motion, which is in non-rigid form. \cite{cedras1995motion} gave an overview on the motion extraction methods using the motion capture systems and focused on action recognition, individual body parts recognition, and body configuration estimation. \cite{aggarwal1997human} used the same taxonomy as in \cite{cedras1995motion}, but engaging different labels for the three classes, that is further dividing the classes into subclasses yielding a more comprehensive taxonomy. \cite{gavrila1999visual} gave an overview on the applications of visual analysis of human movements, and their taxonomy covered the 2D and 3D approaches with and without the explicit shape models.

As the works in this area prosper, public datasets start to gain importance in the vision community to meet different research challenges. The KTH \cite{schuldt2004recognizing} and the Weizmann \cite{zelnik2001event,blank2005actions} datasets were the most popular human actions datasets introduced in the early stages. However, neither of the datasets represent the human actions in a real world environment. In general, each action is performed in a simple manner with just a single actor, static background and fixed view point. KTH however considered a few complex situations such as different lighting conditions, but it is still far away from the real world complex scenarios. Therefore, other datasets were created such as the CAVIAR, ETISEO, CASIA Action, MSR Action, HOLLYWOOD, UCF datasets, Olympic Sports and HMDB51, BEHAVE, TV Human Interaction, UT-Tower, UT-Interaction, etc. Please refer to \cite{chaquet2013survey} for a complete list of the currently available datasets in HMA.

Due to the advancement of the technology, using networks of multiple cameras for monitoring public places such as airports, shopping malls, etc. were emerged.  \cite{hu2004survey,aggarwal1997human,gavrila1999visual,wang2003recent,moeslund2006survey,ji2010advances,poppe2010survey,weinland2011survey,holte2011human} moved ahead to survey on the representation and recognition of the human actions in multiple-views aspect. Various new datasets were created exclusively for this purpose such as the IXMAS, i3DPost, MuHAVi, VideoWeb and CASIA Action. Last but not the least,  \cite{hu2004survey,gavrila1999visual,moeslund2001survey,wang2003recent,turaga2008machine,aggarwal2011human,cristani2013human} surveyed on the various applications of HMA such as the smart surveillance and advanced user interface for human-computer interaction. For the convenience of the readers, we summarize in Table \ref{Table:Survey} and \ref{Table:Survey2} the available survey papers and their respective focuses.

\subsection{Motivation and contributions}
\label{RW}

Fuzzy set theory since its inception in 1965, has played an important role in a variety of applications, for example the subway system in Sendai, Japan; washing machine; digital camera and so on. The research works on the fuzzy set theory in real world problems are abounded. In this paper, we will focus primarily on the solutions that utilize the fuzzy approaches towards HMA. Particularly, our main aim and contribution is to review the early years of the fuzzy set oriented approaches for HMA, individuating how the fuzzy set may improve the HMA, envisaging and delineating the future perspectives. This is in contrast to the past surveys as listed in Table \ref{Table:Survey} and \ref{Table:Survey2} where stochastic solutions were the predominant discussions. 

To the best of our knowledge, there is not found a single survey in the literature that has discussed and reviewed the fuzzy approaches towards HMA. The nearest studies to ours are \cite{huntsberger1986representation,krishnapuram1992fuzzy,sobrevilla2003fuzzy}. \cite{huntsberger1986representation} was the earliest survey that discussed the uncertainties in computer vision using the fuzzy sets. Specifically, it addressed the uncertainty in three levels: image segmentation, edge detection and shape representation. Later, \cite{krishnapuram1992fuzzy} gave a broad overview of the fuzzy set theory towards computer vision with applications in the areas of image modeling, preprocessing, segmentation, boundary detection, object/region recognition, and rule-based scene interpretation. The involved tasks were noise removal, smoothing, and sharpening of the contrast (low-level vision); segmentation of images to isolate the objects and the regions followed by the description and recognition of the segmented regions (intermediate-level vision); and finally the interpretation of the scene (high-level vision). Finally, \cite{sobrevilla2003fuzzy} addressed various aspects of image processing and analysis problems where the theory of fuzzy set was applied.

\begin{figure}[tb]
\centering
\includegraphics[width=6.4in]{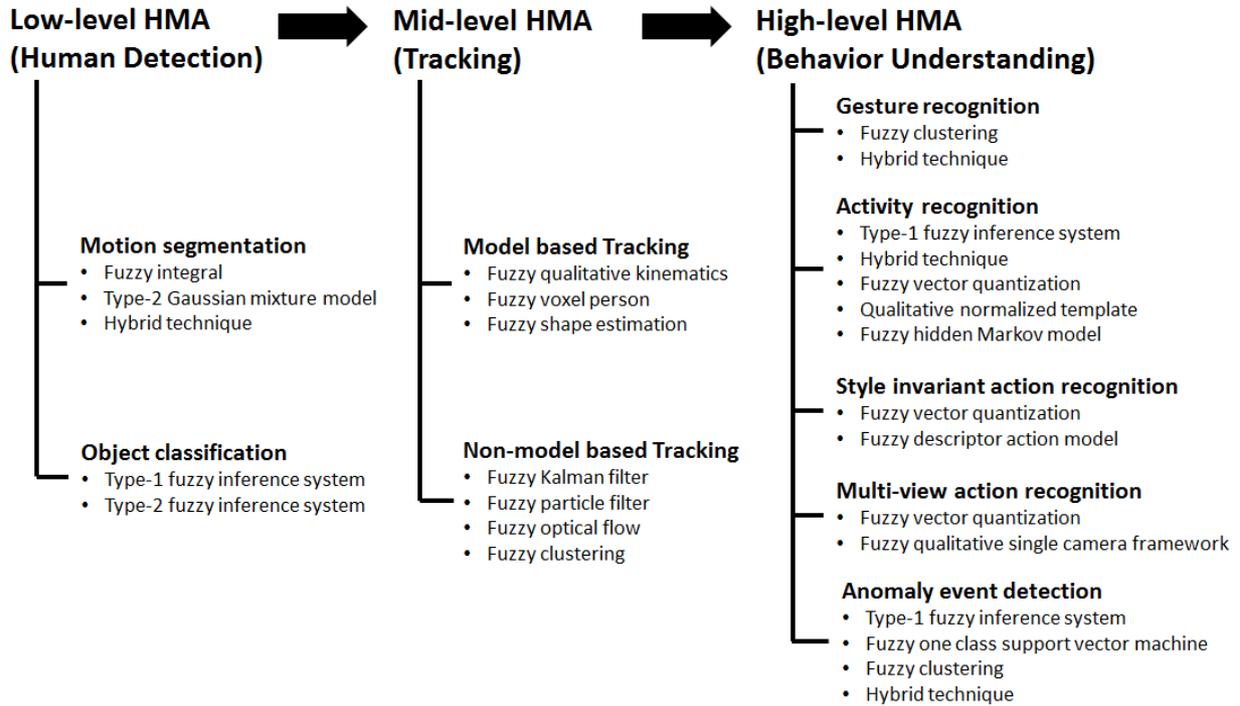}
\caption{Overall taxonomy of the review. It is organized according to the pipeline of HMA from Low-level to High-level with subcategories of the fuzzy approaches that have been employed in the literature.}
\label{fig:Overall}
\end{figure}

For ease of understanding, we conceptually classify the human motion into three broad levels: Low-Level (LoL), Mid-Level (MiL), and High-Level (HiL) HMA, as depicted in Figure \ref{fig:Overall}. The LoL HMA is the background/foreground subtraction which contributes in the pre-processing of the raw images to discover the areas of interest such as the human region. MiL HMA is the object tracking. In this level, it serves as the means to prepare data for pose estimation and activity recognition. HiL HMA is the behavior understanding where the objective is to correctly classify the human motion patterns into activity categories; for example, walking, running, wave hands and so on.

The remainder of this paper is organized as follows. Section \ref{LL} reviews the works on LoL HMA including the motion segmentation and the moving object classification. Section \ref{ML} covers the MiL HMA in terms of model-based and non-model based human tracking. The paper then extends the discussion to the HiL HMA that recognizes the human behavior in the image sequences in Section \ref{L}. Section \ref{dis} provides a detailed discussion on some advantages of the fuzzy approaches and presents some possible directions for the future research at length. Section \ref{con} concludes the paper.

\section{Low-level HMA}
\label{LL}

Human detection is the enabling step in almost every low-level vision-based HMA system before the higher level of processing steps such as tracking and behavior understanding can be performed. Technically, human detection aims at locating and segmenting the regions bounding the people from the rest of the image. This process usually involves first of all, the motion segmentation, and followed by the object classification. 

\subsection{Motion segmentation}

Motion segmentation aims at separating the moving objects from the natural scenes. The extracted motion regions are vital for the next level of processing, e.g. it relaxes the tracking complexity as only the pixels with changes are considered in the process. However, some critical situations in the real world environment such as the illumination changes, dynamic scene movements (e.g. rainy weather, waving tree, rippling water and so on), camera jittering, and shadow effects make it a daunting task. In this section, we will mainly review fuzzy approaches that had addressed the background subtraction problems. 

Background subtraction is one of the popular motion segmentation algorithms that has received much attention in the HMA system. This is due to the usefulness of its output that is capable of preserving the shape information, as well as helps in extracting motion and contour information \cite{bobick2001recognition,weinland2006free,lewandowski2010view}. In general, background subtraction is to differentiate between the image regions which have significantly different characteristics from the background image (normally denoted as the background model). A good background subtraction algorithm comprises of a background model that is robust to the environmental changes, but sensitive to identify all the moving objects of interest. There are some fuzzy approaches that endowed this capability in the background subtraction which will be discussed as follows.

\subsubsection{Fuzzy integral}

Information fusion from a variety of sources is the most straightforward and effective approach to increase the classification confidence, as well as removing the ambiguity and resolving the conflicts in different decisions. Rationally in background modeling, the combination of several measuring criteria (also known as the features or attributes) can strengthen the pixel's classification as background or foreground. However the basic mathematical operators used for aggregation such as the minimum, maximum, average, median, `AND', and `OR' operators provide crisp decisions and utilize only a single feature that tends to result in false positive \cite{zhang2006fusing}. In contrast, the fuzzy integrals take into account the importance of the coalition of any subset of the criteria \cite{el2008fuzz}. 

\begin{figure}[htbp]
\centering
\includegraphics[scale=0.4]{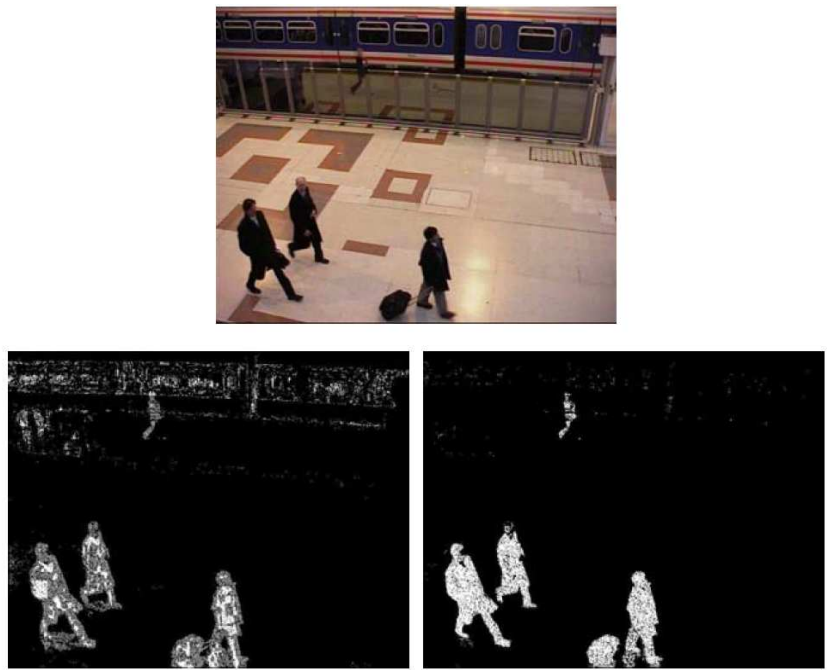}
\caption{Comparison between the Sugeno and the Choquet fuzzy integral methods for background subtraction \cite{el2008fuzz}. First row: The original image. Second row: the output from the Sugeno fuzzy integral on the left and the Choquet fuzzy integral on the right.}
\label{fig:Fig_LoL_FuzzIntegral}
\end{figure}

In general, the fuzzy integral is a non-linear function that is defined with respect to the fuzzy measure such as a belief or a plausibility measure \cite{tahani1990information}, and is employed in the aggregation step. As the fuzzy measure in the fuzzy integral is defined on a set of criteria, it provides precious information about the importance and relevance of the criteria to the discriminative classes. Thus it achieves feature selection with better classification results. \cite{zhang2006fusing} proposed to use the Sugeno integral \cite{marichal2000sugeno} to fuse color and texture features in their works for better classification of the pixel that belongs to either background or foreground, while \cite{el2008fuzz,el2008fuzzy,balcilar2013region} improved \cite{zhang2006fusing} by replacing the Sugeno integral with the Choquet integral \cite{murofushi1989interpretation}. The main reason is that the Choquet integral which was adapted for cardinal aggregation, was found to be more suitable than the Sugeno integral that assumed the measurement scale to be ordinal \cite{sugeno1995new,narukawa2004decision}. The corresponding results for the comparison between the Sugeno integral and the Choquet integral are shown in Figure \ref{fig:Fig_LoL_FuzzIntegral}. The background modeling process using the fusion of color and texture features have shown to achieve better detection of the moving targets against cluttered backgrounds, backgrounds with little movements, shadow effects as well as illumination changes.

\subsubsection{Type-2 Gaussian mixture model}
\label{T2GMM}

The studies on the background subtraction \cite{piccardi2004background,cheung2004robust} have shown that the Gaussian Mixture Model (GMM) is one of the popular approaches used in modeling the dynamic background scene. It solves the limitation in the unimodal model (single Gaussian) which is unable to handle the dynamic backgrounds such as waving tree and water rippling. The expectation-maximization algorithm is normally used in the initialization step of the GMM to estimate the parameters from a training sequence using the Maximum-likelihood (ML) criterion. However, due to insufficient or noisy training data, the GMM may not be able to accurately reflect the underlying distribution of the observations. This is because exact numbers must be used in the likelihood computation and unfortunately, these parameters are bounded by uncertainty. In order to take into account the uncertainty, the fuzzy set theory was explored. 

\begin{figure}[htbp]
\centering
\includegraphics[scale=0.4]{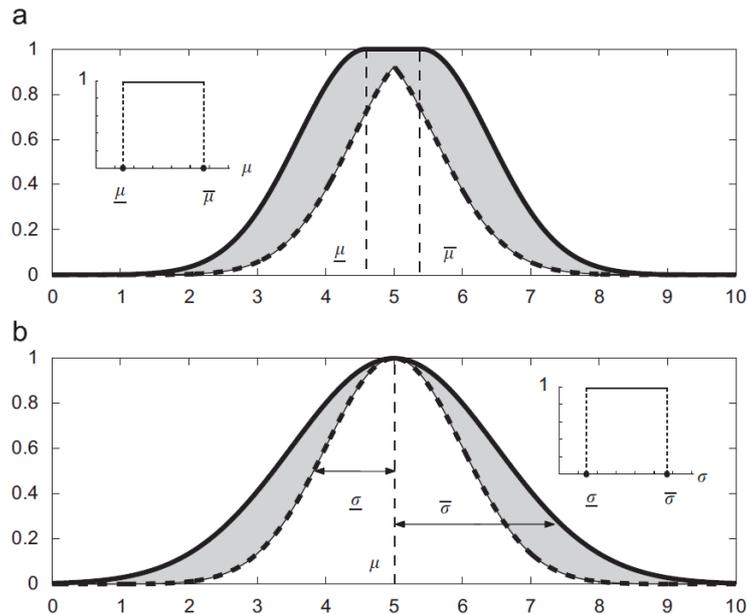}
\caption{Example of the type-2 fuzzy membership function of the Gaussian model with (a) uncertain mean, $\mu$ and (b) uncertain standard deviation, $\sigma$, having uniform possibilities. The shaded region is the Footprint of Uncertainty (FOU). The thick solid and dashed lines denote the lower and upper membership functions \cite{zeng2008type}.} 
\label{fig:Fig_LoL_FGMM}
\end{figure}

However, there has been an argument that type-1 fuzzy set, which is an ordinary fuzzy set \cite{zadeh1965fuzzy}, has limited capability in modeling the uncertainty. This is because the membership function for the type-1 fuzzy set is not associated with uncertainty. Therefore, type-2 fuzzy sets \cite{mendel2002type} emerged from the type-1 fuzzy set by generalizing it to handle more uncertainty in the underlying fuzzy membership function. As a whole, the type-2 fuzzy membership function is itself a fuzzy set and referring to Figure \ref{fig:Fig_LoL_FGMM}, it can be noticed that the uncertainty in the fuzzy membership function is represented in the shaded area known as the Footprint of Uncertainty (FOU). With the capability of type-2 fuzzy set to handle higher dimensions of uncertainty, it was adopted in \cite{zeng2008type} to represent the multivariate Gaussian with an uncertain mean vector or a covariance matrix. In more detail, it was assumed that the mean and the standard deviation vary within the intervals with uniform possibilities (Figure \ref{fig:Fig_LoL_FGMM}), instead of crisp values as in the conventional GMM.

Several works \cite{el2008type,el2009fuzzy,bouwmans2009modeling} have been reported that utilized the type-2 fuzzy GMM to deal with insufficient or noisy data, and resulted in better background subtraction model. In the later stage, \cite{zhao2012fuzzy} made an improvement on these works with the inclusion of spatial-temporal constraints into the  type-2 fuzzy GMM by using the Markov Random Field.

\subsubsection{Hybrid technique}
\label{HM_1}

Although the fuzzy approaches provide superior performance in background subtraction, most of these approaches have a common problem, that is how to optimize the parameters in their algorithms. These parameters can be the intrinsic parameters such as the interval values of the membership function, or the threshold value for the inference step. Optimizing these parameters usually increases the overall system performance. However, such steps require human intervention \cite{zhang2006fusing,el2008fuzzy,el2008fuzz}. For example, the trial and error process to determine a classification threshold value is a tedious job, computationally expensive and subjective \cite{sigari2008fuzzy}.

Fortunately, such limitations can be handled by using hybrid techniques, i.e. the combination of fuzzy approaches with machine learning methods. \cite{lin2000neural} applied neural fuzzy framework to estimate the image motion. The back-propagation learning rule from a five-layered neural fuzzy network was used to choose the best membership functions so that the system is able to adapt to different environments involving occlusions, specularity, shadowing, transparency and so on. Besides that, \cite{maddalena2010fuzzy} introduced a spatial coherence variant incorporated with the self-organizing neural network to formulate a fuzzy model to enhance the robustness against false detection in the background subtraction algorithm. \cite{li2012adaptive} used both the particle swarm optimization and the kernel least mean square to update the system parameters of a fuzzy model, and \cite{calvo2013fuzzy} employed a tuning process using the Marquardt-Levenberg algorithm within a fuzzy system to fine-tune the membership function. In order to determine the appropriate threshold value for the classification task, \cite{shakeri2008novel} proposed a novel fuzzy-cellular method that helps in dynamically learning the optimal threshold value.

\subsection{Object classification}

The outcome from the motion segmentation usually results in a rough estimation of the moving targets in a natural scene. These moving targets in a natural scene can be shadow, vehicle, flying bird and so on. Before the region is further processed at the next level, it is very important to verify and refine the interest object by eliminating the unintended objects.  In this section, we discuss some fuzzy approaches that are beneficial in the human object classification.

\subsubsection{Type-1 fuzzy inference system}
\label{LoL_OC}

\begin{figure}[htb]
\centering
\includegraphics[scale=0.35]{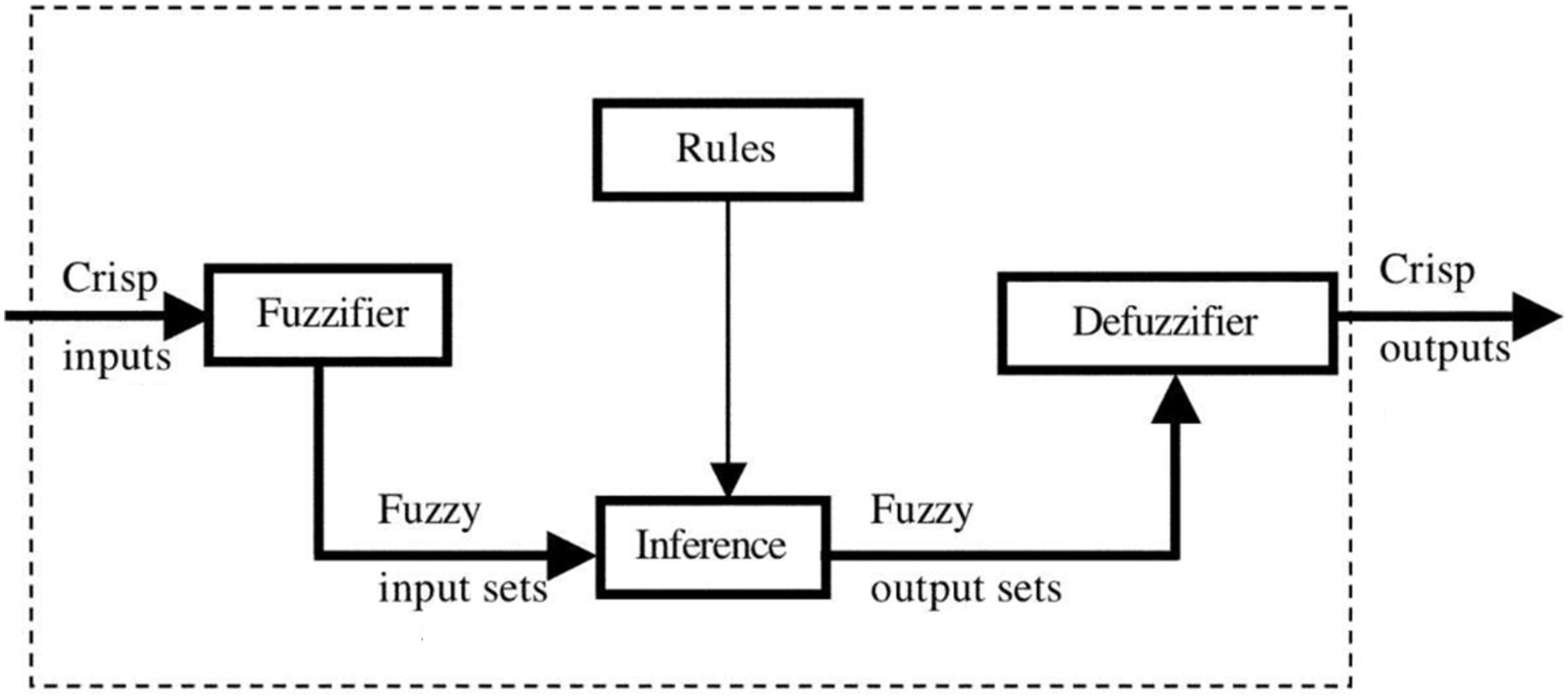}
\caption{Type-1 Fuzzy Inference System \cite{mendel2006interval}.}
\label{fig:Fig_LoL_T1FIS}
\end{figure}

The Type-1 Fuzzy Inference System (FIS) \cite{yager1992introduction} is a complete fuzzy decision making system that utilizes the fuzzy set theory. It has been successfully applied in numerous applications for commercial and research purposes. Its popularity is due to the capability to model the uncertainty and the sophisticated inference mechanism that greatly compromises the vague, noisy, missing, and ill-defined data in the data acquisition step. Figure \ref{fig:Fig_LoL_T1FIS} shows the overall framework of a typical Type-1 FIS, where it includes three important steps: fuzzification, inference, and defuzzification. The fuzzification step maps the crisp input data from a set of sensors (features or attributes) to the membership functions to generate the fuzzy input sets with linguistic support \cite{zadeh1988fuzzy}. Then, the fuzzy input sets go through the inference steps with the support from a set of fuzzy rules to infer the fuzzy output sets. Finally, the fuzzy output sets are defuzzified into the crisp outputs.

\begin{figure}[htb]
\centering
\subfigure[]{\includegraphics[scale=0.24]{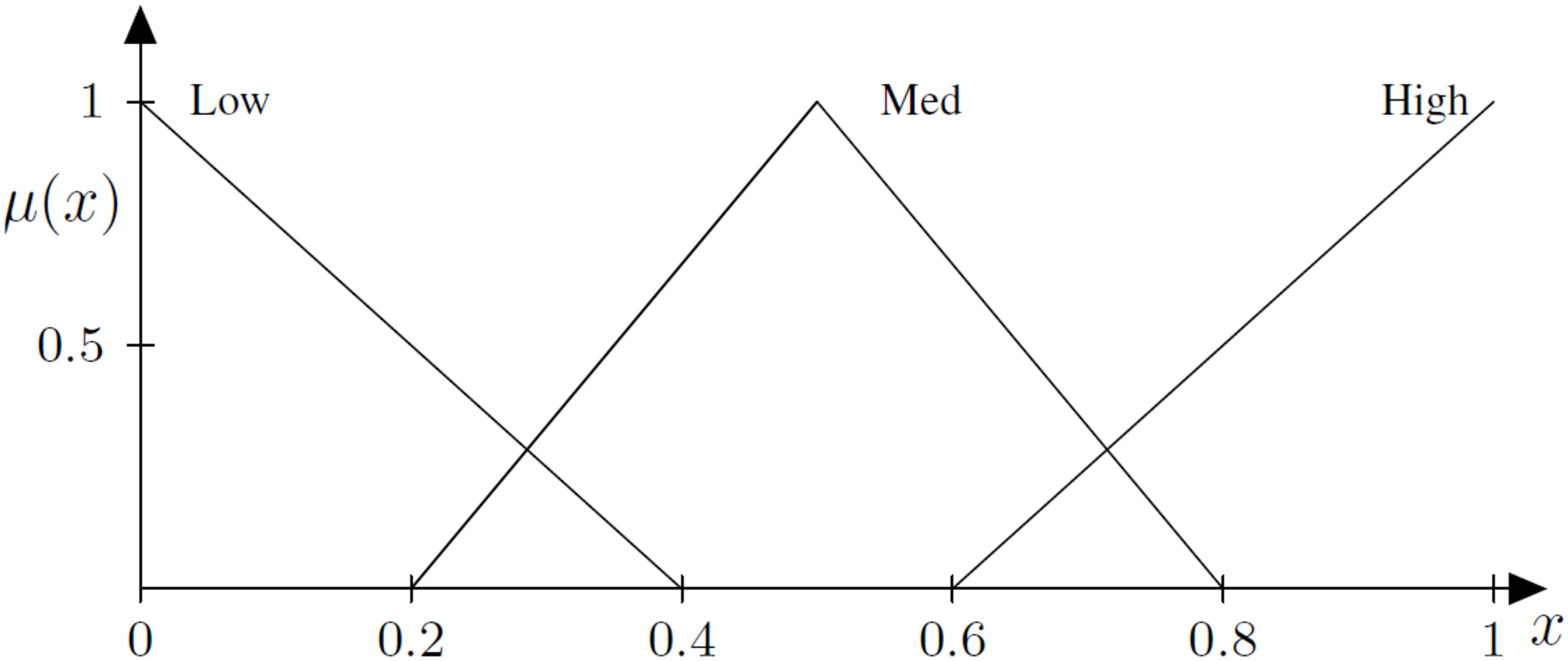}
\label{fig:Fig_LoL_MF}}
\subfigure[]{\includegraphics[scale=0.4]{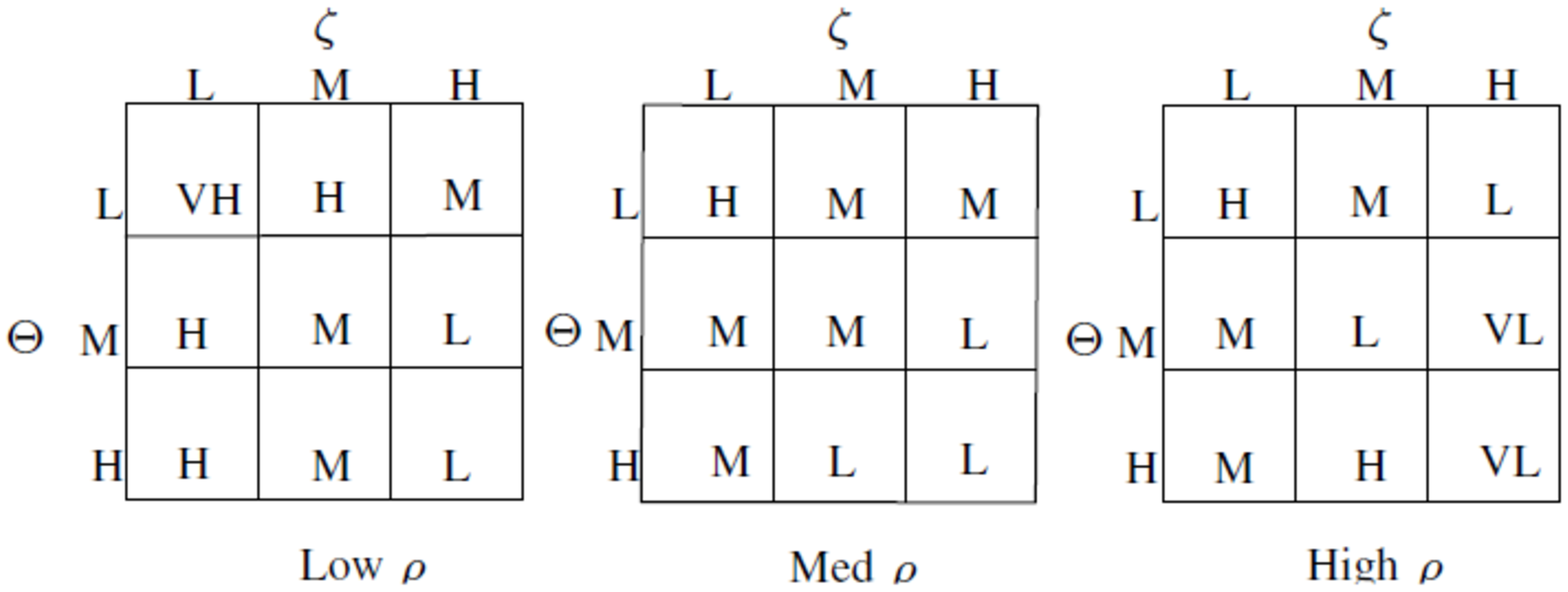}
\label{fig:Fig_LoL_Rules}}
\caption{(a) Example of the membership function for the distance feature where $\mu(x)$ denotes the membership value, and $x$ is the distance value. (b) The fuzzy rules for the fuzzy input for three features (Distance, $\rho$; Angle, $\Theta$ and Cord to Arc Ratio, $\zeta$), and its corresponding fuzzy output (VL=Very low, L=Low, M=Med, H=High, VH=Very High) \cite{mahapatra2013background}. }
\label{fig:Fig_MiL_FIS}
\end{figure}

In human detection, the FIS is an effective and direct approach to distinguish between the human and non-human with different features \cite{see2005human,mahapatra2013background,chowdhury2014detection}. As an example, \cite{mahapatra2013background} extracted three features from the contours of the segmented region, such as the distance to the centroid, angle, and cord to arc ratio, and input them into the FIS. The corresponding fuzzy membership function and a set of fuzzy rules were used to infer the fuzzy output as depicted in Figure \ref{fig:Fig_MiL_FIS}. The fuzzy outputs (VL, L, M, H, VH) were then defuzzified into the crisp outputs, and used to perform human classification. For example, if the crisp output is found to be less than the threshold value, then it is recognized as a human and vice versa.

Besides that, \cite{chen2006adaptive,Chen2006} studied in depth about the problems encountered in the human classification tasks, such as the situations where the unintended objects are attached to the classified human region. This problem often occurs in the silhouette based classification output. In general, silhouette is the binary representation of the segmented regions from the background subtraction techniques, where in HMA, human silhouette has proved its sufficiency to describe the activities captured by the video \cite{bobick2001recognition,weinland2006free,lewandowski2010view}. For example. a chair that is being moved by a person can be misclassified as a part of the segmented region, and included as part of the silhouette image. In order to solve this, \cite{chen2006adaptive,Chen2006} applied the FIS to perform an adaptive silhouette extraction in the complex and dynamic environments. In their works, they used multiple features such as the sum of absolute difference (SAD), fraction of neighbor blocks, and distance between blocks and human body centroid. A set of fuzzy rules were generated, for instance, ``IF SAD is SMALL, AND the fraction of neighboring silhouette blocks belong to the human body is LARGE, AND the distance from the centroid is SMALL, THEN the new block is more likely to be a human silhouette block''. Depending upon the application, the FIS is capable of modeling different sources of features by generating the appropriate fuzzy membership functions and the fuzzy rules.

\subsubsection{Type-2 fuzzy inference system}

\begin{figure}[htb]
\centering
\includegraphics[scale=0.35]{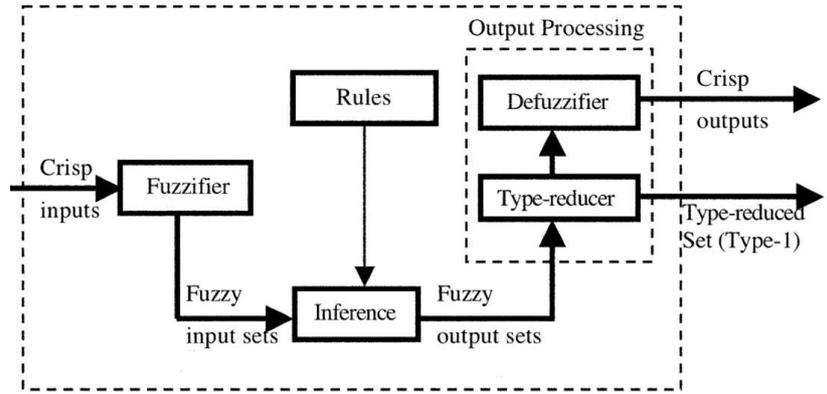}
\caption{Type-2 Fuzzy Inference System \cite{mendel2006interval}.}
\label{fig:Fig_LoL_T2FIS}
\end{figure}

To a certain extent, the overall performance of the system from \cite{chen2006adaptive,Chen2006} may be degraded due to the misclassification of the objects in the proposed type-1 FIS. Taking this into account, \cite{yao2012interval} employed the interval type-2 FIS \cite{liang2000interval} which is capable of handling higher uncertainty levels present in the real world dynamic environments. 

In general, as aforementioned, the type-2 FIS differs from the type-1 FIS in terms of the type-2 FIS offers the capability to support higher dimensions of uncertainty. The main focus in the type-2 FIS is the membership function that is used to represent the input data, where the membership function itself is a fuzzy set with FOU bounded in an ordinary membership function. In consequences, the input data is first fuzzified into type-2 input fuzzy sets, and then go through the inference process where the rules can be similar as the type-1 FIS. Before the defuzzification step takes place, the type-2 output fuzzy sets must be reduced from type-2 to type-1 output fuzzy set. This is processed by using a type-reducer, as depicted in Figure \ref{fig:Fig_LoL_T2FIS}.

\begin{figure}[h]
\centering
\includegraphics[scale=0.4]{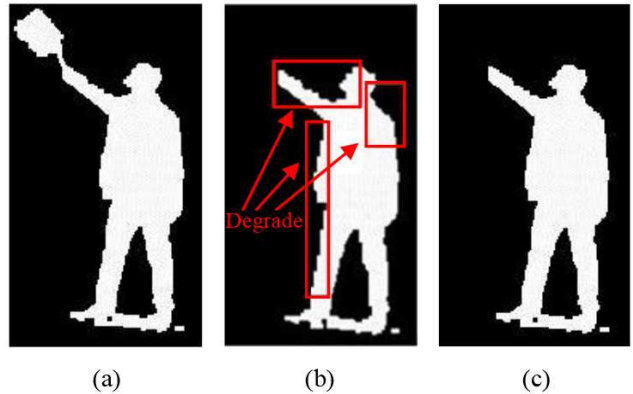}
\caption{Background subtraction on the image of a person raising a book. (a) Extracted silhouette by using the GMM, but is unable to eliminate the unintended object (book). (b) Extracted silhouette after using type-1 FIS to detach the book from the human, but degraded as a result \cite{Chen2006,chen2006adaptive}. (c) Extracted silhouette after using type-2 FIS where the result is much smoother.}
\label{fig:Fig_LoL_T2FIS_2}
\end{figure}

Using the same features as \cite{chen2006adaptive,Chen2006}, \cite{yao2012interval} proposed to fuzzify the input feature values into the type-2 fuzzy sets using the singleton fuzzification method \cite{karnik1998type}. Consequently, it produces the interval type-2 membership functions for the inference process. Their approach was tested on a set of images captured from the real world environment that contains single person, multi-person and the crowded scenes, respectively. The ground truth data was captured from the cameras deployed around their laboratory (i.e. a smart living room) to analyze people's regular activities. Their proposed work showed that the type-2 FIS provides much better results as compared to the type-1 FIS (Figure \ref{fig:Fig_LoL_T2FIS_2}). For ease of understanding of the readers, the problems of the works in LoL HMA with the intuition of using fuzzy approaches as the resolution are summarized in Table \ref{Table:LoLSurvey}.

\begin{table}[htbp]
	\centering
		\caption{Summarization of research works in LoL HMA using the Fuzzy approaches.}
		\label{Table:LoLSurvey}
		\resizebox{15cm}{!} 
		{
		\begin{tabular}{| l | p{7cm}  c  p{7cm}  c |}\hline		
				
		\multirow{2}{*}{LoL processing} & Problem statements / & \multirow{2}{*}{Papers} & \multirow{2}{*}{Why fuzzy?}  & \multirow{2}{*}{Approach} \\  
		  		
		& Sources of Uncertainty & & & \\ \hline \hline
		  		
		& & & & \\
		
		Motion segmentation & Critical situations such as illumination changes, dynamic scene movements, camera jittering, and shadow effects confuse the pixels belonging to the background model or the foreground object. &  \cite{zhang2006fusing,el2008fuzz,el2008fuzzy,balcilar2013region} & Information fusion from a variety of sources using the fuzzy aggregation method relaxes the crisp decision problem that causes confusion in the specific class. & Fuzzy integral\\ 
				
		& & & & \\		
		\cline{2-5}
		& & & & \\	
			
		& Insufficient and noisy training data do not accurately reflect the distribution in an ordinary GMM background modeling process. &\cite{el2008type,el2009fuzzy,bouwmans2009modeling,zhao2012fuzzy} & The uncertainty in GMM is bounded with interval mean and standard deviation instead of the crisp values. Type-2 fuzzy set is utilized to handle higher dimensions of uncertainty within the type-1 membership itself. & Type-2 Fuzzy GMM\\
		 
		& & & & \\		  
		\cline{2-5}	
		& & & & \\
				
		& Difficulty in determining the optimum parameters in the fuzzy system such as the membership function or the threshold value for the decision making process in the background subtraction algorithms. & \cite{lin2000neural,maddalena2010fuzzy,li2012adaptive,calvo2013fuzzy,shakeri2008novel} & Integration of the machine learning techniques with the fuzzy approaches allow the system to learn the optimum parameters that leads to better overall system performance and the feasibility to adapt to various situations depending on the task in hand. & Hybrid technique\\
		
		& & & & \\		
		\hline		
		& & & & \\
				
		Object classification & The confusion between the human and non-human objects, and the unintended objects attached to the human region causes the uncertainty in the classification tasks.  & \cite{mahapatra2013background,see2005human,chowdhury2014detection,chen2006adaptive,Chen2006} & Type-1 FIS is able to model the uncertainty in the features data as the membership function, and perform inference using the fuzzy rules to achieve better classification results. & Type-1 FIS\\ 		
		
		& & & & \\		  
		\cline{2-5}	
		& & & & \\
		
		& The insufficiency of the type-1 FIS causes the misclassification of the objects and the degradation in the silhouette extraction. & \cite{yao2012interval} & Type-2 fuzzy set offers the capability to support higher dimensions of uncertainty where in this case, the smoother classification results can be obtained. & Type-2 FIS\\
		
		& & & & \\
		\hline
								
		\end{tabular}
		}
\end{table}


\section{Mid-level HMA}
\label{ML}

After we have successfully located the human in the frame, the next step is to track the human movements over time for the higher level interpretation. Tracking is a crucial step in HMA as it forms the basis for data preparation for HiL HMA tasks such as action recognition, anomaly event detection and so on. The aim of tracking algorithm is to reliably track the interest objects such as the human body from a sequence of images, and it can be categorized as model based and non-model based motion tracking.

\subsection{Model based tracking}

In the model based human motion tracking, the human body models such as the stick figures, 2D and 3D motion description models are adopted to model the complex, non-rigid structure of the human body \cite{guo1994tracking,leung1995first,iwai1999posture,silaghi1998local,niyogi1994analyzing,ju1996cardboard,rohr1994towards,wachter1997tracking,rehg1995model,kakadiaris1996model}. Readers can refer to \cite{aggarwal1997human,gavrila1999visual,wang2003recent,Moeslund2001} for the detailed reviews. The stick figure model represents the human body as a combination of sticks or line segments connected by the joints \cite{guo1994tracking,leung1995first,iwai1999posture,silaghi1998local}, while the 2D models represents the human body using 2D ribbons or blobs \cite{leung1995first,niyogi1994analyzing,ju1996cardboard}. 3D models are used to depict the human body structure in a more detailed manner using cones, cylinders, spheres, ellipses etc. \cite{rohr1994towards,wachter1997tracking,rehg1995model,kakadiaris1996model}. 

However, tracking human in video sequences is not an easy task. The human body has a complex non-rigid structure consisting of a number of joints (e.g. the leg is connected to the foot by the ankle joint) and each body part can therefore move in a high degree of freedom around its corresponding joints. This often results in self-occlusions of the body parts. 3D models are able to handle such scenarios, but there are other factors that can affect the tracking performance such as the monotone clothes, cluttered background and changing brightness \cite{ning2004kinematics}. Therefore, the fuzzy approaches such as the fuzzy qualitative kinematics, the fuzzy voxel person, and the fuzzy shape estimation are explored in the model based human motion tracking algorithms to handle the uncertainties.

\subsubsection{Fuzzy qualitative kinematics}
\label{MiL_FQK}

\begin{figure}[htbp]
\centering
\subfigure[]{\includegraphics[scale=0.25]{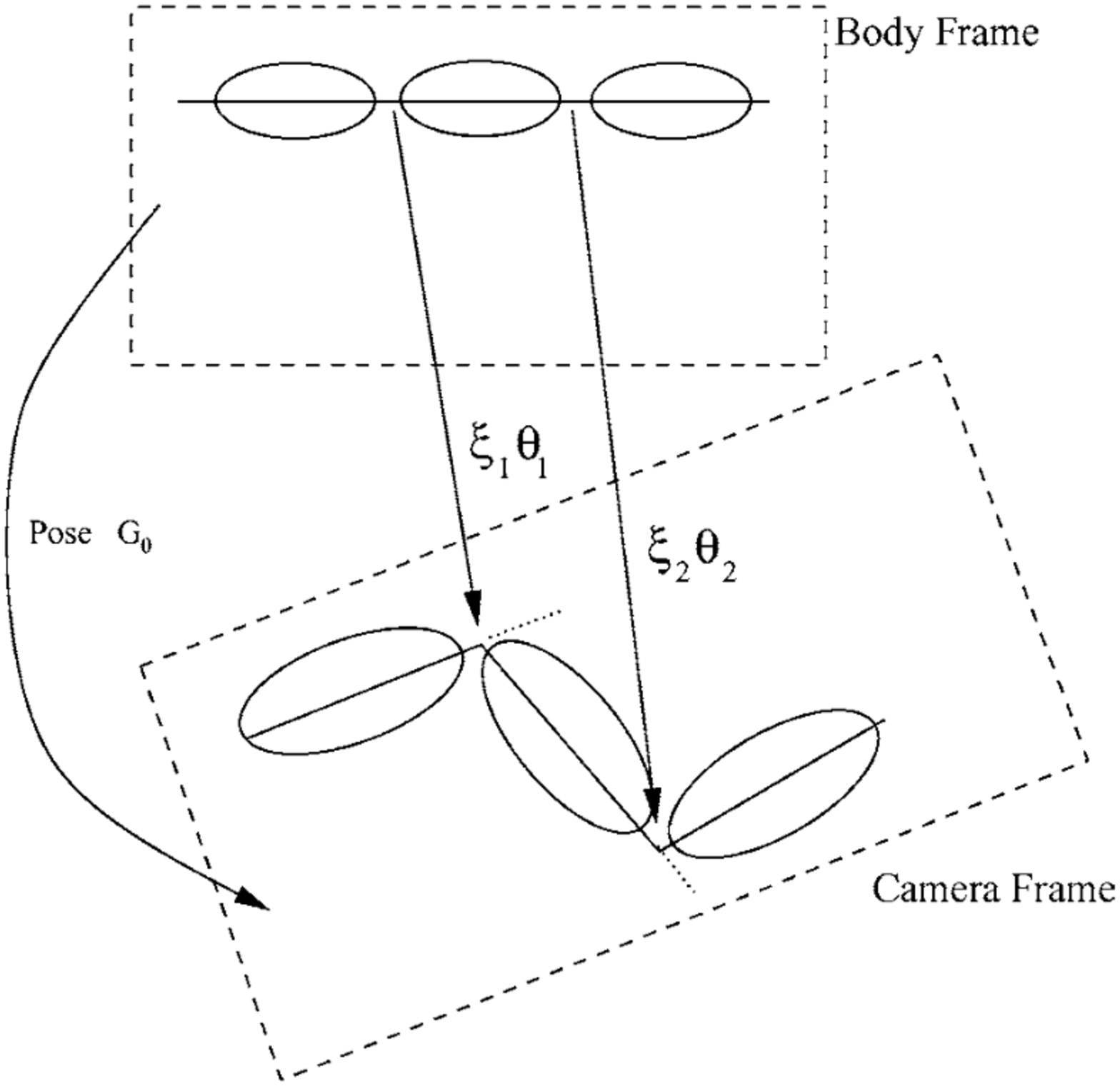}
\label{fig:Fig_MiL_Kinematic}}
\subfigure[]{\includegraphics[scale=0.3]{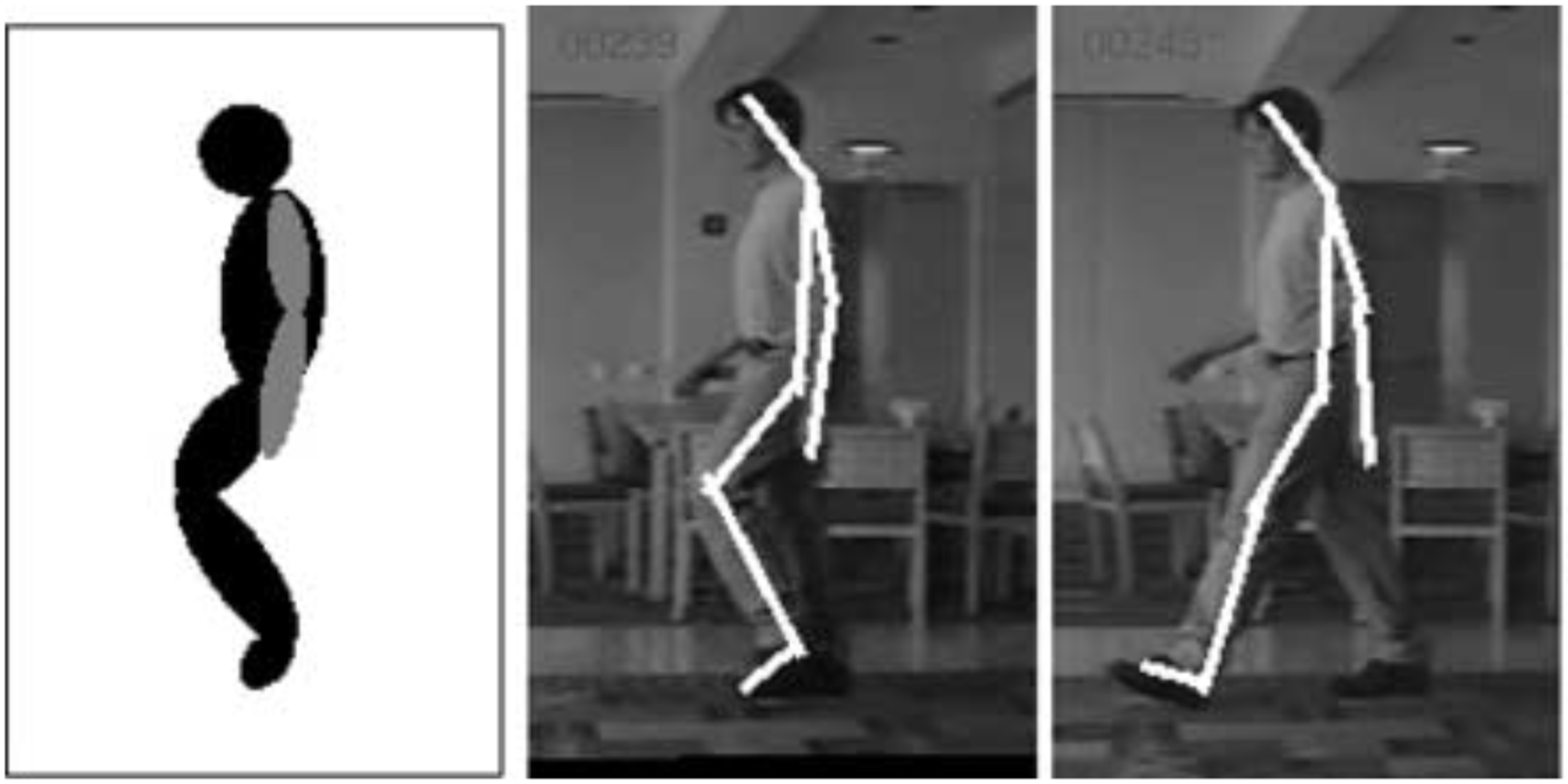}
\label{fig:Fig_MiL_Kinematic2}}
\caption{(a) Kinematic chain defined by twist \cite{bregler2004twist}, and (b) The estimated kinematic chain on the human body while performing the walking action.}
\label{fig:Fig_MiL_KM}
\end{figure}

A variety of works in the model based human motion tracking have employed the kinematic chain \cite{guo1994tracking,leung1995first,iwai1999posture,silaghi1998local,niyogi1994analyzing,ju1996cardboard,rohr1994towards,wachter1997tracking,rehg1995model,kakadiaris1996model}. Bregler et al. \cite{bregler2004twist} demonstrated a comprehensive visual motion estimation technique using the kinematic chain in a complex video sequence, as depicted in Figure \ref{fig:Fig_MiL_KM}. However, the crisp representation of the kinematic chain has a limitation. It suffers from the precision problem \cite{Liu2008a} and the cumulative errors can directly affect the performance of the higher level tasks. Therefore, a better strategy is required to model the kinematic chain, and to this end, the fuzzy qualitative kinematics has been proposed.

\begin{figure}[htbp]
\centering
\subfigure[]{\includegraphics[scale=0.3]{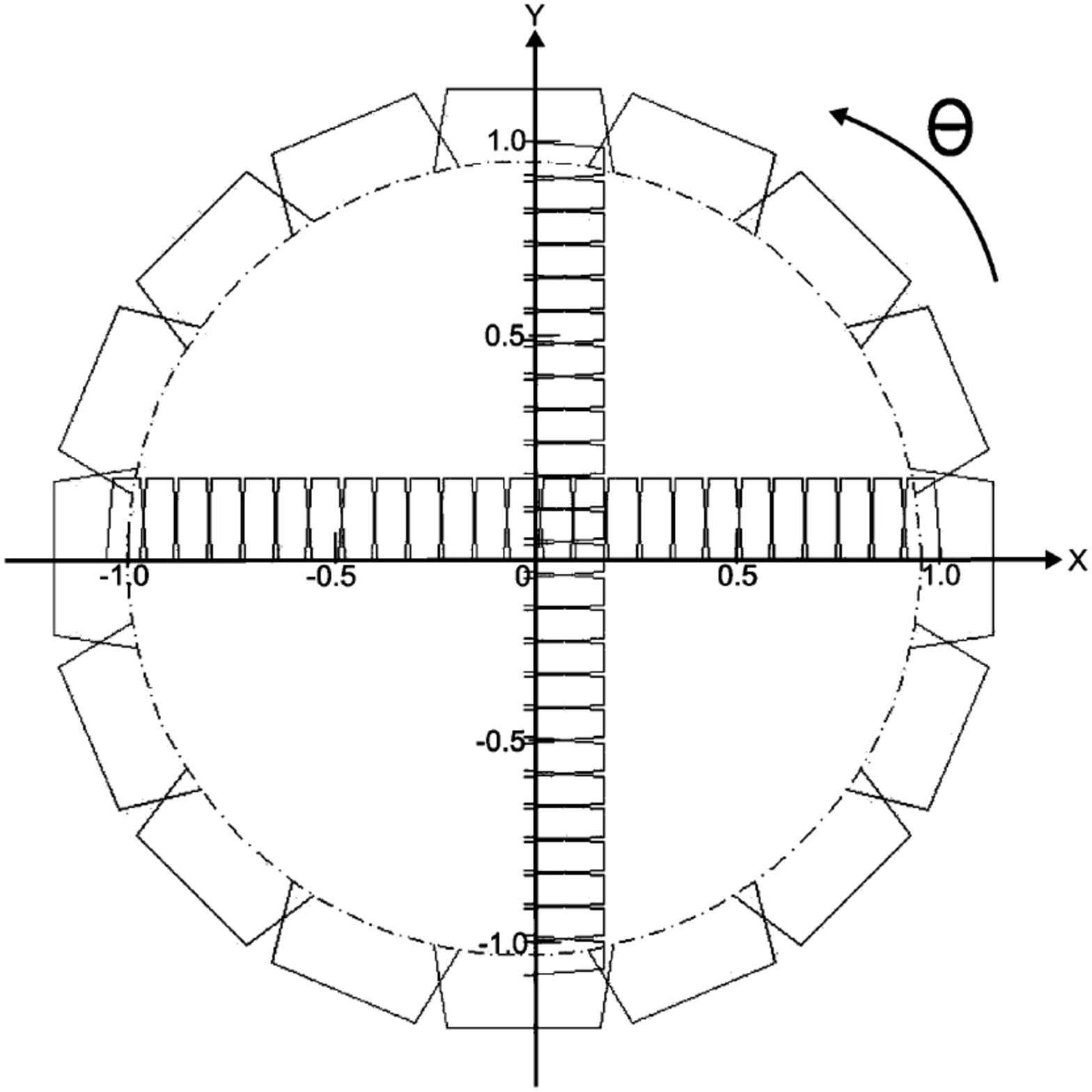}
\label{fig:Fig_MiL_FQS_1}}
\subfigure[]{\includegraphics[scale=0.3]{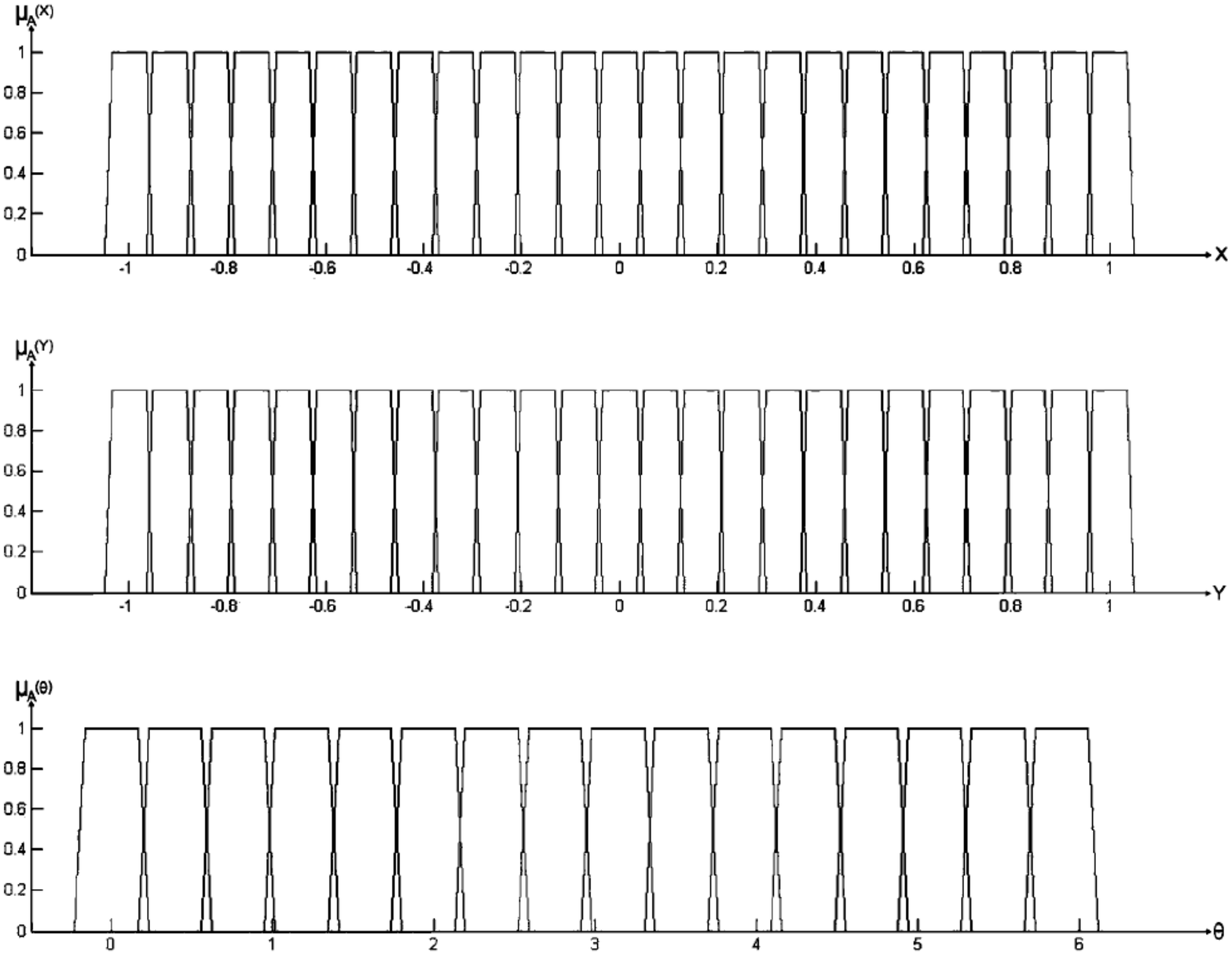}
\label{fig:Fig_MiL_FQS_2}}
\caption{(a) Description of the Cartesian translation and the orientation in the conventional unit circle replaced by the fuzzy quantity space. (b) Element of the fuzzy quantity space for every variable (translation ($X$, $Y$), and orientation $\theta$) in the fuzzy qualitative unit circle is a finite and convex discretization of the real number line \cite{chan2009fuzzy}.}
\label{fig:Fig_MiL_FQS}
\end{figure}

To begin with, the fuzzy qualitative reasoning \cite{shen1993fuzzy,chan2011recent} is a form of approximate reasoning that can be defined as the fusion between the fuzzy set theory \cite{zadeh1965fuzzy} and the qualitative reasoning \cite{kuipers1986qualitative}. The qualitative reasoning operates with the symbolic `quantities', while the fuzzy reasoning reasons with the fuzzy intervals of varying precisions, providing a means to handle the uncertainty in a natural way. Therefore, the fuzzy qualitative reasoning incorporates the advantages of both the approaches to alleviate the hard boundary or the crisp values of the ordinary measurement space. For instance, \cite{liu2009fuzzy} applied this in the Fuzzy Qualitative Trigonometry (Figure \ref{fig:Fig_MiL_FQS}) where the ordinary Cartesian space and the unit circle are substituted with the combination of membership functions yielding the fuzzy qualitative coordinate and the fuzzy qualitative unit circle. Extension from this, a fuzzy qualitative representation of the robot kinematics \cite{Liu2008a,Liu2008} was proposed. The work presented a derivative extension to the Fuzzy Qualitative Trigonometry \cite{liu2009fuzzy}. Motivated by these approaches, \cite{chan2008fuzzy} proposed a data quantization process based on the Fuzzy Qualitative Trigonometry to model the uncertainties during the kinematic chain tracking process; and subsequently constructed a generic activity representation model.

\subsubsection{Fuzzy voxel person}

\begin{figure}[h]
\centering
\includegraphics[width=3.5in]{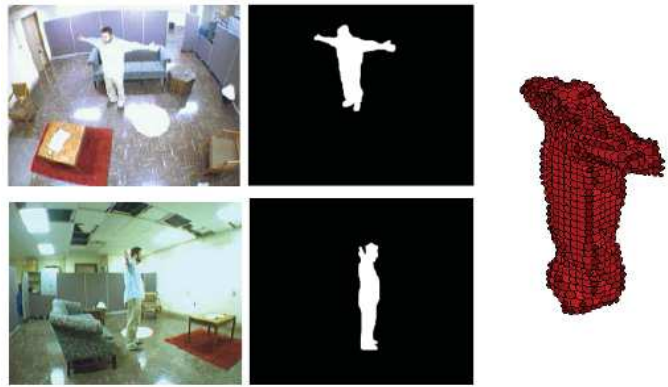}
\caption{Voxel person constructed using multiple cameras from different viewpoints of the silhouette images that resolved the occlusion problem in the single camera system. However, due to the location of the cameras and the person's positions, the information gathered using the crisp voxel person model can be imprecise and inaccurate. Therefore, the fuzzy voxel person representation was proposed \cite{anderson2009fuzzy}.}
\label{fig:crispVoxel}
\end{figure}

As aforementioned, the 3D models provide more useful information than the 2D models as the features (height, centroid, orientation, etc.) in the 3D space are camera-view independent. Inspired by this, \cite{anderson2009modeling,anderson2009linguistic} demonstrated a method to construct a 3D human model in voxel (volume element) space using the human silhouette images called the voxel person (Figure \ref{fig:crispVoxel}). However, due to the location of the cameras and the object's positions, the gathered information using the crisp voxel person model can be sometimes imprecise and inaccurate. The crisp technique works well if and only if there are sufficient number of cameras. But unfortunately, it is hard to find more than a couple of cameras in the same area due to the high cost involved and the limited space area. 

\begin{figure}[h]
\centering
\includegraphics[width=4.5in]{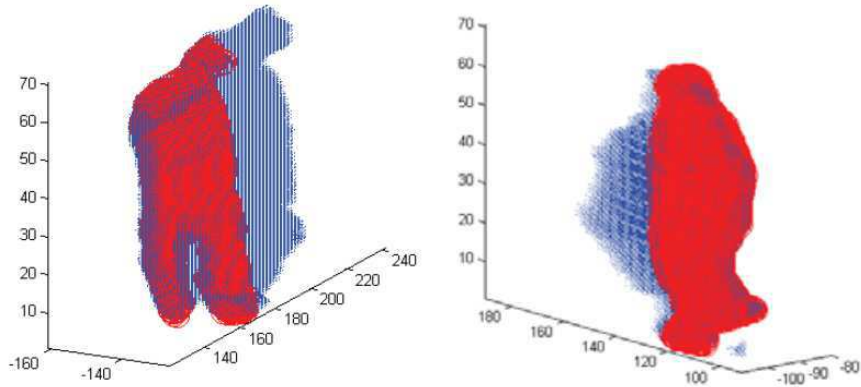}
\caption{The proposed fuzzy voxel person to obtain an improved crisp object. Red areas are the improved voxel person and the blue areas are the rest of the original crisp voxel person \cite{anderson2009fuzzy}. This picture is best viewed in colors.}
\label{fig:FuzzyVoxel}
\end{figure}

Therefore, fuzzy voxel person was utilized in \cite{anderson2009fuzzy} by employing only a few cameras and a minimal prior knowledge about the object. The FIS was used to determine the membership degree of the voxel person, reflecting how likely it belongs to the actual object. Extreme body joints viewing conditions were taken into account and it was observed that the fuzzy acquired results were much better than the crisp approach, both qualitatively (as shown in Figure \ref{fig:FuzzyVoxel}) as well as quantitatively \cite{anderson2009fuzzy}. This concept of the fuzzy voxel person was incorporated in a number of works \cite{anderson2009modeling,anderson2009linguistic}. 

\subsubsection{Fuzzy shape estimation}

The regions of interest extracted from the background subtraction algorithm are normally represented using different shape models, such as ribbons and blobs for 2D images, while cones, cylinders, spheres, ellipses etc. for the 3D images. Here, we will concentrate mainly on the blob representation. For a tracking system with reliance on the shape estimation, problems arise due to imperfect image segmentation techniques. This is because of the image irregularities, shadows, occlusions, etc. that results in multiple blobs generation for a single object. Besides that, in the multiple objects tracking, recovering from the overlapping regions is a big challenge. In order to solve this, \cite{garcia2002robust,garcia2011fuzzy} applied FIS to update both the trajectories and the shape estimated for the targets with a set of image regions. These image regions are represented using the blobs extracted from each frame. Following the general steps of the FIS, heuristic features were extracted from the detected blobs, and used as inputs to the FIS to assess the confidence values assigned to each blob to update the estimators describing the targets' shape and the tracks. With this, the tracking can be locked if the confidence of the target shape is low. This is to prevent the tracking to deviate from the real path caused by the cumulated errors such as the uncertain shape. The tracking resumes once the confidence of the object shape is high.

\subsection{Non-model based tracking}

In non-model based tracking, the objects detected are represented using the random dispersed points instead of the rigid shape models (e.g. stick figure, blob, cylinder, etc.). The association amongst the points that contribute to the motion tracking are based on the hypothesis which takes into account the object's characteristics and behavior. This is a complex problem to be formulated because of the presence of occlusions, misdetections, new object entries etc. that may lead to permanent tracking error. The fuzzy approaches such as the fuzzy Kalman filter, fuzzy particle filter, fuzzy optical flow and fuzzy clustering are widely employed in the non-model based object tracking, where they explicitly take into account the uncertainties to establish the point correspondence between the object motions.

\subsubsection{Fuzzy Kalman filter}

Kalman filter, the popular optimal estimator capable of operating recursively on the streams of noisy input data \cite{kalman1960new}, is a popular choice for tracking a moving object. It has been successfully applied in several previous works on the human motion tracking \cite{kakadiaris1996model,kohler1997using,yun2005implementation,yun2006design,marins2001extended,welch2009history}. There are three basic steps involved in the Kalman filtering for human motion tracking: initialization, prediction and correction \cite{welch1995introduction}. Often the complex dynamic trajectories due to the changes in the acceleration of human motion are not feasible to be modeled by the linear systems. Therefore, instead of the basic Kalman filters, the Extended Kalman filters are used which are capable of modeling the non-linear states. However, all these Kalman filtering algorithms suffer from the divergence problem if the theoretical behavior of a filter and its actual behavior do not agree. The divergence due to modeling errors is a critical issue in the Kalman filtering process. 

In order to solve this, the FIS was adopted in the Kalman filtering \cite{chen1998fuzzy,kobayashi1998accurate,sasiadek1999sensor,sasiadek2000fuzzy,sasiadek2001sensor,senthil2006nonlinear} to detect the bias of measurements and prevent the divergence. The new Kalman filter is called as the fuzzy adaptive Kalman filter. Takagi-Sugeno fuzzy model is used to detect the divergence and the uncertainty of the parameters in the Kalman filter such as the covariance and the mean value are modeled as membership function with the corresponding fuzzy rules for inference. To this extent, \cite{angelov2008autonomous} proposed the evolving Takagi-Sugeno fuzzy model \cite{angelov2004approach,angelov2005simpl_ets} which can be seen as the fuzzy weighted mixture of the Kalman filter for object tracking in the video streams, and the performance is better than the ordinary Kalman Filter.

\subsubsection{Fuzzy particle filter}
\label{MiL_FPF}

Similar to the Kalman filters, the particle filters offer a good way to track the state of a dynamic HMA system. In general, if one has a model of how the system changes with time, and possible observations made in particular states, the particle filters can be employed for tracking. However, as compared to the Kalman filters, the particle filters offer a better tracking mechanism as it provides multiple predictions or hypothesis (i.e. as many as hypothesis as the number of particles) to recover from the lost tracks, which helps to overcome the problems related to the complex human motion. One must note that there is a tradeoff between system precision and computational cost in the particle filter framework, i.e. more number of particles improves the system precision, but also increases the computational cost and vice versa.

As a remedy to the above mentioned problems, a new sequential fuzzy simulation based particle filter was proposed in \cite{wu2008fuzzy} to estimate the state of a dynamic system with noises described as fuzzy variables using the possibility theory. In most of the current particle filtering algorithms, the uncertainty of the tracking process and the measurement of noises are expressed by the probability distributions, which are sometimes hard to construct due to the lack of statistical data. Therefore, it is more suitable to compute the possibility measure using the fuzzy set theory for modeling the uncertain variables with imprecise knowledge. \cite{wu2008fuzzy} found that their proposed fuzzy logic based particle filter outperforms the traditional particle filter even when the number of particles is small. Another variant of this work is \cite{yoon2013object}, where an adaptive model is implemented in the fuzzy particle filter with the capability to adjust the number of particles by using the result from the measurement step, and improve the speed of an object tracking algorithm. Apart from that, \cite{chan2009fuzzy,chan2008fuzzy} handled the tradeoff between the system precision and the computational cost by employing data quantization process that utilizes the Fuzzy Quantity Space \cite{liu2009fuzzy}. In general, the work quantize the particles into finite fuzzy qualitative states. As such, the system able to model the offset of the tracking errors, while retaining the precision when relatively low number of particles are selected to perform the tracking task. Last but not the least, the FIS has also contributed in the particle filters \cite{kamel2005fuzzy,kim2007fuzzy} and achieved better accuracy with lower computational cost.

\subsubsection{Fuzzy optical flow}

Optical flow \cite{horn1981determining,beauchemin1995computation} is another popular motion tracking algorithm. It is an efficient technique for approximating the object motion in two consecutive video frames by computing the intensity variations between them. However, the removal of the incoherent optical flow field is still a great challenge. This is because the incoherent regions can be treated as random noises in the optical flow field due to the sources of disturbances in a natural scene (e.g. dynamic background). Fuzzy hostility index was introduced in \cite{bhattacharyya2009high,bhattacharyya2013target} to overcome this issue and thus improving the time efficiency of the flow computation. The fuzzy hostility index \cite{bhattacharyya2007binary} measures the amount of homogeneity or heterogeneity of the neighborhood pixel in the optical flow field. The more homogeneous is the neighborhood of a pixel, the less is the pixel hostile to its neighbor. This implies that a denser neighborhood indicates a more coherent optical flow neighborhood region. To deal with the uncertain conditions, soft computing is applied where the hostility index computed from the neighborhood pixels is represented as a fuzzy set, where the membership values lie between 0 and 1. This method has shown the capability to track fast moving objects from the video sequences efficiently.

\begin{table}[htb]
	\centering
		\caption{Summarization of research works in MiL HMA using the Fuzzy approaches.}
		\label{Table:MiLSurvey}
		\resizebox{15cm}{!} 
		{
		\begin{tabular}{| l | p{7cm}  c  p{7cm}  c |}\hline		
				
		\multirow{2}{*}{MiL processing} & Problem statements / & \multirow{2}{*}{Papers} & \multirow{2}{*}{Why fuzzy?}  & \multirow{2}{*}{Approach} \\  
		  		
		& Sources of Uncertainty & & & \\ \hline \hline
		  		
		& & & & \\
		
		Model based tracking & Crisp representation of the kinematic chain suffers from the precision problem, and the cumulative errors can directly affect the performance of the tracking process. &  \cite{Liu2008a,chan2009fuzzy,liu2009fuzzy,Liu2008,chan2008fuzzy} & Integration of the fuzzy set theory and the fuzzy qualitative reasoning in the kinematic chain representation provides a means of handling the uncertainty in a natural way. Fuzzy qualitative kinematics solves the precision problem by eliminating the hard boundary problem in the measurement space that can tolerate the offset errors.  & Fuzzy qualitative kinematics\\ 
		 
		& & & & \\		  
		\cline{2-5}	
		& & & & \\
				
		& Due to the location of cameras and object's positions, the information gathered using crisp voxel person model can be imprecise and inaccurate. Crisp approach works fine in multi-camera environment, but it is not feasible due to high cost and limited space. & \cite{anderson2009modeling,anderson2009fuzzy,anderson2009linguistic} & Fuzzy voxel person is able to model different types of uncertainties associated with the construction of the voxel person by using the membership functions, employing only a few cameras and a minimal prior knowledge about the object. & Fuzzy voxel person\\
		
		& & & & \\		  
		\cline{2-5}	
		& & & & \\
			
		& In shape based (blob) tracking, the imperfect image segmentation techniques result in multiple blobs generation for a single object because of the image irregularities, shadows, occlusions, etc. While in the multiple object tracking, recovering from the overlapping regions is a big challenge. & \cite{garcia2002robust,garcia2011fuzzy} & FIS is applied to perform the fuzzy shape estimation to achieve a better tracking performance by taking into account the uncertainty in shape estimation. If the shape is uncertain, the tracking will be locked and it will be recovered once the confidence becomes higher. This is to prevent the tracking errors caused by the uncertain shapes. & Fuzzy shape estimation\\
		
		& & & & \\		
		\hline		
		& & & & \\
				
		Non-model based tracking & Conventional Kalman filter algorithms suffer from the divergence problem and it is difficult to model the complex dynamic trajectories. & \cite{hu2004survey,kim2010intelligent,ko2008survey,aggarwal1997human,gavrila1999visual,chen1998fuzzy,kobayashi1998accurate} & Fuzzy Kalman filters are capable of solving the divergence problem by incorporating the FIS, and are more robust against the streams of random noisy data inputs. & Fuzzy Kalman filter\\ 		
		
		& & & & \\		  
		\cline{2-5}	
		& & & & \\
		
		& Particle filters suffer from the tradeoff between the accuracy and computational cost as its performance usually relies on the number of particles. This means more number of particles will improve the accuracy, but at the same time increases the computational cost. & \cite{chan2009fuzzy,chan2008fuzzy,wu2008fuzzy,yoon2013object,kamel2005fuzzy,kim2007fuzzy} & The fuzzy particle filter effectively handles the system complexity by compromising the low number of particles that were used while retaining the tracking performance. & Fuzzy particle filter\\
		
		& & & & \\		  
		\cline{2-5}	
		& & & & \\
				
		& Random noises in optical flow field due to the sources of disturbances in a natural scene (e.g. dynamic background) affects the tracking performance. & \cite{bhattacharyya2009high,bhattacharyya2013target} & Fuzzy hostility index is used in the optical flow to filter the incoherent optical flow field containing random noises in an efficient manner. & Fuzzy optical flow\\
	
		& & & & \\		  
		\cline{2-5}	
		& & & & \\
				
		& In the conventional methods for multi-object tracking, hard clustering tracking algorithms such as the K-means are used, and involve high complexity and computational cost. Also, they fail in the case of severe occlusions and pervasive disturbances. & \cite{xie2004multi} & FCM tracking algorithm offers more meaningful and stable performance by using soft computing techniques. The integration of component quantization filtering with FCM tracking algorithm provides faster processing speed. & Fuzzy clustering\\		
		
		& & & & \\
		\hline
								
		\end{tabular}
		}
\end{table}

\subsubsection{Fuzzy clustering}

Clustering is an unsupervised machine learning solution that learns the unlabeled data by grouping the similar ones into the corresponding groups autonomously. Inspired from this, multi-object cluster trackings \cite{heisele1997tracking,pece2002cluster} were introduced with the belief that the moving targets always produce a particular cluster of pixels with similar characteristics in the feature space, and the distribution of these clusters changes only little between the consecutive frames. \cite{xie2004multi} proposed a fast fuzzy c-means (FCM) clustering tracking method which offers a solution towards the high complexity and the computational cost involved in the conventional methods on multi-object tracking, and also the hard clustering algorithms such as the k-means that causes failure in the case of severe occlusions and pervasive disturbances. FCM is also recognized as the soft clustering algorithm where it applies data partition to allocate each sample data into more than one clusters with the corresponding membership values which is more meaningful and stable than the hard clustering algorithms. In \cite{xie2004multi}, the component quantization filtering was incorporated with FCM to provide faster processing speed. Table \ref{Table:MiLSurvey} summarizes the intuition of using the fuzzy approaches in MiL HMA.


\section {High-level HMA}
\label{L}

The final aim of the HMA system is to perform human behavior understanding. In this section, we study the feasibility of the fuzzy approaches to achieve this with emphasis on: (a) hand gesture recognition, (b) activity recognition, (c) style invariant action recognition, (d) multi-view action recognition, and (e) anomaly event detection.

\subsection{Hand gesture recognition}
\label{GR}

Gesture recognition aims at recognizing meaningful expressions of the human motion, involving the hands, arms, face, head, or body. The applications of gesture recognition are manifold \cite{lyons1999automatic}, ranging from the sign language to medical rehabilitation and virtual reality. The importance of gesture recognition lies in building efficient and intelligent human-computer interaction applications \cite{wu1999vision} where one can control the system from a distance for a specific task, i.e. without any cursor movements or screen touching. Besides that, nowadays, there exists successful commercialized gesture recognition devices such as the Kinect: a vision-based motion sensing device, capable of inferring the human activities. Unfortunately, in a gesture recognition system, the complex backgrounds, dynamic lighting conditions and sometimes the deformable human limb shapes can lead to high level of uncertainties and ambiguities in recognizing the human gestures. Also, ``pure'' gestures are seldom elicited, as people typically demonstrate ``blends'' of these gestures \cite{mitra2007gesture}. Among all the solutions, the fuzzy clustering algorithms and the integration of fuzzy approaches with machine learning methods are often incorporated to deal with such difficult situations and achieve better system performance. In this section, we review the relevant works with emphasis on the hand gesture recognition.

\subsubsection{Fuzzy clustering}
\label{FCM}

Among the well-known clustering techniques are K-means, GMM, hierarchical model, and FCM. However, in the probabilistic based clustering algorithms (e.g. K-means, GMM, and hierarchical model), the data allocation to each cluster is done in a crisp manner, that is each data element can belong to exactly one cluster. In contrast, the fuzzy clustering algorithm (e.g. FCM), soft computing is applied in the sense that the data partition alleviates the data allocation where each data can belong to more than one clusters and associated with a set of membership values. This solution works better in the challenging environments such as the complex backgrounds, dynamic lighting conditions, and the deformable hand shapes with real-time computational speeds \cite{wachs2002real,wachs2005cluster,li2003gesture,verma2009vision}. 

Using the FCM, \cite{wachs2002real,wachs2005cluster} worked on a fast respond telerobotic gesture-based user interface system. The nature of FCM in relaxing the hard decision allowed the use of smaller portions of the training set and thus shorter training time was required. Empirically, it has proved to be sufficiently reliable and efficient in the recognition tasks with the achievement on high accuracy and real-time performance. \cite{li2003gesture} further improved the work \cite{wachs2002real} in the skin segmentation problem using the color space to solve the skin color variation. Besides spatial information, temporal information is also important in the gesture inference process. In \cite{verma2009vision}, the spatial information of hand gesture using the FCM was trained in order to determine the partitioning of the trajectory points into a number of clusters with the fuzzy pseudo-boundaries. In general, each trajectory point belongs to each cluster specified by a membership degree. Then, the temporal data is obtained through the transitions between the states (cluster of trajectory points) of a series of finite state machines to recognize the gesture motion.

\subsubsection{Hybrid technique}

A few works \cite{al2001recognition,binh2005hand,varkonyi2011human} on fusing the fuzzy approaches with machine learning solutions have been reported in the gesture recognition. \cite{al2001recognition} used the adaptive neuro-fuzzy inference system to recognize the gestures in Arabic sign language. This work was motivated by the transformation of human knowledge into a FIS, but does not produce the exact desired response due to the heuristic or non-sophisticated membership functions and the fuzzy rules generation. Thus, there was a need to fine-tune the parameters in the FIS to enhance its performance, and the adaptive neuro-fuzzy inference system provided this flexibility by applying a learning procedure using a set of training data.

\cite{binh2005hand} introduced a new approach towards gesture recognition based on the idea of incorporating the fuzzy ARTMAP \cite{carpenter1992fuzzy} in the feature recognition neural network \cite{hussain1994novel}. The proposed method reduced the system complexity and performed in real-time manner. Nonetheless, \cite{varkonyi2011human} presented an approach with several novelties and advantages as compared to other hybrid solutions. They introduced a new fuzzy hand-posture model using a modified circular fuzzy neural network architecture to efficiently recognize the hand posture. As a result, the robustness and reliability of the hand-gesture identification was improved, and the complexity and training time involved in the neural networks was significantly reduced.

\subsection{Activity recognition}
\label{AR}

Activity recognition is an important task in the HiL HMA systems. The goal of activity recognition is to autonomously analyze and interpret the ongoing human activities and their context from the video data. For example, in the surveillance systems for detecting suspicious actions, or in sports analysis for monitoring the correctness of the athletes' postures. In recent times, the fuzzy approaches such as type-1 FIS, fuzzy HMM, and hybrid techniques have proved to be beneficial in the human activity recognition, with capability of modeling the uncertainty in the feature data. Nonetheless, Fuzzy Vector Quantization (FVQ) and Qualitative Normalized Template (QNT) provide the capability to handle the complex human activities occurring in our daily life such as walking followed by running, then running followed by jumping, or a hugging activity where two or more people are involved. In this section, we will discuss on the applications of these fuzzy approaches in the activity recognition.

\subsubsection{Type-1 fuzzy inference system}
\label{HiL_FIS1}

The FIS can be efficiently used to distinguish the human motion patterns and recognize the human activities with its capability of modeling the uncertainty and the fusion of different features in the classification process. In the literature of activity recognition, there exists some works \cite{le2012fuzzy,yao2014fuzzy} that employed the FIS to classify different human activities.

Both \cite{le2012fuzzy,yao2014fuzzy} took into account the uncertainties in both the spatial and temporal features for efficient human behavior recognition. Their method aims at handling high uncertainty levels and the complexities occurring in the real world applications. \cite{le2012fuzzy} used the spatial and temporal geometry features to study the importance of the spatio-temporal relations such as `\textit{IsMoving}', `\textit{IsComingCloseTo}', `\textit{IsGoingAway}', `\textit{IsGoingAlong}' with the objective to provide a qualitative interpretation of the behavior of an entity (e.g. a human) in real-time. Another work \cite{yao2014fuzzy} adopted the spatio-temporal features such as the silhouette slices and the movement speed in video sequences as the inputs to the FIS. Extra merit in this work is that they learn the membership functions of the FIS using the FCM which prevents the intervention of human in generating the fuzzy membership function heuristically.

\subsubsection{Hybrid technique}
\label{HiL_hyb}

Owing to the demands of the development of enhanced video surveillance systems that can automatically understand the human behaviors and identify dangerous activities, \cite{acampora2012combining} introduced a semantic human behavioral analysis system based on the hybridization of the neuro-fuzzy approach. In their method, the kinematic data obtained from the tracking algorithm is translated into several semantic labels that characterizes the behaviors of various actors in a scene. To achieve this, the behavioral semantic rules were defined using the theory of time delay neural networks and the fuzzy logic, to identify a human behavior analyzing both the temporal and the contextual features. This means that they analyze how a human activity changes with respect to time along with how it is related to the contexts surrounding the human. Their hybrid method outperformed other approaches and showed high level of scalability and robustness.

Another work \cite{hosseini2013fuzzy} presented a fuzzy rule-based reasoning approach for event detection and annotation of broadcast soccer video, integrating the Decision Tree and the FIS. A flexible system was designed using the fuzzy rules, that can be used with least reliance on the predefined feature sequences and domain knowledge. The FIS was designed as a classifier taking into account the information from a set of audio-visual features as its crisp inputs and generate the semantic concepts corresponding to the events occurred. From the fuzzification of the feature vectors derived from the training data, a set of tuples were created, and using the Decision Tree, the hidden knowledge among these tuples as well as the correlation between the features and the related events were extracted. Then, traversing each path from the root to the leaf nodes of the Decision Tree, a set of fuzzy rules were generated which were inserted in the knowledge base of the FIS and the occurred events were predicted from the input video (i.e. soccer video) with good accuracy.

\subsubsection{Fuzzy vector quantization}
\label{HiL_CAR}

\begin{figure}[htbp]
\centering
\includegraphics[scale=0.5]{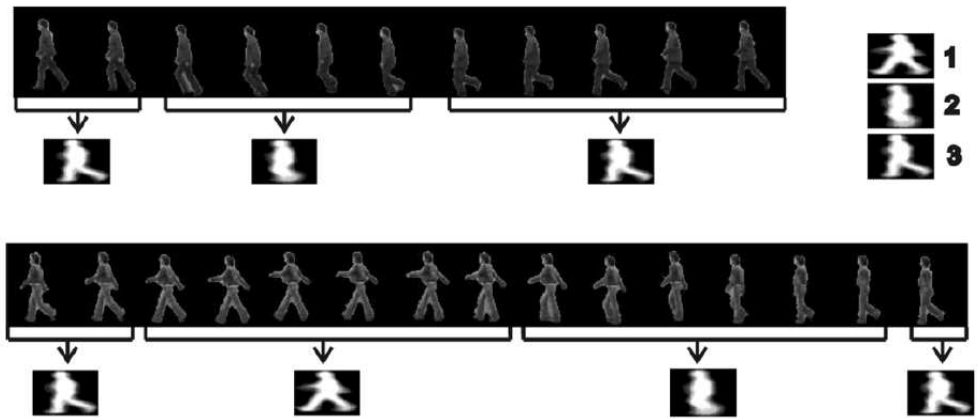}
\caption{Movements of running (top) and walking (bottom) activities, as well as the associated dynemes which are learned from the FCM \cite{gkalelis2008combining}.}
\label{fig:Fig_HiL_FVQ}
\end{figure}

In order to learn the complex actions, \cite{gkalelis2008combining} represented the human movements as a combination of the smallest constructive unit of human motion patterns called the dyneme (Figure \ref{fig:Fig_HiL_FVQ}). It is the basic movement patterns of a continuous action. In the bottom of action hierarchy, dyneme is defined as the smallest constructive unit of human motion; while one level above is the movement which is perceived as a sequence of dynemes with clearly defined temporal boundaries and conceptual meaning. Dyneme can be learned in an unsupervised manner and in \cite{gkalelis2008combining}, the FCM was chosen. Then, fuzzy vector quantization (FVQ) \cite{karayiannis1995fuzzy} as a function that regulates the transition between the crisp and the soft decisions was employed to map an input posture vector into the dyneme space. Finally, each movement was represented as a fuzzy motion model by computing the arithmetic mean of the comprising postures of a movement in the dyneme space. Their algorithm provides good classification rates and exhibits adequate robustness against partial occlusions, different styles of movement execution, viewpoint changes, gentle clothing conditions and other challenging factors.

\begin{figure}[tb]
\centering
\includegraphics[scale=0.3]{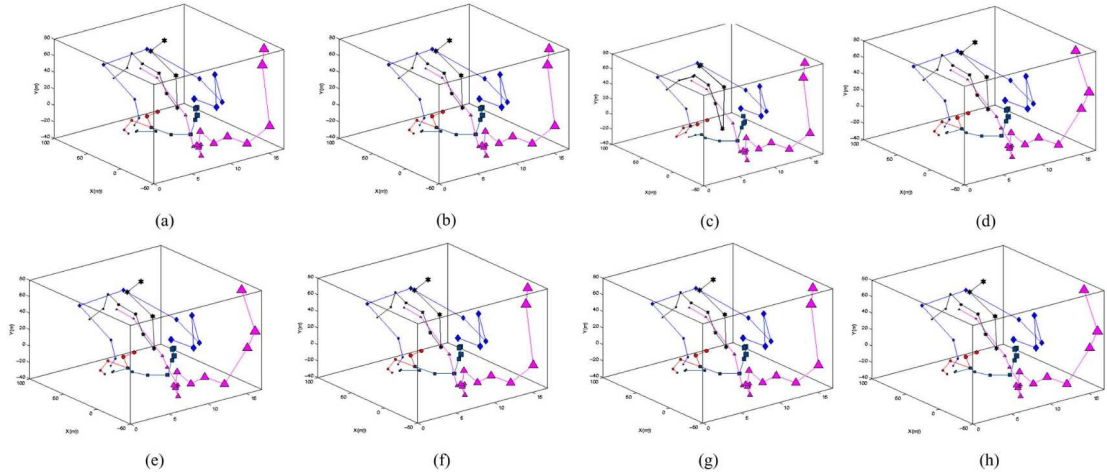}
\caption{Visualization of the QNT model: each of the five activities (walking, running, jogging, one-hand waving (wave1) and two-hands waving(wave2)) from eight subjects (a)-(h) in the quantity space \cite{chan2009fuzzy}.}
\label{fig:Fig_HiL_QNT}
\end{figure}

\subsubsection{Qualitative normalized template}

Utilizing the concept of fuzzy qualitative robot kinematics \cite{Liu2008a,Liu2008}, Chan and Liu \cite{chan2009fuzzy,chan2008fuzzy} built a generative action template, called the Qualitative normalized template (QNT) to perform the human action recognition. First of all, the training data that represents a typical activity is acquired by tracking the human anatomical landmarks in the image sequences. In their work, a data quantization process was employed to handle the tradeoffs between the tracking precision and the computational cost. Then, the QNT as illustrated in Figure \ref{fig:Fig_HiL_QNT} was constructed according to the fuzzy qualitative robot kinematics framework \cite{Liu2008a,Liu2008}. An empirical comparison with the conventional hidden Markov model (HMM) and fuzzy HMM using both the KTH and the Weizmannn datasets has shown the effectiveness of the proposed solution \cite{chan2009fuzzy}.

\subsubsection{Fuzzy Hidden Markov Model}

Hidden Markov model (HMM) \cite{elliott1995hidden} is the statistical Markov model with the state being not directly visible, but the output that is dependent on the state is visible. HMM  have been widely employed in the human action recognition \cite{bobick1995state,campbell1995recognition,oliver2000bayesian,wilson1999parametric,yamato1992recognizing}. These papers have well demonstrated the modeling and recognition of the complex human activities using HMM. In the training stage of HMM, expectation maximization algorithm is adopted. However, in the conventional HMM, each observation vector is assigned only to one cluster. \cite{mozafari2012novel} pointed out that assigning different observation vectors to the same cluster is possible and if their observation probabilities become the same, consequently, the classification performance may decrease. Therefore, HMM was extended to fuzzy HMM where in the training stage, the distance from each observation vector to each cluster center is computed and the inverse of the distance is considered as the membership degree of the observation vector to the cluster. \cite{mozafari2012novel} utilized this concept for human action recognition and the experiment results demonstrate the effectiveness of the fuzzy HMM in human action recognition, with good recognition accuracy for the similar actions such as ``walk'' and ``run''.

\subsection{Style invariant action recognition}
\label{Bio}

A robust action recognition algorithm must be capable of recognizing the actions performed by different person in different styles. Commonly, different person have different styles of executing the same action which can be categorized according to the physical differences (such as human appearances, sizes, postures, etc.) and the dynamic differences (speed, motion pattern, etc.) \cite{lim2013fuzz}. In order to model such variations, several notable works have been reported incorporating the fuzzy approaches.

\subsubsection{Fuzzy vector quantization}

\cite{iosifidis2011person} adopted the concept of FVQ and the dyneme, and proposed a novel person specific activity recognition framework to cope with the style invariant problem. The method is mainly divided into two parts: firstly, the ID of the person is identified, and secondly, the activity is inferred from the person specific fuzzy motion model \cite{gkalelis2008combining}. It was found that the different styles in action execution endowed the capability to distinguish one person from the another. Therefore, \cite{iosifidis2012activity} developed an activity-related biometric authentication system by utilizing the information of different styles by different people. Improvement was made in the computation of the cumulative fuzzy distances between the vectors and the dynemes that outperforms $L_1$, $L_2$, and Mahalanobis distances which were used previously in \cite{gkalelis2008combining}.

\subsubsection{Fuzzy descriptor action model}

There is a limitation in \cite{iosifidis2011person,iosifidis2012activity} where a large storage space is required to store the ID of different person and this makes the system impractical. An alternative approach was proposed in \cite{lim2013fuzz}, where a fuzzy descriptor vector was used to represent the human actions of different styles in a single underlying fuzzy action descriptor. Theoretically, the ordinary descriptor vector was allowed to contain only a single value in each dimension of the vector. In contrast, fuzzy descriptor allows accommodation of a set of possible values where these values hold the different measurements of the feature data obtained from the training data, comprising of an action performed by different person in different styles.

\subsection{Multi-view action recognition}
\label{View}

The capability of multi-view action recognition is emerging as an important aspect for advanced HMA systems. In the real world environment, human are free to perform an action at any angle with no restriction of being frontal parallel to the camera and most of the previous works treat it as a constraint or limitation in their system. This problem has received increasing attention in the HMA research and some of the notable works have been reported \cite{ji2010advances,weinland2006free,lewandowski2010view}. Besides that, fuzzy approaches such as the FVQ, and fuzzy qualitative reasoning are also applied in the study of multi-view action recognition which will be discussed in the following subsections. 

\subsubsection{Fuzzy vector quantization}

\begin{figure}[htb]
\centering
\includegraphics[scale=0.4]{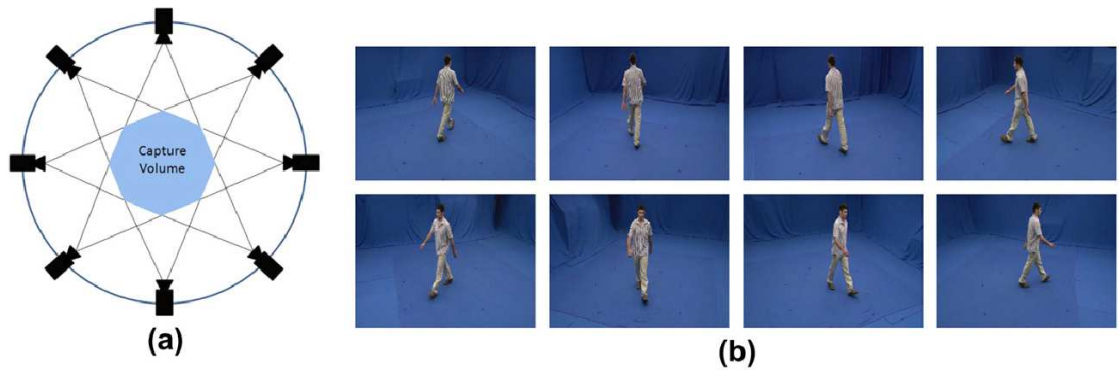}
\caption{(a) A converging eight-view camera setup and its capture volume, and (b) an eight-view video frame \cite{Iosifidis2012}.}
\label{fig:Fig_HiL_Multiview}
\end{figure}

\cite{iosifidis2012activity,Iosifidis2012,iosifidis2013minimum} extended \cite{gkalelis2008combining} to support multi-view action recognition. The motion patterns obtained from different cameras, as in Figure \ref{fig:Fig_HiL_Multiview}, were clustered to determine the number of multi-view posture primitives called the multi-view dynemes. Similar to \cite{gkalelis2008combining}, FVQ was utilized to map every multi-view posture pattern to create the multi-view dyneme space. This new multi-view fuzzy movement representation is motion speed and duration invariant which generalizes over variations within one class and distinguishes between the actions of different classes. In the recognition step, Fourier view invariant posture representation was used to solve the camera viewpoint identification problem before the action classification was performed. Nonetheless, they tackled the problem of interaction recognition i.e. human action recognition involving two persons \cite{Iosifidis2012b}.

\subsubsection{Fuzzy qualitative single camera framework}

\begin{figure}[htb]
\centering
\includegraphics[scale=0.3]{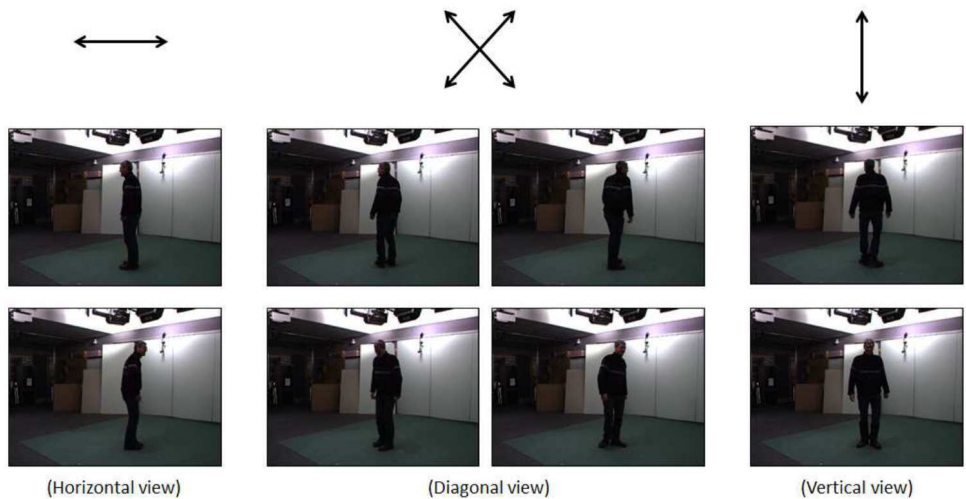}
\caption{Predefined viewpoints from left to right: `horizontal view', `diagonal view' and `vertical view' \cite{lim2013fuzz}.}
\label{fig:Fig_HiL_Viewpoint}
\end{figure}

In most of the multi-view action recognition works, there is an argument that performing view invariant human action recognition using multi-camera approach is not practical in real environment \cite{holte2011human,lewandowski2010view}. The reasons are: firstly, such systems must be deployed in a close environment that has many overlapping regions which is very rare in the open public space. Secondly, the implementation and the maintenance cost is very high due to the usage of multiple cameras. In order to solve this, \cite{lim2013fuzz} proposed a fuzzy action recognition framework for multi-view within a single camera. Their work introduced the concept of learning the action in three predefined viewpoints which are horizontal, diagonal, and verticle view, as depicted in Figure \ref{fig:Fig_HiL_Viewpoint}. The learning is done by mapping the features extracted from the human silhouette onto the fuzzy quantity space. The dominant features are then identified from the fuzzy qualitative states and represented as a fuzzy descriptor \cite{lim2013fuzz}. In the action recognition step, the viewpoint of the person is first estimated and then proceeded with the action recognition by utilizing the viewpoint specific fuzzy descriptor action models \cite{lim2013fuzz}.

\subsection{Anomaly event detection}
\label{AB}

Anomaly detection refers to the problem of finding patterns in the input data that do not conform to the expected behavior. In our daily life, anomaly detection is important to infer the abnormal behavior of a person, such as an action or an activity that is not following the routine or deviated from the normal behavior \cite{hu2004survey,kratz2009anomaly,wu2010chaotic}. For example, in the healthcare domain to prevent unfavorable events from occurring such as the risk of falling down of the patients, and in the surveillance systems, to automatically detect the crime activities.

\subsubsection{Type-1 fuzzy inference system}

As humans gain more knowledge, they are able to make better decisions; similarly if the FIS is provided with sophisticated knowledge (i.e. fuzzy rules), it can deal with the real world problems in a better manner. FIS has been employed in various works for anomaly event detection such as the elderly fall detection in \cite{anderson2006recognizing,anderson2009modeling,anderson2009linguistic}, to address the deficiencies and the inherent uncertainty related to modeling and inferring the human activities. The works emphasized that the non-interpretable likelihood value or the ad-hoc training of the activity models in the conventional approaches is impractical in the area of human action recognition. Therefore, a confidence value (fuzzy membership degree) that can be reliably used to reject unknown activities is more convenient. 

\begin{figure}[htbp]
\centering
\includegraphics[scale=0.4]{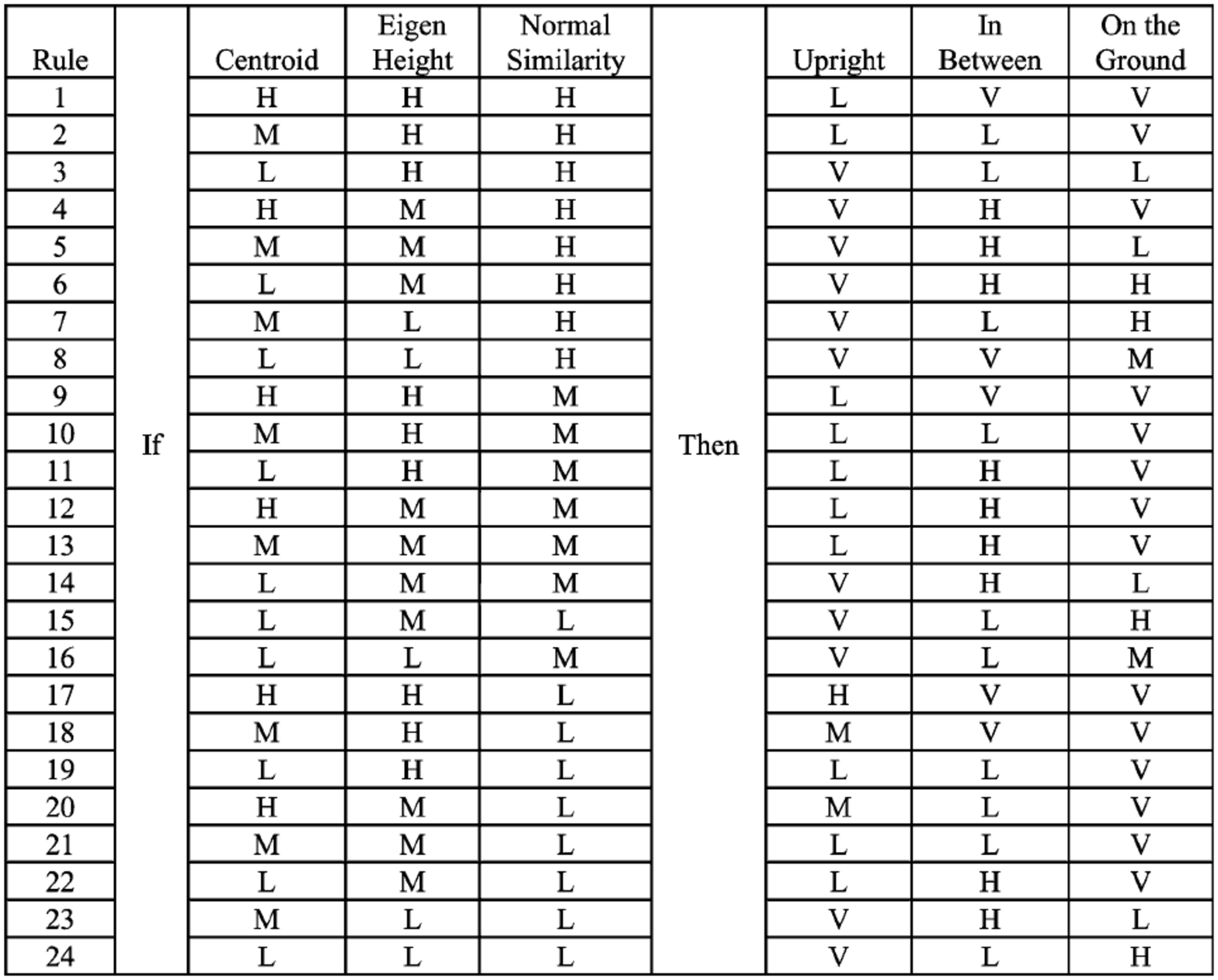}
\caption{Rule table of the human states (Upright, In Between, On the Ground) with V=Very low, L=Low, M=Medium, and H=high which are used to infer the human activities \cite{anderson2009modeling}.}
\label{fig:Fig_HiL_RuleTable}
\end{figure}

\cite{anderson2009modeling} proposed a novel fuzzy rule based method for monitoring the wellness of the elderly people from the video. In this paper, the knowledge base (fuzzy rules as depicted in Figure \ref{fig:Fig_HiL_RuleTable}) was designed under the supervision of nurses for the recognition of falls of the elderly people. Under this framework, the rules can be easily modified, added or deleted, based on the knowledge about the cognitive and functional abilities of the patients. This work was an extension of \cite{anderson2009linguistic} where the linguistic summarizations of the human states (three states: upright, on-the-ground and in-between) based on the voxel person and the FIS were extracted, extended using a hierarchy of the FIS and the linguistic summarization for the inference of the patients' activities. Their technique works well for fall detection, but the question is if this framework can be extended to different activities. The answer is yes where, \cite{anderson2008extension} extended the work to support the additional common elderly activities such as standing, walking, motionless-on-the-chair, and lying-motionless-on-the-couch, with the inclusion of the knowledge about the real world for the identification of the voxels that corresponds to the wall, floor, ceiling, or other static objects or surfaces. Two new states were included to recognize these activities i.e. on-the-chair and on-the couch. These states were different from the previous three states (upright, on-the-ground and in-between) as they were based on the voxel person interacting with a static object in the scene. Further, the fuzzy rules were extended to six new fuzzy rules designed for identifying on-the-chair and on-the-couch activities.

\subsubsection{Fuzzy one class support vector machine}

The fuzzy one class support vector machine (FOCSVM) is an efficient algorithm often used in fall detection systems to distinguish a falling from other activities such as walking, bending, sitting or lying. \cite{yu2011fall} proposed a robust fall detection system using FOCSVM with novel 3D features. In their method, a voxel person was first computed, then the video features obtained from the variation of a persons' 3D angle and centroid information were extracted from the sequences of voxel persons which were used to train the FOCSVM classifier. As compared to the traditional one class support vector machine, FOCSVM obtained more accurate fall detection result with tight decision boundaries under a training dataset with outliers. The success of the proposed method is evident from the experiments on the real video sequences, with less non-fall samples being misclassified as falls by the classifier with imperfect training data.

\subsubsection{Fuzzy clustering}

In order to perform fall detection in multiple camera framework, fuzzy clustering algorithms (e.g. FCM, Gustafson and Kessel Clustering, or Gath and Geva Clustering) along with the fuzzy K-nearest neighbor algorithms were employed in \cite{wongkhuenkaew2013multi}. In particular, Hu moment invariant features were computed from the 2D silhouette images and principal component analysis was utilized to select the principal components. The fuzzy clustering algorithms were used to generate the multi-prototype that represent the action classes such as standing or walking, sitting or bending, lying and lying forward. Fuzzy K-nearest neighbor was then used to deduce the corresponding action classes. For example, if the detected action was ``lying'' or ``lying forward'', it was considered as the falling activity. 

\begin{table}[htbp]
	\centering
		\caption{Summarization of research works in HiL HMA using the Fuzzy approaches.}
		\label{Table:HiLSurvey}
		\resizebox{15cm}{!} 
		{
		\begin{tabular}{| l | p{7cm}  c  p{7cm}  c |}\hline		
				
		\multirow{2}{*}{HiL processing} & Problem statements / & \multirow{2}{*}{Papers} & \multirow{2}{*}{Why fuzzy?}  & \multirow{2}{*}{Approach} \\  
		  		
		& Sources of Uncertainty & & & \\ \hline \hline
		  		
		& & & & \\
		
		Hand gesture recognition & Complex backgrounds, dynamic lighting conditions and sometimes deformable human limbs' shape leads to ineffective clustering outcome with the conventional crisp clustering algorithms. &  \cite{wachs2002real,wachs2005cluster,li2003gesture,verma2009vision} & FCM relaxes the learning and recognition of gesture by using soft computing technique. This reduces the errors caused by the crisp decisions and increases the system efficiency. & Fuzzy clustering\\ 
				
		& & & & \\		
		\cline{2-5}
		& & & & \\	
			
		& Difficulty in determining the optimum parameters in the fuzzy system such as membership function or the threshold value for the decision making in gesture recognition algorithms. &\cite{al2001recognition,binh2005hand,varkonyi2011human} & Integration of the fuzzy approaches with machine learning algorithms help in learning the important parameters for the fuzzy system adaptively based on the training data. & Hybrid technique\\

		& & & & \\		
		\hline		
		& & & & \\
				
		Activity recognition & The uncertainty in the feature data affects the performance of human activity recognition. & \cite{le2012fuzzy,yao2014fuzzy} & FIS effectively distinguishes the human motion patterns and activity recognition with its flexibility in customizing the membership functions and the fuzzy rules with tolerance to the vague feature data. & Type-1 FIS\\

		& & & & \\		  
		\cline{2-5}	
		& & & & \\	
		
		& Difficult to determine the optimum membership functions and the fuzzy rules in the FIS for human activity recognition. & \cite{acampora2012combining,hosseini2013fuzzy} & Integration of fuzzy logic with machine learning techniques allows the generation of the optimum membership function and fuzzy rules to infer the human behavior. & Hybrid technique\\ 
						
		& & & & \\		  
		\cline{2-5}	
		& & & & \\	
		
		& Solving continuous human movements or complex activities over time is a difficult problem. For instance, walk then run. Most of the state-of-the-art methods assumed the activity to be uniform and simple. & \cite{gkalelis2008combining} & FVQ incorporated with FCM is used to model the human movements and provides the flexibility to support complex continuous actions. & FVQ\\

		& & & & \\		  
		\cline{2-5}	
		& & & & \\
		
		& The usage of sophisticated tracking algorithms in the action recognition suffers from the tradeoff between the computational cost and accuracy. & \cite{chan2009fuzzy,chan2008fuzzy,chan2010fuzzy} & QNT fuzzy motion template relaxes the complexity of the representation of the human joints that uses sophisticated tracking algorithms, achieving the efficiency and robustness in complex activity recognition. & QNT\\
		
		& & & & \\		  
		\cline{2-5}	
		& & & & \\
				
		& Conventional HMM is unable to model the uncertainties in the training stage which reduces the classification performance. & \cite{mozafari2012novel} & Fuzzy HMM models apply soft computing in the training stage which effectively increases the performance in the classification of similar actions such as ``walk'' and ``run''. & Fuzzy HMM\\
		
		& & & & \\
		\hline
		& & & & \\
		
		Style invariant action recognition & A similar action can be performed with different styles by different person that causes difficulty in the learning and recognition process. & \cite{iosifidis2011person,iosifidis2012activity} & Style invariant action recognition can be achieved by using person specific fuzzy movement model which is trained using FVQ. & FVQ\\ 		
				
		& & & & \\		  
		\cline{2-5}	
		& & & & \\
		
		& Ordinary descriptor vector that can contain only a single value in each vector dimension, limits the capability to model the actions from different styles. & \cite{lim2013fuzz} & Fuzzy descriptor vector allows to accommodate a set of possible descriptor values in each vector dimension which is able to model the different styles of an action in a single underlying fuzzy action descriptor. & Fuzzy descriptor action model\\
		
		& & & & \\
		\hline
		& & & & \\
		
		Multi-view action recognition & Humans are not restricted to perform an action at a fixed angle from the camera. & \cite{iosifidis2012activity,Iosifidis2012,iosifidis2013minimum,Iosifidis2012b} & Multi-view posture patterns are generated by utilizing FVQ to build a multi-view fuzzy motion model in order to support view invariant human action recognition. & FVQ\\ 			
				
		& & & & \\		  
		\cline{2-5}	
		& & & & \\
						
		& Most of the multi-view action recognition works assume that performing view invariant human action recognition using multi-camera approach is not practical in real environment. & \cite{lim2013fuzz} & Using fuzzy qualitative framework, action recognition is performed for multiple views within single camera in an efficient and robust manner. & Fuzzy qualitative single camera framework\\ 
				
		& & & & \\
		\hline
		& & & & \\	
				
		Anomaly event detection & The difficulty of extension of a framework to deal with new issues and support new activities.  & \cite{anderson2006recognizing,anderson2009modeling,anderson2009linguistic,anderson2008extension} & FIS is flexible in customization where the knowledge base (fuzzy rules) can be modified, added, or removed to adapt to various situations such as falling activities. & Type-1 FIS\\

		& & & & \\		  
		\cline{2-5}	
		& & & & \\
		
		& The imperfect training data (e.g. some samples would be outliers) affect the classification performance in the fall detection system. & \cite{yu2011fall} & FOCSVM is used to reflect the importance of every training sample, by assigning each training data with the membership degree. With this, a good accuracy and decision boundaries are obtained under a training dataset with outliers. & FOCSVM\\
		
		& & & & \\		  
		\cline{2-5}	
		& & & & \\
				
		& Most of the existing elderly fall detection systems are performed in the single camera environment which provides limited information for the inference process. & \cite{wongkhuenkaew2013multi} & Fuzzy clustering algorithms (e.g. FCM, Gustafson and Kessel Clustering, or Gath and Geva Clustering) incorporated with Hu moment invariant features and principle component analysis were employed to learn the multi-prototype action classes in the multiple camera environment. & Fuzzy clustering\\ 
		
		& & & & \\		  
		\cline{2-5}	
		& & & & \\
		
		& Difficulty in determining the optimum parameters in the fuzzy system. & \cite{wang2008detecting,juang2007human,hu2004learning} & Integration of the fuzzy approaches with machine learning algorithms allows the learning of optimum fuzzy membership functions and fuzzy rules that can adapt to newly encountered problems. & Hybrid technique\\
		
		& & & & \\
		\hline
								
		\end{tabular}
		}
\end{table}

\subsubsection{Hybrid technique}

A hybrid model of the FIS and the Fuzzy Associative Memory (FAM) was incorporated in \cite{wang2008detecting}, which basically receives an input and assigns a degree of belongingness to a set of rules. \cite{wang2008detecting} considered the angles of human limbs as the inputs to the FAM with three rules defining the abnormal movement types. FAM then assigns a degree of membership to each rule and determines the anomalous or normal events based on a specific threshold. \cite{juang2007human} also used the neural fuzzy network hybrid model, compensating the lacking of the learning ability of the fuzzy approaches to recognize human poses (e.g. standing, bending, sitting, and lying). Their system with simple fuzzy rules is capable of detecting the emergencies caused by the accidental falls or when a person remains in the lying posture for a period of time. The works evidently show the flexibility of the fuzzy approaches in the alteration or extension of its knowledge base to adapt to newly encountered real world problems. 

Another paper \cite{hu2004learning} proposed fuzzy self-organizing neural network (fuzzy SOM) to learn the activity patterns for anomaly detection in visual surveillance. Their method aims at automatically constructing the activity patterns by self-organizing learning instead of predefining them manually. Traditionally, individual flow vectors were used as inputs to the neural networks. In the proposed method, whole trajectory was taken as an input, simplifying the structure of the neural networks to a great extent. Fuzzy SOM further improved the learning speed and accuracy of the anomaly detection problem, as demonstrated with the support of experimental results. To understand better, a summary of research works in HiL HMA using the fuzzy approaches is shown in Table \ref{Table:HiLSurvey}.

\section{Discussion}
\label{dis}

After reviewing a number of works using the fuzzy approaches in HMA, we identified some important factors that make fuzzy approaches successful in improving the overall system performance and these will be discussed in this section along with the potential future works.

\subsection{Soft boundary}

Human reasoning is a mysterious phenomenon that scientists trying to simulate with machines in the past few decades. With the knowledge that ``soft'' boundaries exist in concepts formation of human beings \cite{Zadeh1997111}, fuzzy set theory has emerged to become one of the most important methodology in capturing notions. In general, fuzzy approach assigns a ``soft'' boundaries, or in other words performing ``soft labeling'' where one subject can be associated with many possible classes with a certain degree of confidence. As such, the fuzzy representation is more beneficial than the ordinary (crisp) representations, as it can represent not only the information stated by a well-determined real interval, but also the knowledge embedded in the soft boundaries of the interval. Thus, it removes, or largely weakens the boundary interpretation problem achieved through the description of a gradual rather than an abrupt change in the degree of membership, closer to how humans make decisions and interpret things in the real world.

This is also supported by a few notable literatures. For example, \cite{bezdek1992computing} in their review on computing with uncertainties emphasized on the fact that the integration of fuzzy models always improves the computer performance in pattern recognition problems. Similarly, \cite{huntsberger1986representation,yager2002uncertainty} presented a survey on how to effectively represent the uncertainty using the FIS. Nevertheless, there are a few studies on the type-2 FIS that have been reported in this regards. \cite{wu2002uncertainty,wu2007uncertainty} explained on how to design an interval type-2 FIS using the uncertainty bounds and introduced the measurement of uncertainty for interval type-2 fuzzy sets using the information such as centroid, cardinality, fuzziness, variance and skewness. A comprehensive review on handling the uncertainty in pattern recognition using the type-2 fuzzy approach was provided by \cite{zeng2006type}.

\subsection{Linguistic support}
Another worth highlighting aspect of human behavior is the way they interpret things in the natural scenarios. Human beings mostly employ words in reasoning, arriving at conclusions expressed as words from premises in natural language or having the form of mental perceptions. As used by humans, words have fuzzy denotations. Therefore, modeling the uncertainties in a natural format for humans (i.e. linguistic summarizations) can yield more succinct description of human activities. Inspired from this, HMA can be modeled efficiently by representing an activity in linguistic terms. This concept was initiated in \cite{zadeh1996fuzzy} where words can be used in place of numbers for computing and reasoning (like done by humans), commonly known as computing with words (CWW).

In CWW, a word is viewed as a fuzzy set of points drawn together by similarity, with the fuzzy set playing the role of a fuzzy constraint on a variable. There are two major imperatives for CWW \cite{zadeh1996fuzzy}. Firstly, CWW is a necessity when the available information is too imprecise to justify by the use of numbers. Secondly, when there is a tolerance for imprecision which can be exploited to achieve tractability, robustness, low solution cost, and better rapport with reality. This concept of using CWW i.e. linguistic support to represent the measurement boundaries can be well applied in the real world scenarios. For example, consider the human activities: walking and running, which can be inferred using simple cue i.e. the speed of a person. Different levels of running speeds of a person can be modeled using the linguistic terms such as `very slow', `slow', `moderate', `fast', and `very fast', instead of representing speed in numerical terms. The use of linguistic terms provide the capability to perform human like reasoning such as the feasibility of defining rules for the inference process. With the integration of the linguistic support in the FIS, the computational complexity of the numeric labeling and the imprecision problem in the interpretation stage are also suppressed. Furthermore, the linguistic terms are more understandable where it mimics how humans interpret things and make decisions. 

The concept of linguistic support is rooted in several papers starting with \cite{zadeh1973outline} in which the concepts of a linguistic variable and the granulation were introduced. \cite{zadeh1996fuzzy} threw light on the role played by the fuzzy logic in CWW and vice-versa. An interesting piece of work on CWW can be found in \cite{rubin1999computing} where the author defined CWW as a symbolic generalization of the fuzzy logic. In the recent years, several papers have been published that utilizes the concept of linguistic summarization in the fuzzy system that are successfully applied in the real world applications \cite{anderson2009linguistic,trivino2008linguistic,kacprzyk2001linguistic,anderson2011linguistic,wilbik2011linguistic,wilbik2013fuzzy}. In these works, a complete sentence instead of numerical data or a crisp answer in a conventional decision making systems is preferable as an output; for example, ``the resident has fallen in the living room and is down for a long time''. Such succinct linguistic summarization output is more understandable and closest to the natural answer.

\subsection{Flexibility of the fuzzy system}

Another advantage of the fuzzy approaches, especially those that utilized the knowledge-based system (fuzzy rules) such as the FIS, is that they possess the flexibility and feasibility to adapt to various system designs. The conventional approaches designed their algorithms to be well-fitted to solve solely some specific problems with low or no extendibility. The world is changing rapidly with the headway of technologies, the flexibility to adapt to such changes is one of the major concerns for a good and long lasting system. Fortunately, the fuzzy approaches allow the alterations to serve the purpose. In addition, the alterations can be made easily on the knowledge base by designing the fuzzy rules.

The knowledge base that comprises of all the rules is considered as the most crucial part of a decision making system where it functions as the ``brain'' of the overall system. As human growth together with knowledge is capable of making better decisions, similarly if a decision making system is provided with sophisticated knowledge, it can deal with the problems in a better manner. The FIS consists of a knowledge base where it can store a number of conditional ``IF-THEN'' rules that are used for the reasoning process in a specific problem domain. These rules are easy to write and as many rules as necessary can be supplied to describe the problem adequately. For example, consider the problem of identifying different human activities e.g. running. Rules can be designed to infer the running activity using a simple cue (speed) as following:

Rule 1: IF (speed is FAST) THEN (person is RUNNING)
		
Rule 2: IF (speed is MODERATE) THEN (person is NOT RUNNING)

However, in the real world scenarios, various factors can affect the speed of a person such as the height, body size, etc. Therefore, in order to make the system closer to natural solution, these rules are needed to be modified accordingly. Intuitively, if one may observe the styles of running of a tall person and a shorter person, due to difference in the step size of their feet, the taller person tends to run with a faster speed due to larger step size as compared to the shorter person, running with moderate speed. However, both are performing the running activity, but with different rules. This situation can be modeled by modifying the ``Rule 2'' as follows:

Rule 2.1: IF (HEIGHT is TALL) \& (SPEED is MODERATE) THEN (person is NOT RUNNING)	
		
Rule 2.2: IF (HEIGHT is SHORT) \& (SPEED is MODERATE) THEN (person is RUNNING)

\noindent Similarly, the body size can also affect the speed of a person, and can be modeled using flexible fuzzy rules that can be easily added, modified or deleted according to the objective of the system.

In a conventional FIS, most of these rules are built with the help of human expert knowledge. For example, in our case, such experts can be doctor, police, forensic expert or researcher, etc. The information that they provide is considered to be the most reliable one as they build it based on their real life experiences and historical analysis. However, human intervention in an intelligent system is becoming a threat due to the heuristic and subjectivity of human decisions. Therefore, automated learning systems have emerged and widely employed in the research society, encouraging learning and generation of fuzzy rules automatically. Several works in the literature have reported efficient methods for the automatic generation of the fuzzy rules such as \cite{wang1992generating,rhee1993fuzzy,wang2001new,cheng1995fast,cordon2001generating}.

For example, \cite{wang1992generating} proposed a method of generating the fuzzy rules by learning from examples, more specifically by the numerical data. Similarly, \cite{rhee1993fuzzy} presented an alternative method to generate the fuzzy rules automatically from the training data with their rules defined in the form of possibility, certainty, gradual, and unless rules. A new approach called the fuzzy extension matrix was proposed in \cite{wang2001new} which incorporated the fuzzy entropy to search for the paths and generalized the concept of the crisp extension matrix. Their method was capable of handling the fuzzy representation and tolerating the noisy or missing data. FCM and its variants (e.g. multi-stage random sampling) with its fast performance have also been adopted in the fuzzy rule generation such as the work by \cite{cheng1995fast}. Apart from that, there are works reported in the fuzzy rule generation incorporated with other machine learning techniques. \cite{mitra2000neuro} provided an exhaustive survey on the neuro-fuzzy rule generation algorithms, while \cite{cordon2001generating} presented an approach to automatically learn the fuzzy rules by incorporating the genetic algorithm.

\begin{table}[htbp]
	\centering
		\caption{The current best results of applying the fuzzy approaches and other stochastic methods on the well known datasets in HMA. RA indicates the recognition accuracy and TP is the tracking precision.}
		\label{Table:dataset_hma}
		\resizebox{15cm}{!} {
		\begin{tabular}{p{4cm} c c c c c }\hline	
		
%
		
  		\multirow{2}{*}{Name} & Dataset & Dataset & Fuzzy paper that & Best accuracy in & Best accuracy in \\
  		
  		 & Established Year & Reference & uses this dataset & fuzzy approach(s) (\%) & other method(s) (\%) \\ \hline \hline \\
		
		KTH & 2004 & \cite{schuldt2004recognizing} & \cite{chan2008fuzzy,chan2009fuzzy,iosifidis2013minimum} & RA = 93.52 \cite{iosifidis2013minimum} & RA = 96.76 \cite{sapienza2014learning}\\ \hline \\
		
		CAVIAR & 2004 & \cite{fisher2004pets04} & - & - & TP = 91.90 \cite{nie2014single}\\ \hline \\

		WEIZMANN \textit{Actions} & 2005 & \cite{blank2005actions} & \cite{yao2014fuzzy,mozafari2012novel,gkalelis2008combining,chan2008fuzzy,chan2009fuzzy} & RA = 100.00 \cite{chan2009fuzzy} & RA = 100.00 \cite{chen2009recognizing} \\ \hline \\

		IXMAS & 2006 & \cite{weinland2006free} & \cite{Iosifidis2012,lim2013fuzz} & RA = 83.47 \cite{Iosifidis2012} & RA = 95.54 \cite{wu2014multi} \\ \hline \\		
		
		CASIA \textit{Action} & 2007 & \cite{wang2007human} & - & - & RA = 99.90 \cite{lu2014application} \\ \hline \\	
		
				ETISEO & 2007 & \cite{nghiem2007etiseo} & - & - & TP = 100.00 \cite{simha2014feature}\\ \hline \\	
				
					UIUC - Complex action & 2007 & \cite{ikizler2007human} & \cite{chan2010fuzzy} & RA > 80.00 \cite{chan2010fuzzy} & - \\
		UIUC & 2008 & \cite{tran2008human} & - & - & RA = 93.30 \cite{tu2014complex}\\ \hline \\
			
					CMU MoCap & 2008 & \cite{de2008guide} &  \cite{gkalelis2008combining} & RA = 98.90 \cite{gkalelis2008combining} & RA = 98.30 \cite{john2014charting} \\ \hline \\

		ViHASi & 2008 & \cite{ragheb2008vihasi} & - & - & RA = 72.00 \cite{zhang2013grassmann}\\ \hline \\
		
			HOLLYWOOD & 2008 & \cite{laptev2008learning} & - & - & RA = 61.50 \cite{du2014recognizing}\\ 
		HOLLYWOOD-2 & 2009 & \cite{marszalek2009actions} & - & - & RA = 64.30 \cite{wang2013action}\\ \hline \\
				
		UCF-Sports & 2008 & \cite{action-shah-action-mach-cvpr08} & \cite{iosifidis2013minimum} & RA = 85.77 \cite{iosifidis2013minimum} & RA = 89.70 \cite{wu2011action}\\		
		UCF-11 \textit{Youtube} & 2009 & \cite{liu2009recognizing} & - & - & RA = 89.79 \cite{sapienza2014learning}\\
		 \hline \\

		i3DPost & 2009 & \cite{gkalelis2009i3dpost} & \cite{Iosifidis2012,Iosifidis2012b,iosifidis2013minimum} & RA = 100.00 \cite{iosifidis2013minimum} & RA = 98.44 \cite{holte2012local}\\ \hline \\
		
				UT-Interaction & 2009 & \cite{ryoo2009spatio} & - & - & RA = 91.67 \cite{fu2014interactive} \\
		UT-Tower & 2009 & \cite{chen2009recognizing} & - & - & \\ \hline \\

		MSR \textit{Action} & 2009 & \cite{yuan2009discriminative} & - & - & \\ 
		
		MSR \textit{3D Action} & 2010 & \cite{li2010action} & - & - & RA = 97.80 \cite{yang2014effective}\\ \hline \\

				BEHAVE & 2010 & \cite{blunsden2010behave} & - & - & RA = 65.50 \cite{cheng2014recognizing} \\ \hline \\
		
		MuHAVi & 2010 & \cite{singh2010muhavi} & - & - & RA = 100.00 \cite{chaaraoui2013optimizing}\\ \hline \\

		Olympic Sports & 2010 & \cite{niebles2010modeling} & - & - & RA = 91.10 \cite{wang2013action}\\ \hline \\
		
		TV Human Interaction & 2010 & \cite{patron2010high} & - & - & RA = 46.00 \cite{marin2014human}\\ \hline \\	
		
		HMDB51 & 2011 & \cite{kuehne2011hmdb} & - & - & RA = 57.20 \cite{wang2013action}\\ \hline \\
			
		VideoWeb & 2011 & \cite{denina2011videoweb} & - & - & RA = 72.00 \cite{zha2013detecting}\\ \hline \\
		
		UCF-101 & 2012 & \cite{soomro2012ucf101} & - & - & RA = 83.50 \cite{caimulti}\\ 
		UCF-50 & 2013 & \cite{reddy2013recognizing} & - & - & RA = 91.20 \cite{wang2013action}\\ \hline
								
		\end{tabular}
		}

\end{table}

\subsection{Potential future works in fuzzy HMA} 

\noindent \textbf{\emph{Datasets:}} In the research society nowadays, public datasets play a very important role in order to show the effectiveness of a proposed algorithm. Even so, from our findings in Table \ref{Table:dataset_hma}, there were not much works from the fuzzy community that had explored these public datasets. Only a handful works in the fuzzy HMA (as referred in Table \ref{Table:dataset_hma}) had employed those datasets and compared their works with other algorithms. In order to justify and improve the competency of the fuzzy approaches in HMA, it is believed that one way forward is to start employing these datasets as the baseline study.

On the other hand, the datasets listed in Table \ref{Table:dataset_hma} undeniably has met the objectives as a baseline evaluation. However, \cite{boutell2004learning,parikh2011relative,lim2012fuzzy} raised an argument that many situations in the real life are ambiguous, especially the human behavior with varied perceptions of the masses. The current datasets, at this stage might be too ideal to reflect the real world scenarios, i.e. the current datasets are mutually exclusive, allowing a data to belong to one class (action) only at a time. Therefore, another potential area which can be explored as future works is having an appropriate psycho-physical dataset with fuzzy ground truths, or in a simpler sense: fuzzy datasets. To the best of our knowledge, there do not exist any fuzzy datasets modeling the human activities and their behavior till date. \\

\noindent \textbf{\emph{Early event detection:}} Apart from that, fuzzy approaches being successful in handling the uncertainties in various real-time applications as highlighted in this survey, can be very well explored to be potentially applied in highly complex HMA applications such as human activity forecasting \cite{kitani2012activity} and early detection of crimes \cite{ryoo2011human,hoai2012max}. There do not exist literature on the fuzzy capability in handling the uncertainties arising in such scenarios, which have high quotient of importance as they are focusing on forecasting an event or early detecting crimes from happening. Therefore, even the minutest level of uncertainty is required to be taken care of for reliable decision making. Fuzzy approaches with its capability in handling the uncertain situations can substantially benefit in performing these complex tasks, and can be explored by the researchers working in this domain as a potential future work. \\

\noindent \textbf{\emph{Human activity recognition in still images:}} Another interesting area to be explored as part of the future works is the recognition of human activities using still image. The work has has received much attention in the recent past in the computer vision community \cite{gupta2009observing,yao2010grouplet,desai2010discriminative,yang2010recognizing,delaitre2010recognizing,maji2011action,prest2012weakly},  but as to our very best knowledge, none was found in the fuzzy domain. In this research topic, most of the works considered it to be same as an image classification problem. Lately, several researchers are trying to obtain a thorough understanding of the human poses, the objects, and the interactions between them in a still images to infer the activities. For example, \cite{yao2012recognizing} proposed a method to recognize the human-object interactions in still images by explicitly modeling the mutual context between the human poses and the objects, so that each can facilitate the recognition of the other. Their mutual context model outperform the state-of-the-art in object detection, human pose estimation, as well as the recognition of human-object interaction activities. However limited information that can be extracted from the still image and the random noises in the image are two major problems that exists in this area. This can be a potential area to explore by the fuzzy community, providing useful solutions in handling the uncertainties, incomplete data or vague information in regards with the human-object interactions, or human-scene context in still images.

\section{Conclusion}
\label{con}

Fuzzy set theory has been effectively applied in many ways that revealed a number of fuzzy approaches such as FIS, FCM, Fuzzy qualitative reasoning, etc. This paper takes the initiative to review the works that employed these fuzzy approaches in the HMA system which has not been done previously. From the studies, one can notice that the fuzzy approaches are capable of handling the uncertainty that abounded in each level of the HMA system (LoL, MiL, and HiL). The fundamental factors that endowed such capability to the fuzzy approaches include the ability to perform soft labeling and the flexibility to adapt to different uncertainties. However, most of the reported works herein did not utilize the standard HMA datasets as their baseline. Anyway, the current datasets are mostly too ideal to reflect the real world scenarios that is full of uncertainties. The generation of the fuzzy dataset for HMA could be one of the potential future works other than the early event detection and the still image action recognition.

\section*{Acknowledgment}
\label{ack}

This research is supported by the High Impact Research MoE Grant UM.C/625/1/HIR/MoE/FCSIT/08, H-22001-00-B0008 from the Ministry of Education Malaysia.

\bibliographystyle{elsarticle-num}
\bibliography{Qiqqa2BibTexExport}

\begin{thebibliography}{100}
\expandafter\ifx\csname url\endcsname\relax
  \def\url#1{\texttt{#1}}\fi
\expandafter\ifx\csname urlprefix\endcsname\relax\def\urlprefix{URL }\fi
\expandafter\ifx\csname href\endcsname\relax
  \def\href#1#2{#2} \def\path#1{#1}\fi

\bibitem{bobick1997movement}
A.~F. Bobick, Movement, activity and action: the role of knowledge in the
  perception of motion, Philosophical Transactions of the Royal Society of
  London. Series B: Biological Sciences 352~(1358) (1997) 1257--1265.

\bibitem{troje2002decomposing}
N.~F. Troje, Decomposing biological motion: A framework for analysis and
  synthesis of human gait patterns, Journal of Vision 2~(5) (2002) 2.

\bibitem{barclay1978temporal}
C.~D. Barclay, J.~E. Cutting, L.~T. Kozlowski, Temporal and spatial factors in
  gait perception that influence gender recognition, Perception \&
  Psychophysics 23~(2) (1978) 145--152.

\bibitem{blake2007perception}
R.~Blake, M.~Shiffrar, Perception of human motion, Annu. Rev. Psychol. 58
  (2007) 47--73.

\bibitem{kirtley2001application}
C.~Kirtley, R.~Smith, Application of multimedia to the study of human movement,
  Multimedia Tools and Applications 14~(3) (2001) 259--268.

\bibitem{haering2008evolution}
N.~Haering, P.~L. Venetianer, A.~Lipton, The evolution of video surveillance:
  an overview, Machine Vision and Applications 19~(5-6) (2008) 279--290.

\bibitem{hu2004survey}
W.~Hu, T.~Tan, L.~Wang, S.~Maybank, A survey on visual surveillance of object
  motion and behaviors, IEEE Transactions on Systems, Man, and Cybernetics,
  Part C: Applications and Reviews 34~(3) (2004) 334--352.

\bibitem{kim2010intelligent}
I.~S. Kim, H.~S. Choi, K.~M. Yi, J.~Y. Choi, S.~G. Kong, Intelligent visual
  surveillance: A survey, International Journal of Control, Automation and
  Systems 8~(5) (2010) 926--939.

\bibitem{ko2008survey}
T.~Ko, A survey on behavior analysis in video surveillance for homeland
  security applications, in: 37th Applied Imagery Pattern Recognition Workshop,
  2008, pp. 1--8.

\bibitem{popoola2012video}
O.~P. Popoola, K.~Wang, Video-based abnormal human behavior recognition - a
  review, IEEE Transactions on Systems, Man, and Cybernetics, Part C:
  Applications and Reviews 42~(6) (2012) 865--878.

\bibitem{geetha2008survey}
P.~Geetha, V.~Narayanan, A survey of content-based video retrieval, Journal of
  Computer Science 4~(6) (2008) 474.

\bibitem{efros2003recognizing}
A.~A. Efros, A.~C. Berg, G.~Mori, J.~Malik, Recognizing action at a distance,
  in: Proceedings. Ninth IEEE International Conference on Computer Vision
  (ICCV), 2003, pp. 726--733.

\bibitem{loy2004monocular}
G.~Loy, M.~Eriksson, J.~Sullivan, S.~Carlsson, Monocular 3d reconstruction of
  human motion in long action sequences, in: European Conference on Computer
  Vision (ECCV), Springer, 2004, pp. 442--455.

\bibitem{sullivan2008action}
M.~Sullivan, M.~Shah, Action mach: Maximum average correlation height filter
  for action recognition, in: IEEE Conference on Computer Vision and Pattern
  Recognition (CVPR), 2008, pp. 1--8.

\bibitem{anderson2006recognizing}
D.~Anderson, J.~M. Keller, M.~Skubic, X.~Chen, Z.~He, Recognizing falls from
  silhouettes, in: 28th Annual International Conference of the IEEE Engineering
  in Medicine and Biology Society (EMBS), 2006, pp. 6388--6391.

\bibitem{anderson2009modeling}
D.~Anderson, R.~H. Luke, J.~M. Keller, M.~Skubic, M.~J. Rantz, M.~A. Aud,
  Modeling human activity from voxel person using fuzzy logic, IEEE
  Transactions on Fuzzy Systems 17~(1) (2009) 39--49.

\bibitem{jaimes2007multimodal}
A.~Jaimes, N.~Sebe, Multimodal human--computer interaction: A survey, Computer
  Vision and Image Understanding 108~(1) (2007) 116--134.

\bibitem{aggarwal1994articulated}
J.~K. Aggarwal, Q.~Cai, W.~Liao, B.~Sabata, Articulated and elastic non-rigid
  motion: A review, in: Proceedings of the IEEE Workshop on Motion of Non-Rigid
  and Articulated Objects, 1994, pp. 2--14.

\bibitem{cedras1995motion}
C.~C{\'e}dras, M.~Shah, Motion-based recognition a survey, Image and Vision
  Computing 13~(2) (1995) 129--155.

\bibitem{aggarwal1997human}
J.~K. Aggarwal, Q.~Cai, Human motion analysis: A review, in: Proceedings of the
  IEEE Nonrigid and Articulated Motion Workshop, 1997, pp. 90--102.

\bibitem{gavrila1999visual}
D.~M. Gavrila, The visual analysis of human movement: A survey, Computer Vision
  and Image Understanding 73~(1) (1999) 82--98.

\bibitem{pentland2000looking}
A.~Pentland, Looking at people: Sensing for ubiquitous and wearable computing,
  IEEE Transactions on Pattern Analysis and Machine Intelligence 22~(1) (2000)
  107--119.

\bibitem{moeslund2001survey}
T.~B. Moeslund, E.~Granum, A survey of computer vision-based human motion
  capture, Computer Vision and Image Understanding 81~(3) (2001) 231--268.

\bibitem{wang2003recent}
L.~Wang, W.~Hu, T.~Tan, Recent developments in human motion analysis, Pattern
  Recognition 36~(3) (2003) 585--601.

\bibitem{moeslund2006survey}
T.~B. Moeslund, A.~Hilton, V.~Kr{\"u}ger, A survey of advances in vision-based
  human motion capture and analysis, Computer Vision and Image Understanding
  104~(2) (2006) 90--126.

\bibitem{poppe2007vision}
R.~Poppe, Vision-based human motion analysis: An overview, Computer Vision and
  Image Understanding 108~(1) (2007) 4--18.

\bibitem{turaga2008machine}
P.~Turaga, R.~Chellappa, V.~S. Subrahmanian, O.~Udrea, Machine recognition of
  human activities: A survey, IEEE Transactions on Circuits and Systems for
  Video Technology 18~(11) (2008) 1473--1488.

\bibitem{ji2010advances}
X.~Ji, H.~Liu, Advances in view-invariant human motion analysis: a review, IEEE
  Transactions on Systems, Man, and Cybernetics, Part C: Applications and
  Reviews 40~(1) (2010) 13--24.

\bibitem{poppe2010survey}
R.~Poppe, A survey on vision-based human action recognition, Image and Vision
  Computing 28~(6) (2010) 976--990.

\bibitem{candamo2010understanding}
J.~Candamo, M.~Shreve, D.~B. Goldgof, D.~B. Sapper, R.~Kasturi, Understanding
  transit scenes: A survey on human behavior-recognition algorithms, IEEE
  Transactions on Intelligent Transportation Systems 11~(1) (2010) 206--224.

\bibitem{aggarwal2011human}
J.~Aggarwal, M.~S. Ryoo, Human activity analysis: A review, ACM Computing
  Surveys 43~(3) (2011) 16.

\bibitem{weinland2011survey}
D.~Weinland, R.~Ronfard, E.~Boyer, A survey of vision-based methods for action
  representation, segmentation and recognition, Computer Vision and Image
  Understanding 115~(2) (2011) 224--241.

\bibitem{holte2011human}
M.~B. Holte, C.~Tran, M.~M. Trivedi, T.~B. Moeslund, Human action recognition
  using multiple views: a comparative perspective on recent developments, in:
  Proceedings of the Joint ACM Workshop on Human Gesture and Behavior
  Understanding, 2011, pp. 47--52.

\bibitem{lara2013survey}
O.~D. Lara, M.~A. Labrador, A survey on human activity recognition using
  wearable sensors, IEEE Communications Surveys \& Tutorials 15~(3) (2013)
  1192--1209.

\bibitem{chen2013survey}
L.~Chen, H.~Wei, J.~Ferryman, A survey of human motion analysis using depth
  imagery, Pattern Recognition Letters 34~(15) (2013) 1995--2006.

\bibitem{cristani2013human}
M.~Cristani, R.~Raghavendra, A.~Del~Bue, V.~Murino, Human behavior analysis in
  video surveillance: A social signal processing perspective, Neurocomputing
  100 (2013) 86--97.

\bibitem{chaquet2013survey}
J.~M. Chaquet, E.~J. Carmona, A.~Fern{\'a}ndez-Caballero, A survey of video
  datasets for human action and activity recognition, Computer Vision and Image
  Understanding 117~(6) (2013) 633--659.

\bibitem{schuldt2004recognizing}
C.~Schuldt, I.~Laptev, B.~Caputo, Recognizing human actions: a local svm
  approach, in: Proceedings of the International Conference on Pattern
  Recognition (ICPR), Vol.~3, 2004, pp. 32--36.

\bibitem{zelnik2001event}
L.~Zelnik-Manor, M.~Irani, Event-based analysis of video, in: IEEE Conference
  on Computer Vision and Pattern Recognition (CVPR), Vol.~2, 2001, pp. II--123.

\bibitem{blank2005actions}
M.~Blank, L.~Gorelick, E.~Shechtman, M.~Irani, R.~Basri, Actions as space-time
  shapes, in: IEEE International Conference on Computer Vision (ICCV), Vol.~2,
  2005, pp. 1395--1402.

\bibitem{huntsberger1986representation}
T.~L. Huntsberger, C.~Rangarajan, S.~N. Jayaramamurthy, Representation of
  uncertainty in computer vision using fuzzy sets, IEEE Transactions on
  Computers 100~(2) (1986) 145--156.

\bibitem{krishnapuram1992fuzzy}
R.~Krishnapuram, J.~M. Keller, Fuzzy set theoretic approach to computer vision:
  An overview, in: IEEE International Conference on Fuzzy Systems (FUZZ), 1992,
  pp. 135--142.

\bibitem{sobrevilla2003fuzzy}
P.~Sobrevilla, E.~Montseny, Fuzzy sets in computer vision: An overview,
  Mathware \& Soft Computing 10~(3) (2008) 71--83.

\bibitem{bobick2001recognition}
A.~F. Bobick, J.~W. Davis, The recognition of human movement using temporal
  templates, IEEE Transactions on Pattern Analysis and Machine Intelligence
  23~(3) (2001) 257--267.

\bibitem{weinland2006free}
D.~Weinland, R.~Ronfard, E.~Boyer, Free viewpoint action recognition using
  motion history volumes, Computer Vision and Image Understanding 104~(2)
  (2006) 249--257.

\bibitem{lewandowski2010view}
M.~Lewandowski, D.~Makris, J.-C. Nebel, View and style-independent action
  manifolds for human activity recognition, in: European Conference on Computer
  Vision (ECCV), 2010, pp. 547--560.

\bibitem{zhang2006fusing}
H.~Zhang, D.~Xu, Fusing color and texture features for background model, in:
  Proceedings of the International Conference on Fuzzy Systems and Knowledge
  Discovery (FSKD), 2006, pp. 887--893.

\bibitem{el2008fuzz}
F.~El~Baf, T.~Bouwmans, B.~Vachon, Fuzzy integral for moving object detection,
  in: IEEE International Conference on Fuzzy Systems (FUZZ), 2008, pp.
  1729--1736.

\bibitem{tahani1990information}
H.~Tahani, J.~M. Keller, Information fusion in computer vision using the fuzzy
  integral, Systems, Man and Cybernetics, IEEE Transactions on 20~(3) (1990)
  733--741.

\bibitem{marichal2000sugeno}
J.-L. Marichal, On sugeno integral as an aggregation function, Fuzzy Sets and
  Systems 114~(3) (2000) 347--365.

\bibitem{el2008fuzzy}
F.~El~Baf, T.~Bouwmans, B.~Vachon, A fuzzy approach for background subtraction,
  in: IEEE International Conference on Image Processing (ICIP), 2008, pp.
  2648--2651.

\bibitem{balcilar2013region}
M.~Balcilar, A.~C. Sonmez, Region based fuzzy background subtraction using
  choquet integral, in: Adaptive and Natural Computing Algorithms, Springer,
  2013, pp. 287--296.

\bibitem{murofushi1989interpretation}
T.~Murofushi, M.~Sugeno, An interpretation of fuzzy measures and the choquet
  integral as an integral with respect to a fuzzy measure, Fuzzy Sets and
  Systems 29~(2) (1989) 201--227.

\bibitem{sugeno1995new}
M.~Sugeno, S.-H. Kwon, A new approach to time series modeling with fuzzy
  measures and the choquet integral, in: Proceedings of IEEE International
  Joint Conference of the Fourth IEEE International Conference on Fuzzy Systems
  and The Second International Fuzzy Engineering Symposium, Vol.~2, 1995, pp.
  799--804.

\bibitem{narukawa2004decision}
Y.~Narukawa, T.~Murofushi, Decision modelling using the choquet integral, in:
  Modeling Decisions for Artificial Intelligence, Springer, 2004, pp. 183--193.

\bibitem{piccardi2004background}
M.~Piccardi, Background subtraction techniques: a review, in: IEEE
  International Conference on Systems, Man and Cybernetics (SMC), Vol.~4, 2004,
  pp. 3099--3104.

\bibitem{cheung2004robust}
S.-C.~S. Cheung, C.~Kamath, Robust techniques for background subtraction in
  urban traffic video, in: Proceedings of SPIE, Vol. 5308, 2004, pp. 881--892.

\bibitem{zeng2008type}
J.~Zeng, L.~Xie, Z.-Q. Liu, Type-2 fuzzy gaussian mixture models, Pattern
  Recognition 41~(12) (2008) 3636--3643.

\bibitem{zadeh1965fuzzy}
L.~Zadeh, Fuzzy sets, Information and Control 8~(3) (1965) 338--353.

\bibitem{mendel2002type}
J.~M. Mendel, R.~B. John, Type-2 fuzzy sets made simple, IEEE Transactions on
  Fuzzy Systems 10~(2) (2002) 117--127.

\bibitem{el2008type}
F.~El~Baf, T.~Bouwmans, B.~Vachon, Type-2 fuzzy mixture of gaussians model:
  application to background modeling, in: Advances in Visual Computing,
  Springer, 2008, pp. 772--781.

\bibitem{el2009fuzzy}
F.~El~Baf, T.~Bouwmans, B.~Vachon, Fuzzy statistical modeling of dynamic
  backgrounds for moving object detection in infrared videos, in: IEEE Computer
  Society Conference on Computer Vision and Pattern Recognition (CVPRW), 2009,
  pp. 60--65.

\bibitem{bouwmans2009modeling}
T.~Bouwmans, F.~El~Baf, et~al., Modeling of dynamic backgrounds by type-2 fuzzy
  gaussians mixture models, MASAUM Journal of of Basic and Applied Sciences
  1~(2) (2009) 265--276.

\bibitem{zhao2012fuzzy}
Z.~Zhao, T.~Bouwmans, X.~Zhang, Y.~Fang, A fuzzy background modeling approach
  for motion detection in dynamic backgrounds, in: Multimedia and Signal
  Processing, Springer, 2012, pp. 177--185.

\bibitem{sigari2008fuzzy}
M.~H. Sigari, N.~Mozayani, H.~R. Pourreza, Fuzzy running average and fuzzy
  background subtraction: concepts and application, International Journal of
  Computer Science and Network Security 8~(2) (2008) 138--143.

\bibitem{lin2000neural}
C.~Lin, I.~Chung, L.~Sheu, A neural fuzzy system for image motion estimation,
  Fuzzy Sets and Systems 114~(2) (2000) 281--304.

\bibitem{maddalena2010fuzzy}
L.~Maddalena, A.~Petrosino, A fuzzy spatial coherence-based approach to
  background/foreground separation for moving object detection, Neural
  Computing and Applications 19~(2) (2010) 179--186.

\bibitem{li2012adaptive}
Z.~Li, W.~Liu, Y.~Zhang, Adaptive fuzzy apporach to background modeling using
  pso and klms, in: 10th World Congress on Intelligent Control and Automation
  (WCICA), 2012, pp. 4601--4607.

\bibitem{calvo2013fuzzy}
E.~Calvo-Gallego, P.~Brox, S.~S{\'a}nchez-Solano, A fuzzy system for background
  modeling in video sequences, in: Fuzzy Logic and Applications, Springer,
  2013, pp. 184--192.

\bibitem{shakeri2008novel}
M.~Shakeri, H.~Deldari, H.~Foroughi, A.~Saberi, A.~Naseri, A novel fuzzy
  background subtraction method based on cellular automata for urban traffic
  applications, in: International Conference on Signal Processing (ICSP), 2008,
  pp. 899--902.

\bibitem{mendel2006interval}
J.~M. Mendel, R.~I. John, F.~Liu, Interval type-2 fuzzy logic systems made
  simple, IEEE Transactions on Fuzzy Systems 14~(6) (2006) 808--821.

\bibitem{yager1992introduction}
R.~R. Yager, L.~Zadeh, An introduction to fuzzy logic applications in
  intelligent systems, Springer, 1992.

\bibitem{zadeh1988fuzzy}
L.~Zadeh, Fuzzy logic, Computer 21~(4) (1988) 83--93.

\bibitem{mahapatra2013background}
A.~Mahapatra, T.~K. Mishra, P.~K. Sa, B.~Majhi, Background subtraction and
  human detection in outdoor videos using fuzzy logic, in: IEEE International
  Conference on Fuzzy Systems (FUZZ), 2013, pp. 1--7.

\bibitem{see2005human}
J.~See, S.~Lee, M.~Hanmandlu, Human motion detection using fuzzy rule-base
  classification of moving blob regions, in: Proc. Int. Conf. on Robotics,
  Vision, Information and Signal Processing 2005, 2005, pp. 398--402.

\bibitem{chowdhury2014detection}
A.~Chowdhury, S.~S. Tripathy, Detection of human presence in a surveillance
  video using fuzzy approach, in: International Conference on Signal Processing
  and Integrated Networks (SPIN), 2014, pp. 216--219.

\bibitem{chen2006adaptive}
X.~Chen, Z.~He, D.~Anderson, J.~Keller, M.~Skubic, Adaptive silouette
  extraction and human tracking in complex and dynamic environments, in: IEEE
  International Conference on Image Processing (ICIP), 2006, pp. 561--564.

\bibitem{Chen2006}
X.~Chen, Z.~He, J.~M. Keller, D.~Anderson, M.~Skubic, Adaptive silhouette
  extraction in dynamic environments using fuzzy logic, in: IEEE International
  Conference on Fuzzy Systems (FUZZ), 2006, pp. 236--243.

\bibitem{yao2012interval}
B.~Yao, H.~Hagras, D.~Al~Ghazzawi, M.~J. Alhaddad, An interval type-2 fuzzy
  logic system for human silhouette extraction in dynamic environments, in:
  Autonomous and Intelligent Systems, Springer, 2012, pp. 126--134.

\bibitem{liang2000interval}
Q.~Liang, J.~M. Mendel, Interval type-2 fuzzy logic systems: theory and design,
  IEEE Transactions on Fuzzy Systems 8~(5) (2000) 535--550.

\bibitem{karnik1998type}
N.~N. Karnik, J.~M. Mendel, Type-2 fuzzy logic systems: type-reduction, in:
  IEEE International Conference on Systems, Man, and Cybernetics (SMC), Vol.~2,
  1998, pp. 2046--2051.

\bibitem{guo1994tracking}
Y.~Guo, G.~Xu, S.~Tsuji, Tracking human body motion based on a stick figure
  model, Journal of Visual Communication and Image Representation 5~(1) (1994)
  1--9.

\bibitem{leung1995first}
M.~K. Leung, Y.-H. Yang, First sight: A human body outline labeling system,
  IEEE Transactions on Pattern Analysis and Machine Intelligence 17~(4) (1995)
  359--377.

\bibitem{iwai1999posture}
Y.~Iwai, K.~Ogaki, M.~Yachida, Posture estimation using structure and motion
  models, in: IEEE International Conference on Computer Vision (ICCV), Vol.~1,
  1999, pp. 214--219.

\bibitem{silaghi1998local}
M.-C. Silaghi, R.~Pl{\"a}nkers, R.~Boulic, P.~Fua, D.~Thalmann, Local and
  global skeleton fitting techniques for optical motion capture, in: Modelling
  and Motion Capture Techniques for Virtual Environments, Springer, 1998, pp.
  26--40.

\bibitem{niyogi1994analyzing}
S.~A. Niyogi, E.~H. Adelson, Analyzing and recognizing walking figures in xyt,
  in: IEEE Conference on Computer Vision and Pattern Recognition (CVPR), 1994,
  pp. 469--474.

\bibitem{ju1996cardboard}
S.~X. Ju, M.~J. Black, Y.~Yacoob, Cardboard people: A parameterized model of
  articulated image motion, in: Proceedings of the Second International
  Conference on Automatic Face and Gesture Recognition (FG), 1996, pp. 38--44.

\bibitem{rohr1994towards}
K.~Rohr, Towards model-based recognition of human movements in image sequences,
  CVGIP: Image understanding 59~(1) (1994) 94--115.

\bibitem{wachter1997tracking}
S.~Wachter, H.-H. Nagel, Tracking of persons in monocular image sequences, in:
  Proceedings of IEEE Nonrigid and Articulated Motion Workshop, 1997, pp. 2--9.

\bibitem{rehg1995model}
J.~M. Rehg, T.~Kanade, Model-based tracking of self-occluding articulated
  objects, in: Proceedings of International Conference on Computer Vision
  (ICCV), 1995, pp. 612--617.

\bibitem{kakadiaris1996model}
I.~A. Kakadiaris, D.~Metaxas, Model-based estimation of 3d human motion with
  occlusion based on active multi-viewpoint selection, in: IEEE Conference on
  Computer Vision and Pattern Recognition (CVPR), IEEE, 1996, pp. 81--87.

\bibitem{Moeslund2001}
T.~B. Moeslund, E.~Granum, A survey of computer vision-based human motion
  capture, Computer Vision and Image Understanding 81~(3) (2001) 231--268.

\bibitem{ning2004kinematics}
H.~Ning, T.~Tan, L.~Wang, W.~Hu, Kinematics-based tracking of human walking in
  monocular video sequences, Image and Vision Computing 22~(5) (2004) 429--441.

\bibitem{bregler2004twist}
C.~Bregler, J.~Malik, K.~Pullen, Twist based acquisition and tracking of animal
  and human kinematics, International Journal of Computer Vision 56~(3) (2004)
  179--194.

\bibitem{Liu2008a}
H.~Liu, Fuzzy qualitative robot kinematics, IEEE Transactions on Fuzzy Systems
  16~(6) (2008) 1522--1530.

\bibitem{chan2009fuzzy}
C.~S. Chan, H.~Liu, Fuzzy qualitative human motion analysis, IEEE Transactions
  on Fuzzy Systems 17~(4) (2009) 851--862.

\bibitem{shen1993fuzzy}
Q.~Shen, R.~Leitch, Fuzzy qualitative simulation, IEEE Transactions on Systems,
  Man and Cybernetics 23~(4) (1993) 1038--1061.

\bibitem{chan2011recent}
C.~S. Chan, G.~M. Coghill, H.~Liu, Recent advances in fuzzy qualitative
  reasoning, International Journal of Uncertainty, Fuzziness and
  Knowledge-Based Systems 19~(03) (2011) 417--422.

\bibitem{kuipers1986qualitative}
B.~Kuipers, Qualitative simulation, Artificial Intelligence 29~(3) (1986)
  289--338.

\bibitem{liu2009fuzzy}
H.~Liu, G.~M. Coghill, D.~P. Barnes, Fuzzy qualitative trigonometry,
  International Journal of Approximate Reasoning 51~(1) (2009) 71--88.

\bibitem{Liu2008}
H.~Liu, D.~J. Brown, G.~M. Coghill, A fuzzy qualitative framework for
  connecting robot qualitative and quantitative representations, IEEE
  Transactions on Fuzzy Systems 16~(3) (2008) 808--822.

\bibitem{chan2008fuzzy}
C.~S. Chan, H.~Liu, D.~Brown, N.~Kubota, A fuzzy qualitative approach to human
  motion recognition, in: IEEE International Conference on Fuzzy Systems
  (FUZZ), 2008, pp. 1242--1249.

\bibitem{anderson2009fuzzy}
D.~Anderson, R.~H. Luke~III, E.~E. Stone, J.~M. Keller, Fuzzy voxel object.,
  in: IFSA/EUSFLAT Conf., 2009, pp. 282--287.

\bibitem{anderson2009linguistic}
D.~Anderson, R.~H. Luke, J.~M. Keller, M.~Skubic, M.~Rantz, M.~Aud, Linguistic
  summarization of video for fall detection using voxel person and fuzzy logic,
  Computer Vision and Image Understanding 113~(1) (2009) 80--89.

\bibitem{garcia2002robust}
J.~Garc{\'\i}a, J.~M. Molina, J.~A. Besada, J.~I. Portillo, J.~R. Casar, Robust
  object tracking with fuzzy shape estimation, in: Proceedings of the
  International Conference on Information Fusion, Vol.~1, 2002, pp. 64--71.

\bibitem{garcia2011fuzzy}
J.~Garcia, M.~A. Patricio, A.~Berlanga, J.~M. Molina, Fuzzy region assignment
  for visual tracking, Soft Computing 15~(9) (2011) 1845--1864.

\bibitem{kalman1960new}
R.~E. Kalman, A new approach to linear filtering and prediction problems,
  Journal of Basic Engineering 82~(1) (1960) 35--45.

\bibitem{kohler1997using}
M.~Kohler, Using the Kalman filter to track human interactive motion: modelling
  and initialization of the Kalman filter for translational motion, Citeseer,
  1997.

\bibitem{yun2005implementation}
X.~Yun, C.~Aparicio, E.~R. Bachmann, R.~B. McGhee, Implementation and
  experimental results of a quaternion-based kalman filter for human body
  motion tracking, in: Proceedings of the IEEE International Conference on
  Robotics and Automation (ICRA), 2005, pp. 317--322.

\bibitem{yun2006design}
X.~Yun, E.~R. Bachmann, Design, implementation, and experimental results of a
  quaternion-based kalman filter for human body motion tracking, IEEE
  Transactions on Robotics 22~(6) (2006) 1216--1227.

\bibitem{marins2001extended}
J.~L. Marins, X.~Yun, E.~R. Bachmann, R.~B. McGhee, M.~J. Zyda, An extended
  kalman filter for quaternion-based orientation estimation using marg sensors,
  in: Proceedings of IEEE/RSJ International Conference on Intelligent Robots
  and Systems, Vol.~4, 2001, pp. 2003--2011.

\bibitem{welch2009history}
G.~F. Welch, History: The use of the kalman filter for human motion tracking in
  virtual reality, Presence: Teleoperators and Virtual Environments 18~(1)
  (2009) 72--91.

\bibitem{welch1995introduction}
G.~Welch, G.~Bishop, An introduction to the kalman filter (1995).

\bibitem{chen1998fuzzy}
G.~Chen, Q.~Xie, L.~S. Shieh, Fuzzy kalman filtering, Information Sciences
  109~(1) (1998) 197--209.

\bibitem{kobayashi1998accurate}
K.~Kobayashi, K.~C. Cheok, K.~Watanabe, F.~Munekata, Accurate differential
  global positioning system via fuzzy logic kalman filter sensor fusion
  technique, IEEE Transactions on Industrial Electronics 45~(3) (1998)
  510--518.

\bibitem{sasiadek1999sensor}
J.~Sasiadek, Q.~Wang, Sensor fusion based on fuzzy kalman filtering for
  autonomous robot vehicle, in: Proceedings of the International Conference on
  Robotics and Automation (ICRA), Vol.~4, 1999, pp. 2970--2975.

\bibitem{sasiadek2000fuzzy}
J.~Sasiadek, Q.~Wang, M.~Zeremba, Fuzzy adaptive kalman filtering for ins/gps
  data fusion, in: Proceedings of the IEEE International Symposium on
  Intelligent Control, 2000, pp. 181--186.

\bibitem{sasiadek2001sensor}
J.~Sasiadek, J.~Khe, Sensor fusion based on fuzzy kalman filter, in:
  Proceedings of the Second International Workshop on Robot Motion and Control,
  2001, pp. 275--283.

\bibitem{senthil2006nonlinear}
R.~Senthil, K.~Janarthanan, J.~Prakash, Nonlinear state estimation using fuzzy
  kalman filter, Industrial \& Engineering Chemistry Research 45~(25) (2006)
  8678--8688.

\bibitem{angelov2008autonomous}
P.~Angelov, R.~Ramezani, X.~Zhou, Autonomous novelty detection and object
  tracking in video streams using evolving clustering and takagi-sugeno type
  neuro-fuzzy system, in: IEEE International Joint Conference on Neural
  Networks (IJCNN), 2008, pp. 1456--1463.

\bibitem{angelov2004approach}
P.~P. Angelov, D.~P. Filev, An approach to online identification of
  takagi-sugeno fuzzy models, IEEE Transactions on Systems, Man, and
  Cybernetics, Part B: Cybernetics 34~(1) (2004) 484--498.

\bibitem{angelov2005simpl_ets}
P.~Angelov, D.~Filev, Simpl\_ets: a simplified method for learning evolving
  takagi-sugeno fuzzy models, in: IEEE International Conference on Fuzzy
  Systems (FUZZ), 2005, pp. 1068--1073.

\bibitem{wu2008fuzzy}
H.~Wu, F.~Sun, H.~Liu, Fuzzy particle filtering for uncertain systems, IEEE
  Transactions on Fuzzy Systems 16~(5) (2008) 1114--1129.

\bibitem{yoon2013object}
C.~Yoon, M.~Cheon, M.~Park, Object tracking from image sequences using adaptive
  models in fuzzy particle filter, Information Sciences 253 (2013) 74--99.

\bibitem{kamel2005fuzzy}
H.~Kamel, W.~Badawy, Fuzzy logic based particle filter for tracking a
  maneuverable target, in: 48th Midwest Symposium on Circuits and Systems,
  2005, pp. 1537--1540.

\bibitem{kim2007fuzzy}
Y.-J. Kim, C.-H. Won, J.-M. Pak, M.-T. Lim, Fuzzy adaptive particle filter for
  localization of a mobile robot, in: Knowledge-Based Intelligent Information
  and Engineering Systems, Springer, 2007, pp. 41--48.

\bibitem{horn1981determining}
B.~K. Horn, B.~G. Schunck, Determining optical flow, in: 1981 Technical
  Symposium East, International Society for Optics and Photonics, 1981, pp.
  319--331.

\bibitem{beauchemin1995computation}
S.~Beauchemin, J.~L. Barron, The computation of optical flow, ACM Computing
  Surveys 27~(3) (1995) 433--466.

\bibitem{bhattacharyya2009high}
S.~Bhattacharyya, U.~Maulik, P.~Dutta, High-speed target tracking by fuzzy
  hostility-induced segmentation of optical flow field, Applied Soft Computing
  9~(1) (2009) 126--134.

\bibitem{bhattacharyya2013target}
S.~Bhattacharyya, U.~Maulik, Target tracking using fuzzy hostility induced
  segmentation of optical flow field, in: Soft Computing for Image and
  Multimedia Data Processing, Springer, 2013, pp. 97--107.

\bibitem{bhattacharyya2007binary}
S.~Bhattacharyya, P.~Dutta, U.~Maulik, Binary object extraction using
  bi-directional self-organizing neural network (bdsonn) architecture with
  fuzzy context sensitive thresholding, Pattern Analysis and Applications
  10~(4) (2007) 345--360.

\bibitem{xie2004multi}
D.~Xie, W.~Hu, T.~Tan, J.~Peng, A multi-object tracking system for surveillance
  video analysis, in: Proceedings of the 17th International Conference on
  Pattern Recognition (ICPR), Vol.~4, 2004, pp. 767--770.

\bibitem{heisele1997tracking}
B.~Heisele, U.~Kressel, W.~Ritter, Tracking non-rigid, moving objects based on
  color cluster flow, in: IEEE Conference on Computer Vision and Pattern
  Recognition (CVPR), 1997, pp. 257--260.

\bibitem{pece2002cluster}
A.~E. Pece, From cluster tracking to people counting, in: IEEE Workshop on
  Performance Evaluation of Tracking and Surveillance (PETS), 2002, pp. 9--17.

\bibitem{lyons1999automatic}
M.~J. Lyons, J.~Budynek, S.~Akamatsu, Automatic classification of single facial
  images, IEEE Transactions on Pattern Analysis and Machine Intelligence
  21~(12) (1999) 1357--1362.

\bibitem{wu1999vision}
Y.~Wu, T.~S. Huang, Vision-based gesture recognition: A review, in:
  Gesture-based communication in human-computer interaction, Springer, 1999,
  pp. 103--115.

\bibitem{mitra2007gesture}
S.~Mitra, T.~Acharya, Gesture recognition: A survey, IEEE Transactions on
  Systems, Man, and Cybernetics, Part C: Applications and Reviews 37~(3) (2007)
  311--324.

\bibitem{wachs2002real}
J.~Wachs, U.~Kartoun, H.~Stern, Y.~Edan, Real-time hand gesture telerobotic
  system using fuzzy c-means clustering, in: Proceedings of the 5th Biannual
  World Automation Congress, Vol.~13, 2002, pp. 403--409.

\bibitem{wachs2005cluster}
J.~P. Wachs, H.~Stern, Y.~Edan, Cluster labeling and parameter estimation for
  the automated setup of a hand-gesture recognition system, IEEE Transactions
  on Systems, Man and Cybernetics, Part A: Systems and Humans 35~(6) (2005)
  932--944.

\bibitem{li2003gesture}
X.~Li, Gesture recognition based on fuzzy c-means clustering algorithm,
  Department Of Computer Science The University Of Tennessee Knoxville.

\bibitem{verma2009vision}
R.~Verma, A.~Dev, Vision based hand gesture recognition using finite state
  machines and fuzzy logic, in: International Conference on Ultra Modern
  Telecommunications \& Workshops (ICUMT), 2009, pp. 1--6.

\bibitem{al2001recognition}
O.~Al-Jarrah, A.~Halawani, Recognition of gestures in arabic sign language
  using neuro-fuzzy systems, Artificial Intelligence 133~(1) (2001) 117--138.

\bibitem{binh2005hand}
N.~D. Binh, T.~Ejima, Hand gesture recognition using fuzzy neural network, in:
  Proc. ICGST Conf. Graphics, Vision and Image Proces, 2005, pp. 1--6.

\bibitem{varkonyi2011human}
A.~R. V{\'a}rkonyi-K{\'o}czy, B.~Tusor, Human--computer interaction for smart
  environment applications using fuzzy hand posture and gesture models, IEEE
  Transactions on Instrumentation and Measurement 60~(5) (2011) 1505--1514.

\bibitem{carpenter1992fuzzy}
G.~A. Carpenter, S.~Grossberg, N.~Markuzon, J.~H. Reynolds, D.~B. Rosen, Fuzzy
  artmap: A neural network architecture for incremental supervised learning of
  analog multidimensional maps, IEEE Transactions on Neural Networks 3~(5)
  (1992) 698--713.

\bibitem{hussain1994novel}
B.~Hussain, M.~R. Kabuka, A novel feature recognition neural network and its
  application to character recognition, IEEE Transactions on Pattern Analysis
  and Machine Intelligence 16~(1) (1994) 98--106.

\bibitem{le2012fuzzy}
J.-M. Le~Yaouanc, J.-P. Poli, A fuzzy spatio-temporal-based approach for
  activity recognition, in: Advances in Conceptual Modeling, Springer, 2012,
  pp. 314--323.

\bibitem{yao2014fuzzy}
B.~Yao, H.~Hagras, M.~J. Alhaddad, D.~Alghazzawi, A fuzzy logic-based system
  for the automation of human behavior recognition using machine vision in
  intelligent environments, Soft Computing (2014) 1--8.

\bibitem{acampora2012combining}
G.~Acampora, P.~Foggia, A.~Saggese, M.~Vento, Combining neural networks and
  fuzzy systems for human behavior understanding, in: IEEE Ninth International
  Conference on Advanced Video and Signal-Based Surveillance (AVSS), 2012, pp.
  88--93.

\bibitem{hosseini2013fuzzy}
M.-S. Hosseini, A.-M. Eftekhari-Moghadam, Fuzzy rule-based reasoning approach
  for event detection and annotation of broadcast soccer video, Applied Soft
  Computing 13~(2) (2013) 846--866.

\bibitem{gkalelis2008combining}
N.~Gkalelis, A.~Tefas, I.~Pitas, Combining fuzzy vector quantization with
  linear discriminant analysis for continuous human movement recognition, IEEE
  Transactions on Circuits and Systems for Video Technology 18~(11) (2008)
  1511--1521.

\bibitem{karayiannis1995fuzzy}
N.~B. Karayiannis, P.-I. Pai, Fuzzy vector quantization algorithms and their
  application in image compression, IEEE Transactions on Image Processing 4~(9)
  (1995) 1193--1201.

\bibitem{elliott1995hidden}
R.~J. Elliott, L.~Aggoun, J.~B. Moore, Hidden Markov Models, Springer, 1995.

\bibitem{bobick1995state}
A.~F. Bobick, A.~D. Wilson, A state-based technique for the summarization and
  recognition of gesture, in: International Conference on Computer Vision
  (ICCV), 1995, pp. 382--388.

\bibitem{campbell1995recognition}
L.~W. Campbell, A.~F. Bobick, Recognition of human body motion using phase
  space constraints, in: International Conference on Computer Vision (ICCV),
  1995, pp. 624--630.

\bibitem{oliver2000bayesian}
N.~M. Oliver, B.~Rosario, A.~P. Pentland, A bayesian computer vision system for
  modeling human interactions, IEEE Transactions on Pattern Analysis and
  Machine Intelligence 22~(8) (2000) 831--843.

\bibitem{wilson1999parametric}
A.~D. Wilson, A.~F. Bobick, Parametric hidden markov models for gesture
  recognition, IEEE Transactions on Pattern Analysis and Machine Intelligence
  21~(9) (1999) 884--900.

\bibitem{yamato1992recognizing}
J.~Yamato, J.~Ohya, K.~Ishii, Recognizing human action in time-sequential
  images using hidden markov model, in: IEEE Conference on Computer Vision and
  Pattern Recognition (CVPR), 1992, pp. 379--385.

\bibitem{mozafari2012novel}
K.~Mozafari, N.~M. Charkari, H.~S. Boroujeni, M.~Behrouzifar, A novel fuzzy hmm
  approach for human action recognition in video, in: Knowledge Technology,
  Springer, 2012, pp. 184--193.

\bibitem{lim2013fuzz}
C.~H. Lim, C.~S. Chan, Fuzzy action recognition for multiple views within
  single camera, in: IEEE International Conference on Fuzzy Systems (FUZZ),
  2013, pp. 1--8.

\bibitem{iosifidis2011person}
A.~Iosifidis, A.~Tefas, I.~Pitas, Person specific activity recognition using
  fuzzy learning and discriminant analysis, in: Proceedings of the 19th
  European Signal Processing Conference (EUSIPCO), 2011, pp. 1974--1978.

\bibitem{iosifidis2012activity}
A.~Iosifidis, A.~Tefas, I.~Pitas, Activity-based person identification using
  fuzzy representation and discriminant learning, IEEE Transactions on
  Information Forensics and Security 7~(2) (2012) 530--542.

\bibitem{Iosifidis2012}
A.~Iosifidis, A.~Tefas, N.~Nikolaidis, I.~Pitas, Multi-view human movement
  recognition based on fuzzy distances and linear discriminant analysis,
  Computer Vision and Image Understanding 116~(3) (2012) 347--360.

\bibitem{iosifidis2013minimum}
A.~Iosifidis, A.~Tefas, I.~Pitas, Minimum class variance extreme learning
  machine for human action recognition, IEEE Transactions on Circuits and
  Systems for Video Technology 23~(11) (2013) 1968--1979.

\bibitem{Iosifidis2012b}
A.~Iosifidis, A.~Tefas, I.~Pitas, Multi-view action recognition based on action
  volumes, fuzzy distances and cluster discriminant analysis, Signal Processing
  93~(6) (2012) 1445--1457.

\bibitem{kratz2009anomaly}
L.~Kratz, K.~Nishino, Anomaly detection in extremely crowded scenes using
  spatio-temporal motion pattern models, in: IEEE Conference on Computer Vision
  and Pattern Recognition (CVPR), 2009, pp. 1446--1453.

\bibitem{wu2010chaotic}
S.~Wu, B.~E. Moore, M.~Shah, Chaotic invariants of lagrangian particle
  trajectories for anomaly detection in crowded scenes, in: IEEE Conference on
  Computer Vision and Pattern Recognition (CVPR), 2010, pp. 2054--2060.

\bibitem{anderson2008extension}
D.~Anderson, R.~H. Luke, J.~M. Keller, M.~Skubic, Extension of a soft-computing
  framework for activity analysis from linguistic summarizations of video, in:
  IEEE International Conference on Fuzzy Systems (FUZZ), 2008, pp. 1404--1410.

\bibitem{yu2011fall}
M.~Yu, S.~M. Naqvi, A.~Rhuma, J.~Chambers, Fall detection in a smart room by
  using a fuzzy one class support vector machine and imperfect training data,
  in: IEEE International Conference on Acoustics, Speech and Signal Processing
  (ICASSP), 2011, pp. 1833--1836.

\bibitem{wongkhuenkaew2013multi}
R.~Wongkhuenkaew, S.~Auephanwiriyakul, N.~Theera-Umpon, Multi-prototype fuzzy
  clustering with fuzzy k-nearest neighbor for off-line human action
  recognition, in: IEEE International Conference on Fuzzy Systems (FUZZ), 2013,
  pp. 1--7.

\bibitem{chan2010fuzzy}
C.~S. Chan, H.~Liu, W.~K. Lai, Fuzzy qualitative complex actions recognition,
  in: IEEE International Conference on Fuzzy Systems (FUZZ), 2010, pp. 1--8.

\bibitem{wang2008detecting}
Z.~Wang, J.~Zhang, Detecting pedestrian abnormal behavior based on fuzzy
  associative memory, in: Fourth International Conference on Natural
  Computation (ICNC), Vol.~6, 2008, pp. 143--147.

\bibitem{juang2007human}
C.-F. Juang, C.-M. Chang, Human body posture classification by a neural fuzzy
  network and home care system application, EEE Transactions on Systems, Man
  and Cybernetics, Part A: Systems and Humans 37~(6) (2007) 984--994.

\bibitem{hu2004learning}
W.~Hu, D.~Xie, T.~Tan, S.~Maybank, Learning activity patterns using fuzzy
  self-organizing neural network, IEEE Transactions on Systems, Man, and
  Cybernetics, Part B: Cybernetics 34~(3) (2004) 1618--1626.

\bibitem{Zadeh1997111}
L.~A. Zadeh, Toward a theory of fuzzy information granulation and its
  centrality in human reasoning and fuzzy logic, Fuzzy Sets and Systems 90~(2)
  (1997) 111 -- 127.

\bibitem{bezdek1992computing}
J.~C. Bezdek, Computing with uncertainty, IEEE Communications Magazine 30~(9)
  (1992) 24--36.

\bibitem{yager2002uncertainty}
R.~R. Yager, Uncertainty representation using fuzzy measures, IEEE Transactions
  on Systems, Man, and Cybernetics, Part B: Cybernetics 32~(1) (2002) 13--20.

\bibitem{wu2002uncertainty}
H.~Wu, J.~M. Mendel, Uncertainty bounds and their use in the design of interval
  type-2 fuzzy logic systems, IEEE Transactions on Fuzzy Systems 10~(5) (2002)
  622--639.

\bibitem{wu2007uncertainty}
D.~Wu, J.~M. Mendel, Uncertainty measures for interval type-2 fuzzy sets,
  Information Sciences 177~(23) (2007) 5378--5393.

\bibitem{zeng2006type}
J.~Zeng, Z.-Q. Liu, Type-2 fuzzy sets for handling uncertainty in pattern
  recognition, in: IEEE International Conference on Fuzzy Systems (FUZZ), 2006,
  pp. 1247--1252.

\bibitem{zadeh1996fuzzy}
L.~Zadeh, Fuzzy logic= computing with words, IEEE Transactions on Fuzzy Systems
  4~(2) (1996) 103--111.

\bibitem{zadeh1973outline}
L.~Zadeh, Outline of a new approach to the analysis of complex systems and
  decision processes, IEEE Transactions on Systems, Man and Cybernetics~(1)
  (1973) 28--44.

\bibitem{rubin1999computing}
S.~H. Rubin, Computing with words, IEEE Transactions on Systems, Man, and
  Cybernetics, Part B: Cybernetics 29~(4) (1999) 518--524.

\bibitem{trivino2008linguistic}
G.~Trivino, A.~van~der Heide, Linguistic summarization of the human activity
  using skin conductivity and accelerometers, in: Proc. 12th Int. Conf. on
  Information Processing and Management of Uncertainty in Knowledge-based
  Systems (IPMU), 2008.

\bibitem{kacprzyk2001linguistic}
J.~Kacprzyk, R.~R. Yager, Linguistic summaries of data using fuzzy logic,
  International Journal of General System 30~(2) (2001) 133--154.

\bibitem{anderson2011linguistic}
D.~T. Anderson, J.~M. Keller, M.~Anderson, D.~J. Wescott, Linguistic
  description of adult skeletal age-at-death estimations from fuzzy integral
  acquired fuzzy sets, in: IEEE International Conference on Fuzzy Systems
  (FUZZ), 2011, pp. 2274--2281.

\bibitem{wilbik2011linguistic}
A.~Wilbik, J.~M. Keller, G.~L. Alexander, Linguistic summarization of sensor
  data for eldercare, in: IEEE International Conference on Systems, Man, and
  Cybernetics (SMC), 2011, pp. 2595--2599.

\bibitem{wilbik2013fuzzy}
A.~Wilbik, J.~Keller, A fuzzy measure similarity between sets of linguistic
  summaries, IEEE Transactions on Fuzzy Systems 21~(1) (2013) 183--189.

\bibitem{wang1992generating}
L.-X. Wang, J.~M. Mendel, Generating fuzzy rules by learning from examples,
  IEEE Transactions on Systems, Man and Cybernetics 22~(6) (1992) 1414--1427.

\bibitem{rhee1993fuzzy}
F.~C.-H. Rhee, R.~Krishnapuram, Fuzzy rule generation methods for high-level
  computer vision, Fuzzy Sets and Systems 60~(3) (1993) 245--258.

\bibitem{wang2001new}
X.~Wang, Y.~Wang, X.~Xu, W.~Ling, D.~S. Yeung, A new approach to fuzzy rule
  generation: fuzzy extension matrix, Fuzzy Sets and Systems 123~(3) (2001)
  291--306.

\bibitem{cheng1995fast}
T.~W. Cheng, D.~Goldgof, L.~Hall, Fast clustering with application to fuzzy
  rule generation, in: IEEE International Conference on Fuzzy Systems (FUZZ),
  Vol.~4, 1995, pp. 2289--2295.

\bibitem{cordon2001generating}
O.~Cord{\'o}n, F.~Herrera, P.~Villar, Generating the knowledge base of a fuzzy
  rule-based system by the genetic learning of the data base, IEEE Transactions
  on Fuzzy Systems 9~(4) (2001) 667--674.

\bibitem{mitra2000neuro}
S.~Mitra, Y.~Hayashi, Neuro-fuzzy rule generation: survey in soft computing
  framework, IEEE Transactions on Neural Networks 11~(3) (2000) 748--768.

\bibitem{sapienza2014learning}
M.~Sapienza, F.~Cuzzolin, P.~H. Torr, Learning discriminative space--time
  action parts from weakly labelled videos, International Journal of Computer
  Vision (2014) 1--18.

\bibitem{fisher2004pets04}
R.~B. Fisher, The pets04 surveillance ground-truth data sets, in: Proc. 6th
  IEEE International Workshop on Performance Evaluation of Tracking and
  Surveillance (PETS), 2004, pp. 1--5.

\bibitem{nie2014single}
W.~Nie, A.~Liu, Y.~Su, H.~Luan, Z.~Yang, L.~Cao, R.~Ji, Single/cross-camera
  multiple-person tracking by graph matching, Neurocomputing 139 (2014)
  220--232.

\bibitem{chen2009recognizing}
C.-C. Chen, J.~Aggarwal, Recognizing human action from a far field of view, in:
  Workshop on Motion and Video Computing, 2009, pp. 1--7.

\bibitem{wu2014multi}
D.~Wu, L.~Shao, Multi-max-margin support vector machine for multi-source human
  action recognition, Neurocomputing 127 (2014) 98--103.

\bibitem{wang2007human}
Y.~Wang, K.~Huang, T.~Tan, Human activity recognition based on r transform, in:
  IEEE Conference on Computer Vision and Pattern Recognition (CVPR), 2007, pp.
  1--8.

\bibitem{lu2014application}
Y.~Lu, K.~Boukharouba, J.~Boon{\ae}rt, A.~Fleury, S.~Lec{\oe}uche, Application
  of an incremental svm algorithm for on-line human recognition from video
  surveillance using texture and color features, Neurocomputing 126 (2014)
  132--140.

\bibitem{nghiem2007etiseo}
A.~T. Nghiem, F.~Bremond, M.~Thonnat, V.~Valentin, Etiseo, performance
  evaluation for video surveillance systems, in: IEEE Conference on Advanced
  Video and Signal Based Surveillance (AVSS), 2007, pp. 476--481.

\bibitem{simha2014feature}
S.~M. Simha, D.~P. Chau, F.~Bremond, et~al., Feature matching using co-inertia
  analysis for people tracking, in: The 9th International Conference on
  Computer Vision Theory and Applications (VISAPP), 2014.

\bibitem{ikizler2007human}
N.~Ikizler, P.~Duygulu, Human action recognition using distribution of oriented
  rectangular patches, in: Human Motion--Understanding, Modeling, Capture and
  Animation, Springer, 2007, pp. 271--284.

\bibitem{tran2008human}
D.~Tran, A.~Sorokin, Human activity recognition with metric learning, in:
  European Conference on Computer Vision (ECCV), Springer, 2008, pp. 548--561.

\bibitem{tu2014complex}
H.-b. Tu, L.-m. Xia, Z.-w. Wang, The complex action recognition via the
  correlated topic model, The Scientific World Journal 2014.

\bibitem{de2008guide}
F.~De~la Torre, J.~Hodgins, A.~Bargteil, X.~Martin, J.~Macey, A.~Collado,
  P.~Beltran, {Guide to the Carnegie Mellon University Multimodal Activity
  (CMU-MMAC) Database}, in: Tech. report CMU-RI-TR-08-22, Robotics Institute,
  Carnegie Mellon University, 2008.

\bibitem{john2014charting}
V.~John, E.~Trucco, Charting-based subspace learning for video-based human
  action classification, Machine Vision and Applications 25~(1) (2014)
  119--132.

\bibitem{ragheb2008vihasi}
H.~Ragheb, S.~Velastin, P.~Remagnino, T.~Ellis, Vihasi: virtual human action
  silhouette data for the performance evaluation of silhouette-based action
  recognition methods, in: Second ACM/IEEE International Conference on
  Distributed Smart Cameras (ICDSC), 2008, pp. 1--10.

\bibitem{zhang2013grassmann}
L.~Zhang, D.~Tao, X.~Liu, L.~Sun, M.~Song, C.~Chen, Grassmann multimodal
  implicit feature selection, Multimedia Systems (2013) 1--16.

\bibitem{laptev2008learning}
I.~Laptev, M.~Marszalek, C.~Schmid, B.~Rozenfeld, Learning realistic human
  actions from movies, in: IEEE Conference on Computer Vision and Pattern
  Recognition (CVPR), 2008, pp. 1--8.

\bibitem{du2014recognizing}
J.-X. Du, C.-M. Zhai, Y.-L. Guo, Y.-Y. Tang, P.~C. Chun~Lung, Recognizing
  complex events in real movies by combining audio and video features,
  Neurocomputing 137 (2014) 89--95.

\bibitem{marszalek2009actions}
M.~Marszalek, I.~Laptev, C.~Schmid, Actions in context, in: IEEE Conference on
  Computer Vision and Pattern Recognition (CVPR), 2009, pp. 2929--2936.

\bibitem{wang2013action}
H.~Wang, C.~Schmid, Action recognition with improved trajectories, in:
  International Conference on Computer Vision (ICCV), 2013.

\bibitem{action-shah-action-mach-cvpr08}
M.~Rodriguez, J.~Ahmed, M.~Shah, Action mach: Maximum average correlation
  height filter for action recognition, in: IEEE Conference on Computer Vision
  and Pattern Recognition (CVPR), 2008, pp. 1--8.

\bibitem{wu2011action}
S.~Wu, O.~Oreifej, M.~Shah, Action recognition in videos acquired by a moving
  camera using motion decomposition of lagrangian particle trajectories, in:
  IEEE International Conference on Computer Vision (ICCV), 2011, pp.
  1419--1426.

\bibitem{liu2009recognizing}
J.~Liu, J.~Luo, M.~Shah, Recognizing realistic actions from videos in the wild,
  in: IEEE Conference on Computer Vision and Pattern Recognition (CVPR), 2009,
  pp. 1996--2003.

\bibitem{gkalelis2009i3dpost}
N.~Gkalelis, H.~Kim, A.~Hilton, N.~Nikolaidis, I.~Pitas, The i3dpost multi-view
  and 3d human action/interaction database, in: Conference for Visual Media
  Production (CVMP), 2009, pp. 159--168.

\bibitem{holte2012local}
M.~B. Holte, B.~Chakraborty, J.~Gonzalez, T.~B. Moeslund, A local 3-d motion
  descriptor for multi-view human action recognition from 4-d spatio-temporal
  interest points, IEEE Journal of Selected Topics in Signal Processing 6~(5)
  (2012) 553--565.

\bibitem{ryoo2009spatio}
M.~S. Ryoo, J.~K. Aggarwal, Spatio-temporal relationship match: Video structure
  comparison for recognition of complex human activities, in: IEEE
  International Conference on Computer Vision (ICCV), 2009, pp. 1593--1600.

\bibitem{fu2014interactive}
Y.~Fu, Y.~Jia, Y.~Kong, Interactive phrases: Semantic descriptions for human
  interaction recognition, IEEE Transactions on Pattern Analysis and Machine
  Intelligence.

\bibitem{yuan2009discriminative}
J.~Yuan, Z.~Liu, Y.~Wu, Discriminative subvolume search for efficient action
  detection, in: IEEE Conference on Computer Vision and Pattern Recognition
  (CVPR), 2009, pp. 2442--2449.

\bibitem{li2010action}
W.~Li, Z.~Zhang, Z.~Liu, Action recognition based on a bag of 3d points, in:
  2010 IEEE Computer Society Conference on Computer Vision and Pattern
  Recognition Workshops (CVPRW), 2010, pp. 9--14.

\bibitem{yang2014effective}
X.~Yang, Y.~Tian, Effective 3d action recognition using eigenjoints, Journal of
  Visual Communication and Image Representation 25~(1) (2014) 2--11.

\bibitem{blunsden2010behave}
S.~Blunsden, R.~Fisher, The behave video dataset: ground truthed video for
  multi-person behavior classification, Annals of the BMVA 2010~(4) (2010)
  1--12.

\bibitem{cheng2014recognizing}
Z.~Cheng, L.~Qin, Q.~Huang, S.~Yan, Q.~Tian, Recognizing human group action by
  layered model with multiple cues, Neurocomputing 136 (2014) 124--135.

\bibitem{singh2010muhavi}
S.~Singh, S.~A. Velastin, H.~Ragheb, Muhavi: A multicamera human action video
  dataset for the evaluation of action recognition methods, in: 2010 Seventh
  IEEE International Conference on Advanced Video and Signal Based Surveillance
  (AVSS), 2010, pp. 48--55.

\bibitem{chaaraoui2013optimizing}
A.~A. Chaaraoui, F.~Fl{\'o}rez-Revuelta, Optimizing human action recognition
  based on a cooperative coevolutionary algorithm, Engineering Applications of
  Artificial Intelligence 31 (2013) 116--125.

\bibitem{niebles2010modeling}
J.~C. Niebles, C.-W. Chen, L.~Fei-Fei, Modeling temporal structure of
  decomposable motion segments for activity classification, in: European
  Conference on Computer Vision (ECCV), Springer, 2010, pp. 392--405.

\bibitem{patron2010high}
A.~Patron-Perez, M.~Marszalek, A.~Zisserman, I.~Reid, High five: Recognising
  human interactions in tv shows, in: Proceedings of the British Machine Vision
  Conference (BMVC), 2010, pp. 50.1--50.11.

\bibitem{marin2014human}
M.~Mar{\'\i}n-Jim{\'e}nez, R.~Mu{\~n}oz-Salinas, E.~Yeguas-Bolivar, N.~P. de~la
  Blanca, Human interaction categorization by using audio-visual cues, Machine
  Vision and Applications 25~(1) (2014) 71--84.

\bibitem{kuehne2011hmdb}
H.~Kuehne, H.~Jhuang, E.~Garrote, T.~Poggio, T.~Serre, Hmdb: a large video
  database for human motion recognition, in: IEEE International Conference on
  Computer Vision (ICCV), 2011, pp. 2556--2563.

\bibitem{denina2011videoweb}
G.~Denina, B.~Bhanu, H.~T. Nguyen, C.~Ding, A.~Kamal, C.~Ravishankar,
  A.~Roy-Chowdhury, A.~Ivers, B.~Varda, Videoweb dataset for multi-camera
  activities and non-verbal communication, in: Distributed Video Sensor
  Networks, Springer, 2011, pp. 335--347.

\bibitem{zha2013detecting}
Z.-J. Zha, H.~Zhang, M.~Wang, H.~Luan, T.-S. Chua, Detecting group activities
  with multi-camera context, IEEE Transactions on Circuits and Systems for
  Video Technology 23~(5) (2013) 856--869.

\bibitem{soomro2012ucf101}
K.~Soomro, A.~R. Zamir, M.~Shah, Ucf101: A dataset of 101 human actions classes
  from videos in the wild, Tech. Rep. CRCV-TR-12-01, CRCV, University of
  Central Florida (November 2012).

\bibitem{caimulti}
Z.~Cai, L.~Wang, X.~Peng, Y.~Qiao, Multi-view super vector for action
  recognition, in: IEEE International Conference on Computer Vision and Pattern
  Recognition (CVPR), 2014.

\bibitem{reddy2013recognizing}
K.~K. Reddy, M.~Shah, Recognizing 50 human action categories of web videos,
  Machine Vision and Applications 24~(5) (2013) 971--981.

\bibitem{boutell2004learning}
M.~R. Boutell, J.~Luo, X.~Shen, C.~M. Brown, Learning multi-label scene
  classification, Pattern recognition 37~(9) (2004) 1757--1771.

\bibitem{parikh2011relative}
D.~Parikh, K.~Grauman, Relative attributes, in: IEEE International Conference
  on Computer Vision (ICCV), 2011, pp. 503--510.

\bibitem{lim2012fuzzy}
C.~H. Lim, C.~S. Chan, A fuzzy qualitative approach for scene classification,
  in: IEEE International Conference on Fuzzy Systems (FUZZ), 2012, pp. 1--8.

\bibitem{kitani2012activity}
K.~M. Kitani, B.~D. Ziebart, J.~A. Bagnell, M.~Hebert, Activity forecasting,
  in: European Conference on Computer Vision (ECCV), Springer, 2012, pp.
  201--214.

\bibitem{ryoo2011human}
M.~Ryoo, Human activity prediction: Early recognition of ongoing activities
  from streaming videos, in: IEEE International Conference on Computer Vision
  (ICCV), 2011, pp. 1036--1043.

\bibitem{hoai2012max}
M.~Hoai, F.~De~la Torre, Max-margin early event detectors, in: IEEE Conference
  on Computer Vision and Pattern Recognition (CVPR), 2012, pp. 2863--2870.

\bibitem{gupta2009observing}
A.~Gupta, A.~Kembhavi, L.~S. Davis, Observing human-object interactions: Using
  spatial and functional compatibility for recognition, IEEE Transactions on
  Pattern Analysis and Machine Intelligence 31~(10) (2009) 1775--1789.

\bibitem{yao2010grouplet}
B.~Yao, L.~Fei-Fei, Grouplet: A structured image representation for recognizing
  human and object interactions, in: IEEE Conference on Computer Vision and
  Pattern Recognition (CVPR), 2010, pp. 9--16.

\bibitem{desai2010discriminative}
C.~Desai, D.~Ramanan, C.~Fowlkes, Discriminative models for static human-object
  interactions, in: IEEE Computer Society Conference on Computer Vision and
  Pattern Recognition Workshops (CVPRW), 2010, pp. 9--16.

\bibitem{yang2010recognizing}
W.~Yang, Y.~Wang, G.~Mori, Recognizing human actions from still images with
  latent poses, in: IEEE Conference on Computer Vision and Pattern Recognition
  (CVPR), 2010, pp. 2030--2037.

\bibitem{delaitre2010recognizing}
V.~Delaitre, I.~Laptev, J.~Sivic, Recognizing human actions in still images: a
  study of bag-of-features and part-based representations, in: Proceedings of
  the British Machine Vision Conference (BMVC), 2010, pp. 97.1--97.11.

\bibitem{maji2011action}
S.~Maji, L.~Bourdev, J.~Malik, Action recognition from a distributed
  representation of pose and appearance, in: IEEE Conference on Computer Vision
  and Pattern Recognition (CVPR), 2011, pp. 3177--3184.

\bibitem{prest2012weakly}
A.~Prest, C.~Schmid, V.~Ferrari, Weakly supervised learning of interactions
  between humans and objects, IEEE Transactions on Pattern Analysis and Machine
  Intelligence 34~(3) (2012) 601--614.

\bibitem{yao2012recognizing}
B.~Yao, L.~Fei-Fei, Recognizing human-object interactions in still images by
  modeling the mutual context of objects and human poses, IEEE Transactions on
  Pattern Analysis and Machine Intelligence 34~(9) (2012) 1691--1703.

\end{thebibliography}

%







\end{document}